\documentclass[10pt,journal,compsoc]{IEEEtran}
\ifCLASSOPTIONcompsoc
  \usepackage[nocompress]{cite}
\else
  \usepackage{cite}
\fi
\ifCLASSINFOpdf
\else
\fi
\usepackage{graphicx}
\usepackage{algorithm}
\usepackage{algorithmic}
\usepackage{amssymb}
\usepackage{subfigure}
\usepackage{diagbox}
\usepackage{tabularx}
\usepackage{amsmath}
\usepackage{indentfirst} 
\usepackage{color}
\usepackage{mathrsfs}

\begin{document}
%
\title{Asynchronous Decentralized Federated Learning for Collaborative Fault Diagnosis of PV Stations}
%
%
%
%

\author{Qi~Liu,
        Bo~Yang,~\IEEEmembership{Senior Member,~IEEE,}
        Zhaojian~Wang,~\IEEEmembership{Member,~IEEE,}
        Dafeng~Zhu,
        Xinyi~Wang,  \\
        Kai Ma,~\IEEEmembership{Member,~IEEE,}      
        and~Xinping~Guan,~\IEEEmembership{Fellow,~IEEE}
\IEEEcompsocitemizethanks{\IEEEcompsocthanksitem Qi Liu, Bo Yang, Zhaojian Wang, Dafeng Zhu, Xinyi Wang, and Xinping Guan are with the Department of Automation, Shanghai Jiao Tong University, Shanghai 200240, China; Key Laboratory of System Control and Information Processing, Ministry of Education of China, Shanghai 200240, China; Shanghai Engineering Research Center of Intelligent Control and Management, Shanghai 200240, China. Kai Ma is with School of Electrical Engineering, Yanshan University, Qinhuangdao 066004, China. Bo Yang is the corresponding author. 
E-mail: bo.yang@sjtu.edu.cn}}

\IEEEtitleabstractindextext{%
\begin{abstract}
Due to the different losses caused by various photovoltaic (PV) array faults, accurate diagnosis of fault types is becoming increasingly important.
Compared with a single one, multiple PV stations collect sufficient fault samples, but their data is not allowed to be shared directly due to potential conflicts of interest.
Therefore, federated learning can be exploited to train a collaborative fault diagnosis model.
However, the modeling efficiency is seriously affected by the model update mechanism since each PV station has a different computing capability and amount of data. 
Moreover, for the safe and stable operation of the PV system, the robustness of collaborative modeling must be guaranteed rather than simply being processed on a central server.  
To address these challenges, a novel asynchronous decentralized federated learning (ADFL) framework is proposed. 
Each PV station not only trains its local model but also participates in collaborative fault diagnosis by exchanging model parameters to improve the generalization without losing accuracy.
The global model is aggregated distributedly to avoid central node failure. 
By designing the asynchronous update scheme, the communication overhead and training time are greatly reduced.
Both the experiments and numerical simulations are carried out to verify the effectiveness of the proposed method.
\end{abstract}

}

\maketitle

\IEEEdisplaynontitleabstractindextext

%
\IEEEpeerreviewmaketitle

\IEEEraisesectionheading{\section{Introduction}\label{sec:introduction}}
%
%
%
%
\IEEEPARstart{T}{he} large consumption of fossil fuels has caused serious issues, such as energy shortages, environmental pollution, and global warming \cite{1ElisaPapadis}. As a renewable and sustainable energy source, solar energy is attracting more and more attention, where photovoltaics is one of the primary ways \cite{2Sainbold}.
The photovoltaic (PV) arrays are the most fundamental part of the PV system and are prone to suffer from various faults due to the harsh outdoor environment \cite{30ZhicongChen}.
Since these faults have different levels of harm to the PV system, it is of great significance to accurately diagnose the types of faults, thereby reducing the economic loss of power generation and avoiding safety accidents.

In the last decades, many studies focused on PV fault diagnosis have been carried out. 
The existing methods can be mainly categorized into three types: sensing-based methods \cite{3MC,4TakumiTakashima,5Saleh}, electrical characteristics-based methods \cite{6ChenLeian,7Kumar,9SainboldSaranchimeg} and machine learning-based (ML) methods \cite{11FengJian,12YiZhehan}. 
The sensing-based methods and electrical characteristics-based methods rely on manual fault feature extraction and some key parameters need to be determined through specific experiments according to the topology, which reduces the generalization of models and is not suitable for increasingly complex PV stations. 
Compared with them, the ML-based methods are data-driven and automatically extract fault features through a large number of samples, which has occupied a dominant position in the field of fault diagnosis.

However, the existing ML-based methods developed fault diagnosis models based on a single PV station without considering the case that both the types and number of fault samples are insufficient. They either assumed that the collected fault samples were sufficient and improved the accuracy of fault diagnosis through advanced neural networks, or assumed that the number of fault samples was insufficient but already contained all types of faults to be diagnosed. 
Although these methods are practically important and achieve high accuracy, it is equally significant to study scenarios involving multiple PV stations to collaboratively train a shared fault diagnosis model, which is considered an important way to break data island \cite{33Lu,34Savazzi},\cite{9184854}. 
In fact, it is difficult for a single PV station to collect samples of all fault types, and there is an unbalanced distribution of samples from multiple PV stations. 
If these samples can be fully utilized together, the generalization of the model can be significantly improved without losing accuracy.
Thus, it is important to design a new ML framework to make multiple PV stations learn the fault features contained in the local data from each other.

To design a new collaborative fault diagnosis framework for multiple PV stations, the following three challenges need to be addressed: data island, framework robustness, and modeling efficiency.
First, the collected samples contain the privacy information of the PV system. Since multiple PV stations belong to different operators, they are not allowed to share the original data directly. A privacy-preserving data interaction framework should be adopted. 
Second, the validity and accuracy of the fault diagnosis model are essential to the safe and stable operation of the PV system. The robustness of the collaborative fault diagnosis modeling must be guaranteed. 
Third, the modeling efficiency is seriously affected by model update strategy due to different amounts of data and computing capabilities \cite{9224971}. 
Hence it is important to design a novel global model aggregation scheme to reduce the communication overhead and training time.

In this paper, we propose a novel federated learning (FL) method that combines the decentralized (serverless) framework and asynchronous update strategy for PV fault diagnosis.
The problem of the unbalanced distribution of fault samples and data island among multiple PV stations is addressed through collaborative fault diagnosis while ensuring data privacy, as well as the framework robustness and modeling efficiency issues.
The proposed method fully utilizes the data of multiple PV stations to improve the generalization of the model without losing accuracy and is suitable for various meteorological conditions.

We mainly focus on three common PV array faults in this paper: 1) short-circuit faults, defined as the accidental connection or low impedance between two points in the PV array \cite{22Alam}; 
2) degradation faults, defined as the increase in the equivalent series resistance or the decrease in the parallel resistance after a certain time of operation \cite{27Manish};  
3) partial shading faults, defined as the reduction of actual effective irradiance of the PV module due to surface dust, clouds, or other objects blocking the sunlight \cite{28Belhaouas}.

Our major contributions are summarized as follows:
\begin{itemize}
	\item This paper focuses on the collaborative modeling method in PV fault diagnosis, which enables multiple PV stations to learn the fault features from others without sharing original data, and improves model generalization. To solve the problem of server paralysis in the existing centralized FL methods, a novel asynchronous decentralized federated learning (ADFL) framework is proposed to achieve the distributed aggregation of global models.
	\item Considering the difference in computing capability and amount of data for each PV station, a new global model aggregation algorithm is proposed to dynamically determine the participating agents in each round. An asynchronous update scheme is designed to accelerate the global model aggregation, which greatly reduces communication overhead and training time. The theoretical convergence analysis of the proposed algorithm is given.
	\item Extensive experiments are carried out based on simulation and real-world data to show that our method enables each PV station to accurately identify more fault types and is suitable for various meteorological conditions. 
\end{itemize}

The remainder of this paper is organized as follows. Section 2 reviews the related work. Section 3 describes the system model. Section 4 introduces the ADFL based collaborative fault diagnosis method. Section 5 presents the experimental and numerical simulation results. Section 6 concludes the paper.

\section{Related Work}
In recent years, ML technology has been widely used in PV fault diagnosis, such as deep neural networks (DNNs) \cite{10Nikolaos}, support vector machines (SVM) \cite{12YiZhehan,13Matthias} and extreme learning machine (ELM) \cite{14HuangJunMing,15ZhicongChen}, etc.
However, these methods rely on a large amount of labeled PV data, which is usually difficult or expensive to obtain.
Aiming at the problem of insufficient fault samples, 
Feng \emph{et al.} \cite{11FengJian} integrated domain knowledge into the learning process and proposed a domain-knowledge-based deep-broad learning framework (DK-DBLF) to reduce demand for labeled samples and improve the model flexibility.
Zhao \emph{et al.} \cite{16ZhaoYe} proposed a graph-based semi-supervised learning model, which only used a few labeled training data to detect line-line faults and open-circuit faults in PV arrays.
Lu \emph{et al.} \cite{17LuShibo} proposed a deep convolutional generative adversarial network (DA-DCGAN) to convert normal data from the target domain into virtual fault data by employing domain adaptation.
However, the conventional ML methods mentioned above do not consider the case that both the types and number of fault samples are insufficient. They are mostly based on the assumption that samples of all fault types to be diagnosed have been included in the labeled dataset. 

\begin{figure}[!t]
	\centering
	\includegraphics[width=2.5in]{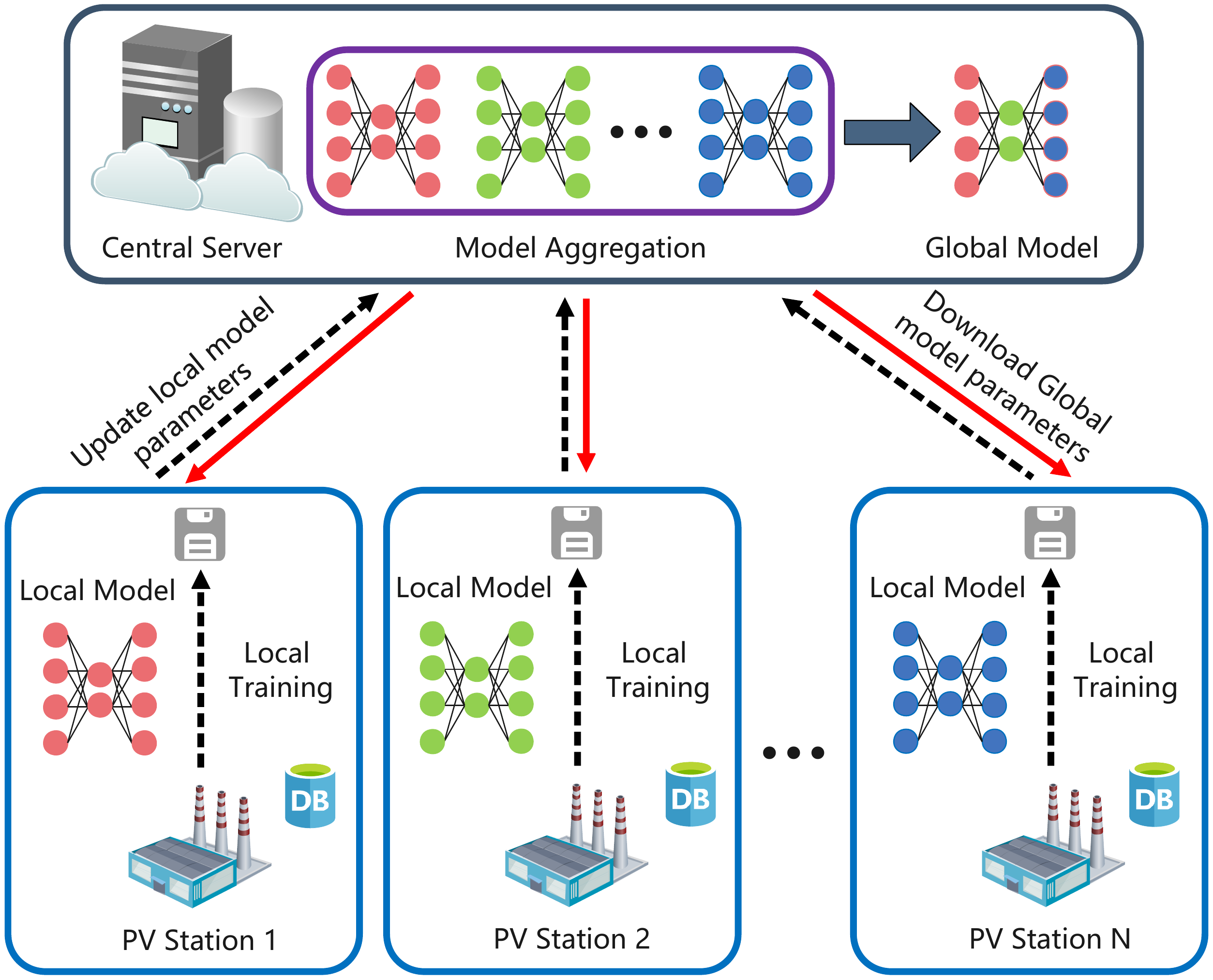}
	\caption{The traditional centralized FL framework}
	\label{fig_FL}
\end{figure}

FL is a privacy-preserving distributed machine learning paradigm \cite{9252911}. It is suitable for scenarios where the distribution of fault samples of multiple PV stations is unbalanced and they cannot share the original data directly due to privacy.
Specifically, as illustrated by Fig. \ref{fig_FL}, participants in FL only share model parameters without providing their original data.
After that, a shared global model is obtained by aggregating the model parameters of all participants, which has excellent accuracy and generalization.
The concept of FL was introduced by Google \cite{18mcmahan} and has been used in several fields \cite{19LiuYi,20GaoYujia,21LeeSangyoon,31Zhang}.
Y. Liu \emph{et al.} proposed an FL-based gated recurrent unit neural network for traffic flow prediction \cite{19LiuYi}.
Y. Gao \emph{et al.} in \cite{20GaoYujia} proposed a federated region-learning (FRL) framework to comprehensively use the information of multiple weather monitoring stations for weather forecasting. 
S. Lee \emph{et al.} in \cite{21LeeSangyoon} applied federated reinforcement learning to home energy management systems to reduce energy costs.
W. Zhang \emph{et al.} proposed an FL method based on dynamic fusion to collaboratively detect COVID-19 infection \cite{31Zhang}.
However, the methods mentioned above either used centralized FL or updated the global model synchronously.
Once the central server crashes, the collaborative modeling cannot be carried out. 
In addition, since the difference in computing capability and amount of data is not considered, long waiting time for synchronous update mechanism results in low modeling efficiency.

\section{System Model}
In this section, we first introduce the system configuration and then present the proposed ADFL framework.

\subsection{System Configuration}
\begin{figure}[htbp]
	\centering
	\includegraphics[width=2.5in]{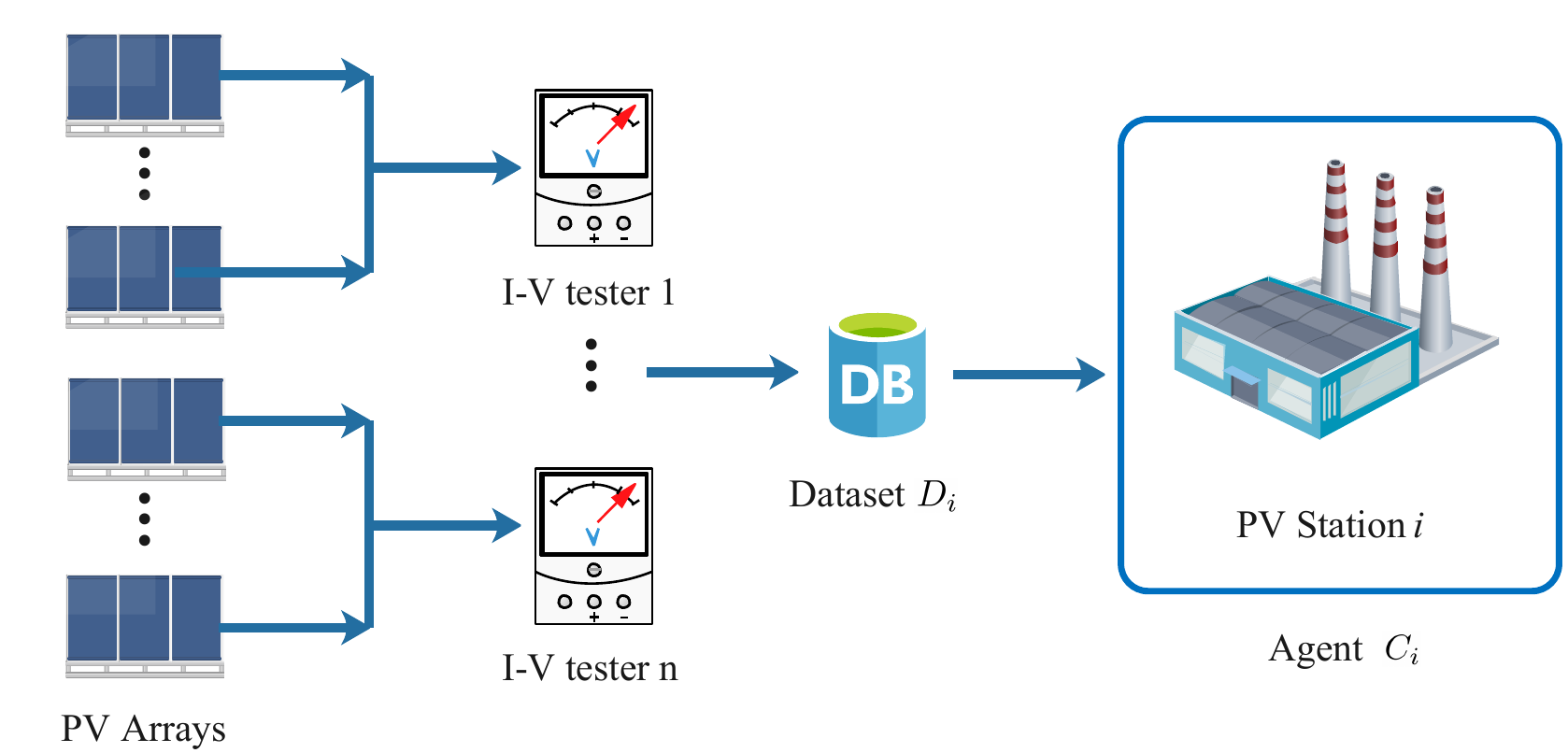}
	\caption{The structure of each agent}
	\label{fig_client}
\end{figure}

We assume that there are $N$ PV stations participating in collaborative fault diagnosis, and they belong to different owners.
All of these PV stations have data storage, processing, and transmission capability, and use their data to train the local fault diagnosis models based on the neural networks. We regard each PV station as an agent, and use the set $\mathcal{C}=\{{C}_{1}, {C}_{2}, \ldots {C}_{N}\}$ to represent all agents, where $C_{i}$ is responsible for PV station $i$. 
As shown in Fig. \ref{fig_client}, each agent is composed of several arrays and has its local dataset $D_{i}$. Specifically, the I-V characteristic curves, corresponding temperature and irradiance of each array can be collected by the I-V tester equipped with an environmental tester.

Considering the situation where both the types and number of samples are insufficient, we assume that each agent only has part of fault samples, and the fault types involved in the local data of each agent are not the same.
To implement the global model aggregation, each agent is supposed to use neural networks with the same structure to train its local fault diagnosis model.

\subsection{ADFL Framework}
\begin{figure}[!h]
	\centering
	\includegraphics[width=2.5in]{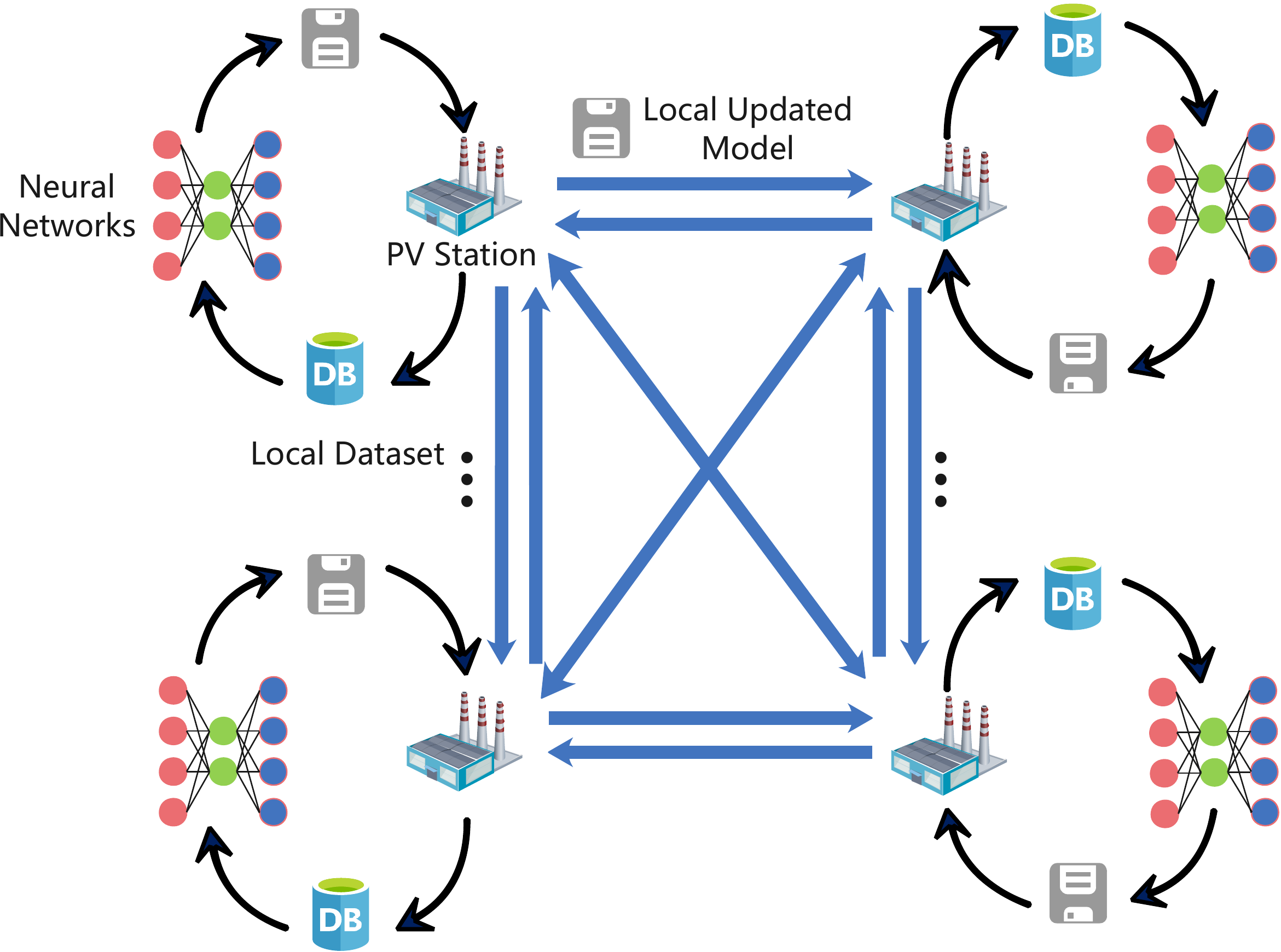}
	\caption{The proposed ADFL framework}
	\label{fig_framework}
\end{figure}

Different from the centralized FL methods, the proposed ADFL framework is fully decentralized and does not require the participation of the central server, as shown in Fig. \ref{fig_framework}.
Each agent first records the I-V characteristic curves and environmental information as the dataset to train the local fault diagnosis model, and then broadcasts the model parameters to other participants while receiving updated models sent by them. More details are described as follows:

\subsubsection{\textbf{Functions of agents}}
Each agent mainly contains four functions: (a) Update the local fault diagnosis model based on its fault samples; (b) Distribute local model parameters to other agents; (c) Receive updated models from other agents and complete model aggregation distributedly; (d) Compare the aggregated global model with the current local model, and reserve the model with better accuracy as the new local model for subsequent communication. Repeat the above steps until the global model converges.
Since all agents in the FL framework use the neural networks with the same structure, the global model is obtained by aggregating the model parameters of multiple PV stations without sharing the original data.

\begin{figure}[htbp]
	\centering
	\includegraphics[width=0.35\textwidth]{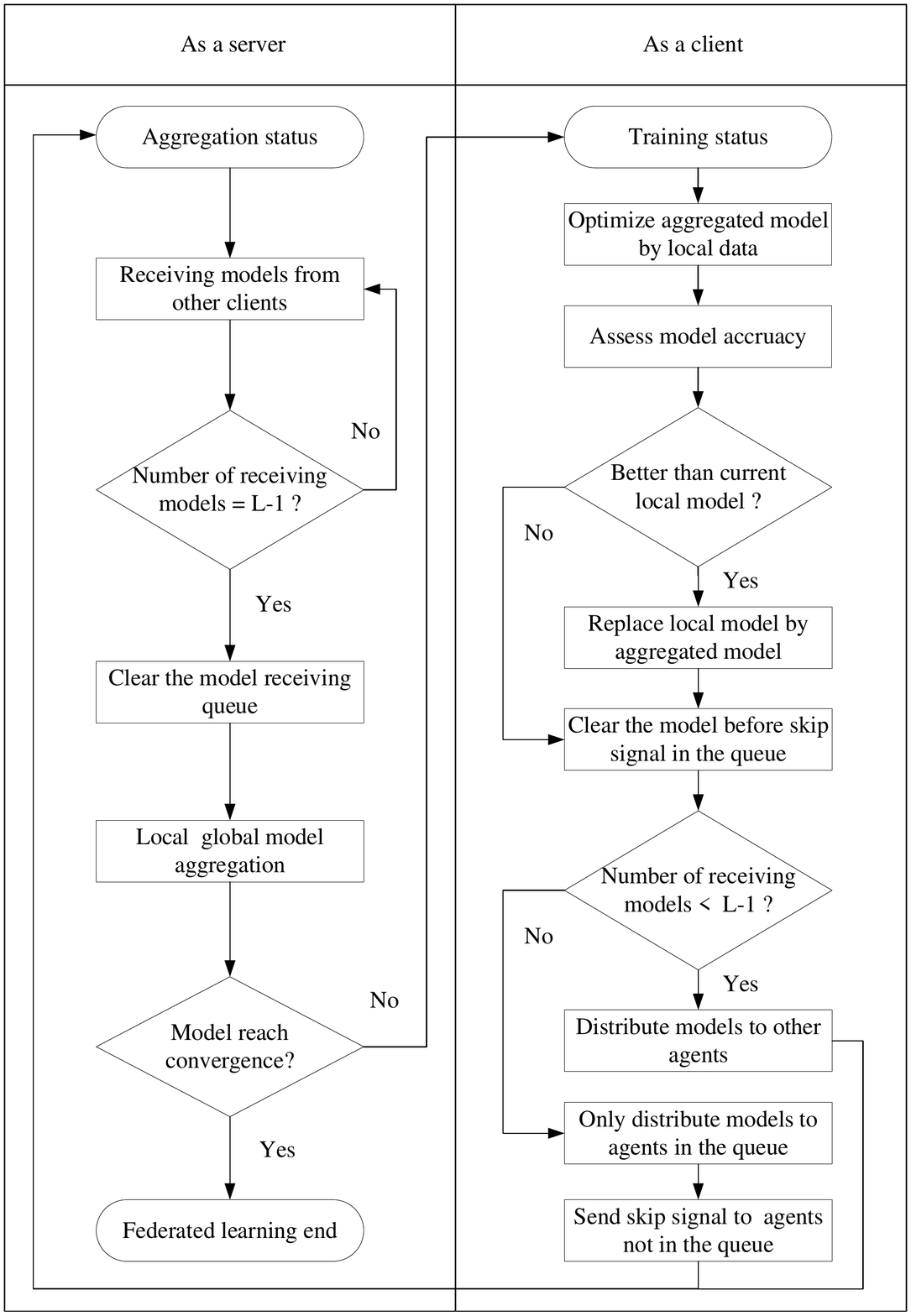}
	\caption{Two roles of each agent}
	\label{fig_CS}
\end{figure}

In the process of collaborative fault diagnosis modeling, each agent is both a client and a server, as shown in Fig. \ref{fig_CS}.
The model receiving termination threshold is introduced as a flag to update the global model.
Each agent checks the current model receiving queue after updating its local model. Once the termination threshold is triggered, it only sends model parameters to the agents in the queue to complete the global model update. Meanwhile, it sends skip signals to the remaining agents and clears the queue, indicating that they skip the current round of model aggregation. The data volume of the skip signal is very small and is not considered to consume communication resources.
Furthermore, an optimal model selection mechanism is executed to compare the latest global model with the current local model. The model with higher fault diagnosis accuracy is retained for the next round of communication. 

\subsubsection{\textbf{Asynchronous update strategy}}
\begin{figure*}[!htbp]
	\centering
	\includegraphics[width=0.7\textwidth]{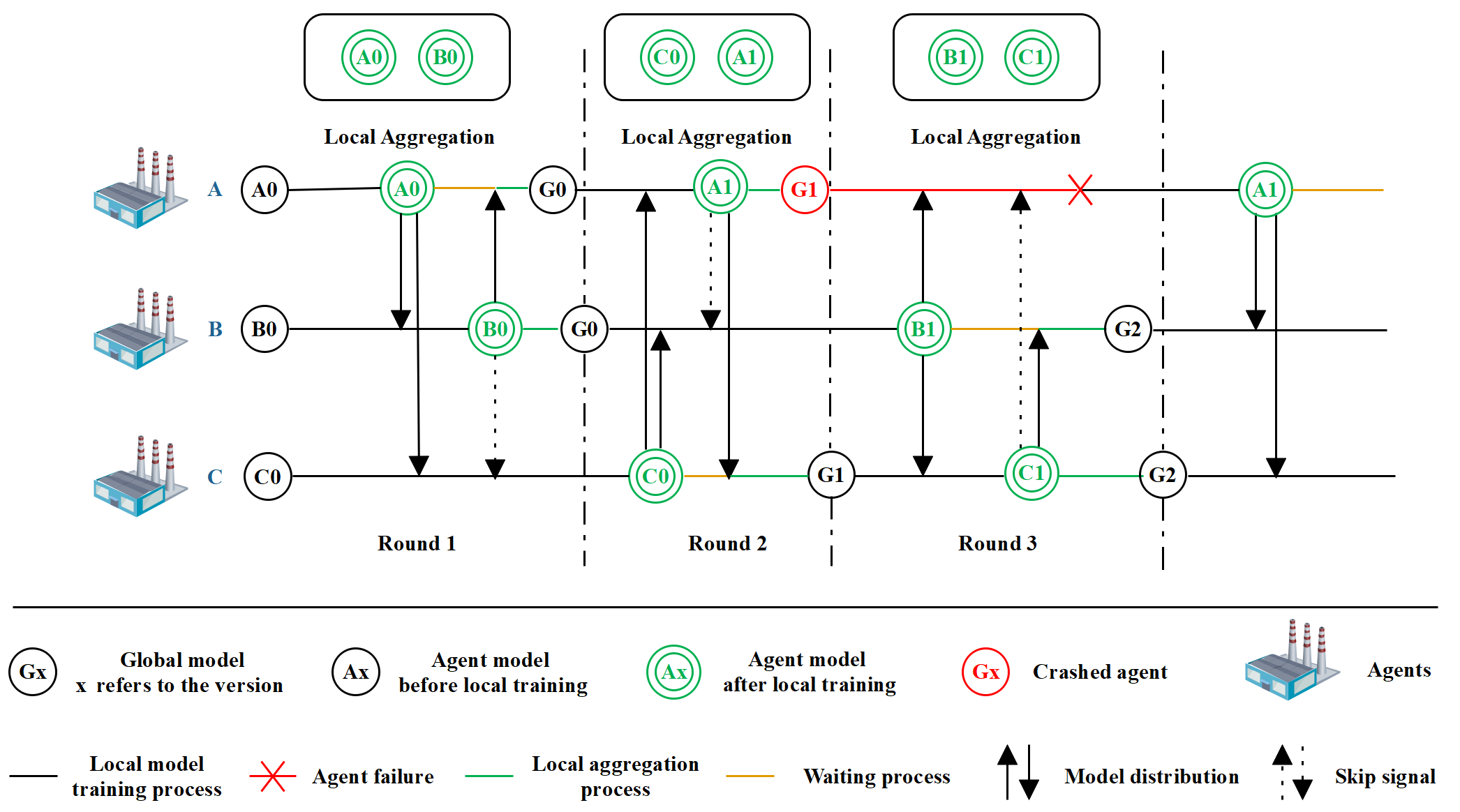}
	\caption{The diagram of asynchronous update strategy}
	\label{FRL_yb}
\end{figure*}

Due to the difference in computing capability and the amount of data, the time required for each agent to update the local model is different. 
The synchronous update scheme leads to low modeling efficiency due to the long waiting time. 
This motivates us to design an asynchronous strategy, where each agent joins the global model aggregation in different rounds, as shown in Fig. \ref{FRL_yb}.

Since the proposed ADFL framework is decentralized, the global model is aggregated without considering the status of other agents when the model parameters collected by any agent reach a certain amount.
In the synchronous update strategy, a certain number of agents are reselected in each round to train their models, and the slowest agent greatly affects the update efficiency \cite{35Wu}.
Although the number of agents participating in each round is certain due to the termination threshold, the proposed asynchronous strategy performs more efficiently.
This is because not all of these agents start to update their local models from the round in which they participate in global model aggregation. 
For agents with low computing capability, it takes a long time to update their local model parameters while the global model may have been updated several times by other agents during this time. 
When the local model update is completed, they will immediately participate in the global model aggregation. 
For example, agent $C$ joins the global update in round 2, but it starts to train the local model since round 1, which effectively reduces the waiting time.
It should be stressed that even if agent $A$ crashes in round 3, agent $B$ and agent $C$ still update the global model.
Compared to the existing centralized FL framework, the proposed ADFL method has several advantages. 
\begin{itemize}
	\item[$\bullet$] \emph{Save computing resources:} It enables the local model updated by each agent to be used in a specific global model aggregation round rather than being discarded directly like the synchronous update mechanism, which avoids wasting local computing resources of each participant.
\end{itemize}
\begin{itemize}
	\item[$\bullet$] \emph{Enhance model diversity:} It ensures that agents with different computing capabilities and amounts of data have the chance to join the model aggregation, which enhances the diversity and generalization of the model.
\end{itemize}
\begin{itemize}
	\item[$\bullet$] \emph{Improve training efficiency:} It speeds up the aggregation frequency of the global model and saves time waiting for other agents to update, thereby improving training efficiency.
\end{itemize}
\begin{itemize}
	\item[$\bullet$] \emph{Guarantee the robustness:} The global model is aggregated distributedly, which avoids the issue that collaborative fault diagnosis cannot be carried out once the central node crashes to a certain extent.
\end{itemize}

\section{ADFL for Collaborative Fault Diagnosis}
In this section, we specifically introduce the implementation of the collaborative fault diagnosis based on the proposed ADFL framework.
It is mainly composed of two parts, including each agent separately training the local fault diagnosis model and multiple agents jointly aggregating a shared global model.

\subsection{Data Preprocessing}
\begin{figure}[htbp]
	\centering
	\includegraphics[width=2.5in]{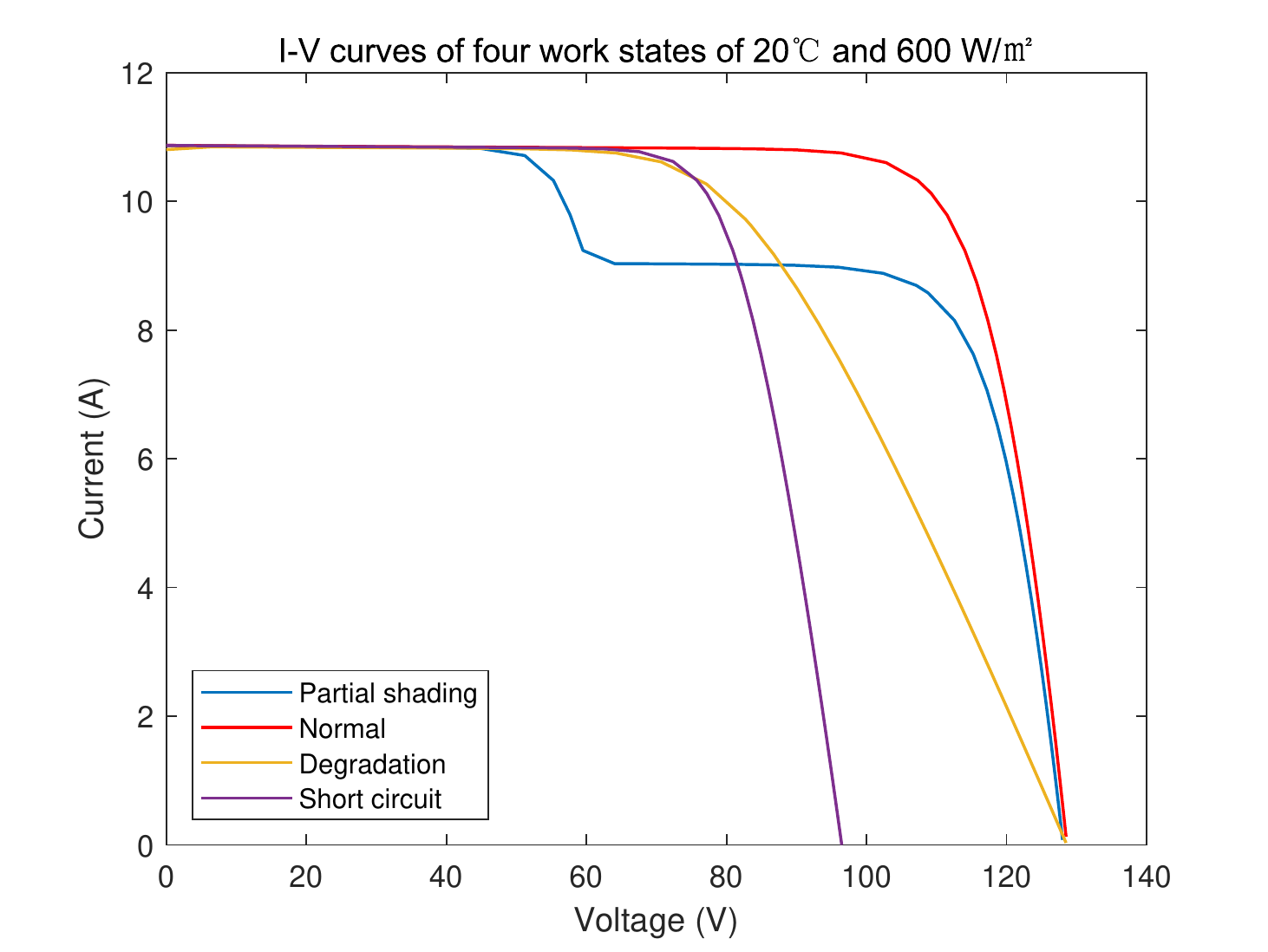}
	\caption{I-V characteristic curves under different work status}
	\label{fig_IV}
\end{figure}

The I-V characteristic curves of PV arrays under different operating status are quite different \cite{29Rosario}. 
Specifically, the I-V characteristic curves of the normal and three fault states under the temperature of 20$^{\circ}$C and the irradiance of 600$W/m^{2}$ are shown in Fig. \ref{fig_IV}.

The original I-V curves are not directly used for fault diagnosis for two reasons. 
First, there are many sampling points for the collected I-V characteristic curves, which contain redundant information and occupy additional computing resources.
Second, the sampling points on the curves are unevenly distributed, which makes it difficult to accurately reflect the fault features. 
To solve these problems, the original I-V characteristic curves are compressed and enhanced by combining the down-sampling and bilinear interpolation methods.
To retain as much effective feature information as possible, 20 new data points $V_{\rm sample\_i}$ and $I_{\rm sample\_i}$ are sampled equidistantly in the range of $[0, V_{\rm oc}]$ and $[0, I_{\rm sc}]$ respectively, $i=1, 2, \ldots, 20$.
The corresponding current $I_{\rm n}$ and voltage values $V_{\rm n}$, $n=1, 2, \ldots, 20$, are calculated from the original data points according to bilinear interpolation.
Then, they are sorted in ascending order of voltage.  
Specifically, the voltage and current of the 40 new sampling points are respectively calculated by (\ref{Current}) and (\ref{Voltage}): 
\begin{equation}
I_{\rm n}=\frac{\left(V_{\rm sample\_i}-V_{\rm left}\right) \cdot I_{\rm right}+\left(V_{\rm right}-V_{\rm sample\_i}\right) \cdot I_{\rm left}}{V_{\rm right}-V_{\rm left}},
\label{Current}
\end{equation}
\begin{equation}
V_{\rm n}=\frac{\left(I_{\rm sample\_i}-I_{\rm left}\right) \cdot V_{\rm right}+\left(I_{\rm right}-I_{\rm sample\_i}\right) \cdot V_{\rm left}}{I_{\rm right}-I_{\rm left}},
\label{Voltage}
\end{equation}
where $V_{\rm left}$, $I_{\rm left}$ and $V_{\rm right}$, $I_{\rm right}$ respectively represent the current and voltage values of the left and right samples closest to $V_{\rm sample\_i}$ and $I_{\rm sample\_i}$.

\begin{figure}[htbp] 
	\centering 
	\subfigure[Before downsampling]{ 
		\includegraphics[width=0.2\textwidth]{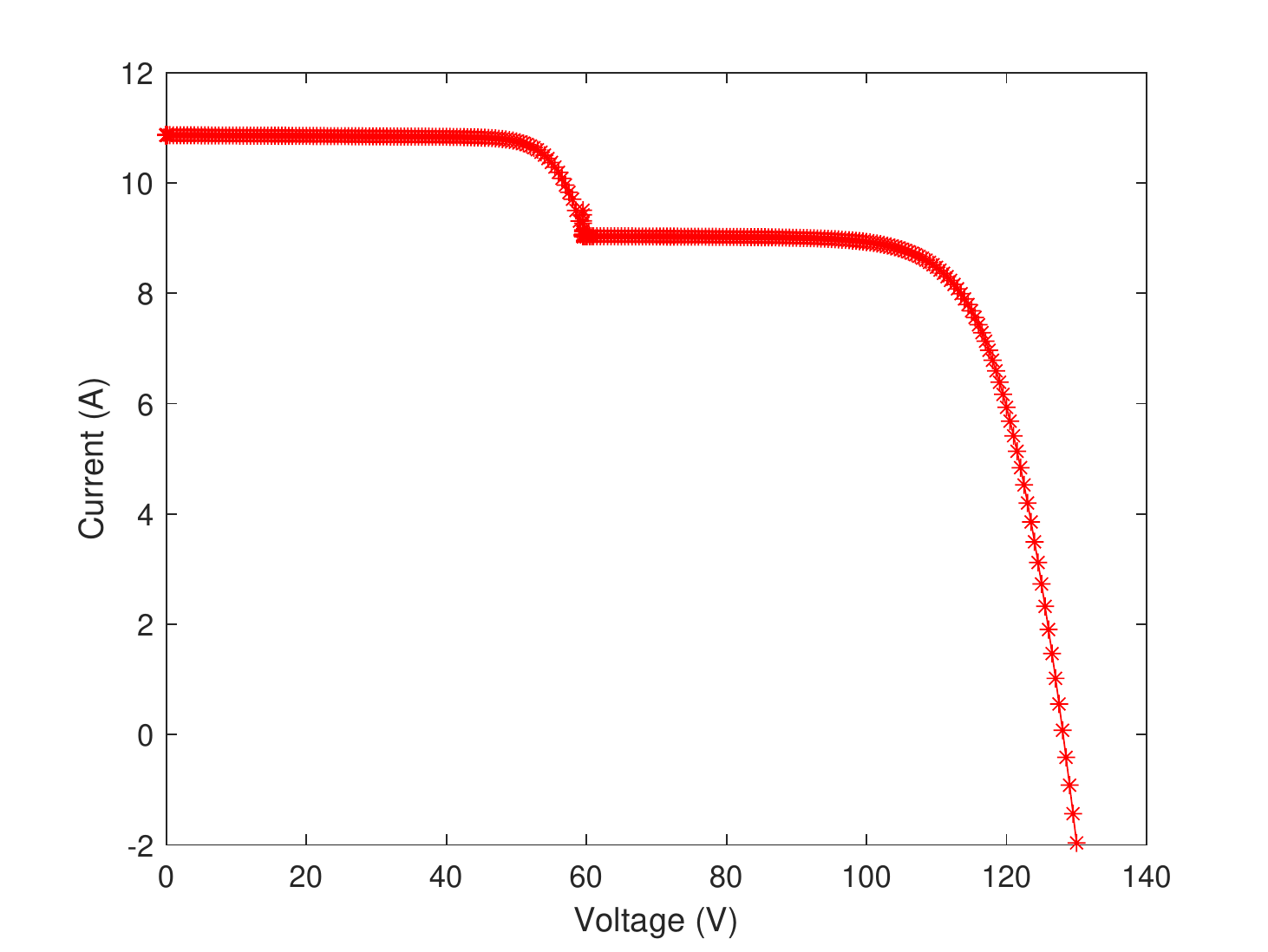} 
	} 
	\subfigure[After downsampling]{ 
		\includegraphics[width=0.2\textwidth]{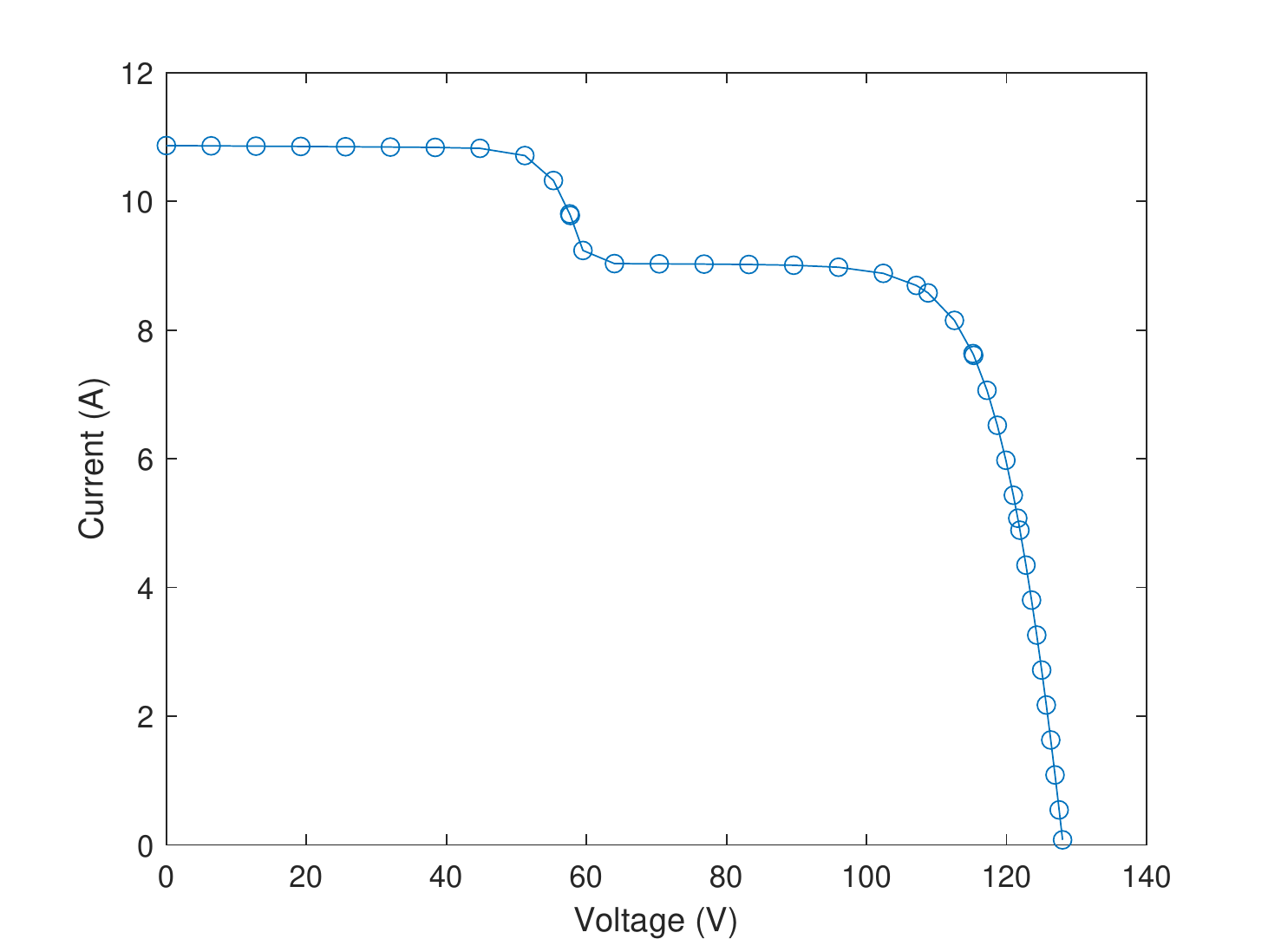} 
	} 
	\caption{I-V characteristic curve of partial shading fault} 
	\label{fig_downsampling}
\end{figure}

Specifically, the I-V characteristic curves of partial shading faults before and after data preprocessing under the temperature of 20$^{\circ}$C and the irradiance of 600$W/m^{2}$ are shown in Fig. \ref{fig_downsampling}.
Through the above data preprocessing method, the original curve is compressed from approximately 400 data points to 40, which are evenly distributed on the curve. 
The current and voltage of the new I-V curve are expressed as a 40*2 I-V vector. 
Meanwhile, considering the influence of temperature and irradiance, we add them to the I-V vector as a 40*2 environmental feature vector. Then the reconstructed 40*4 two-dimensional array is regarded as a fault sample and used as the input of the later CNN.

\subsection{CNN Unit}
In the proposed ADFL framework, the global model is obtained by exchanging and aggregating local model parameters between agents.
Therefore, a neural network with the same structure is required to train a local fault diagnosis model for each agent separately.  
Considering that the collected samples are 40*4 feature matrices, a plain CNN proposed in \cite{23ZhicongChen3} is selected to automatically extract the two-dimensional data features and train the local fault diagnosis model.
The structure of the CNN includes a 2-D CNN layer, 1-D CNN layers, Maxpool layers, a fully connected (FC) layer, and the Softmax function. The architecture of the network is shown in Table. \ref{table_CNN}. The following subsection specifically introduces the process of collaborative fault diagnosis modeling for multiple PV stations.
\begin{table}[htbp]
	\setlength\tabcolsep{2pt}
	\caption{Configuration for the plain CNN}
	\label{table_CNN}
	\centering
	\begin{tabular}{c c c}
		\hline
		\textbf{Layer} & \textbf{Output demension} & \textbf{Detailed architecture of CNN} \\
		\hline
		2-D CNN & 37×1×1 & $k=4×4, C_{out}=1,ss=1,p=0$\\ 
		Squeeze & 37×1 & Demension squeeze\\
		1-D CNN & 18×3 & $k=3, C_{out}=3,ss=2,p=0$\\ 
		1-D CNN & 18×5 & $k=3, C_{out}=5,ss=1,p=1$\\ 
		MaxPool 1D & 8×5 & $k=4, ss=2$\\ 
		1-D CNN & 8×8 & $k=3, C_{out}=8,ss=1,p=1$\\ 
		MaxPool 1D & 4×8 & $k=2, ss=2$\\ 
		1-D CNN & 4×16 & $k=3, C_{out}=16,ss=1,p=1$\\ 
		MaxPool 1D & 1×16 & $k=4, ss=1$\\
		FC & 10 & Fully-connected Layer\\ 
		Linear Classifier & 10×9 & FC + Softmax\\ 
		\hline
	\end{tabular}
\end{table}

\subsection{Collaborative Fault Diagnosis Modeling}
The goal of collaborative fault diagnosis is to enable multiple PV stations to identify more fault types by making full use of their local data while ensuring data privacy. 
Therefore, an asynchronous decentralized aggregation algorithm is proposed to obtain a high-precision collaborative (global) model by integrating their local model parameters.
Specifically, based on the local data set $D_{k}$ and the current local model $w_{t}^{k}$, agent $C_{k}$ obtains the updated local model $w_{t+1}^{k}$ by running the adaptive moment estimation (Adam) \cite{24adam} method on a mini-batch of $B<D_{k}$, and thus we have:
\begin{equation}
f\left(x^{(i)} ; w\right) = \hat{y}_{i},
\end{equation}
\begin{equation}
J(w)=\frac{1}{B} \sum_{i=1}^{B}\left[y_{i} \log \hat{y}_{i}+\left(1-y_{i}\right) \log \left(1-\hat{y}_{i}\right)\right],
\end{equation}
\begin{equation}
g_{t}=\frac{1}{B} \nabla_{w} \sum_{i=1}^{B} J\left(\hat{y}_{i}, y^{(i)}\right),
\end{equation}
\begin{equation}
m_{t}=\beta_{1} \cdot m_{t-1}+\left(1-\beta_{1}\right) \cdot g_{t}, 
\end{equation}
\begin{equation}
v_{t}=\beta_{2} \cdot v_{t-1}+\left(1-\beta_{2}\right) \cdot g_{t}^{2},
\end{equation}
\begin{equation}
\hat{m}_{t}=\frac{m_{t}}{1-\beta_{1} { t }} , \hat{v}_{t}=\frac{v_{t}}{1-\beta_{2} {t}}, 
\end{equation}
\begin{equation}
w_{t+1}^{k}=w_{t}^{k}- {l_{r}} \frac{\hat{m}_{t}}{\sqrt{v_{t}}+\epsilon}, 
\end{equation}
where $x^{(i)}$is a sample in a small batch $B$ and the model parameters of the CNN are represented by $w$. $\hat{y}_{i}$ is the probability vector of the CNN softmax function and ${y}_{i}$ is the real label of the sample. $B$ is the mini-batch size and $g$ represents the gradient of the loss function $J$. The initial values of $m_{t}$ and $v_{t}$ are equal to 0 when $t=0$. $\beta_{1}$ and $\beta_{2}$ are the average moving coefficients. $l_{r}$ is the initial learning rate and $\epsilon$ is a small constant to avoid a zero denominator. In the following, we take agent $C_{k}$ as an example to illustrate.

During each round of communication, agent $C_{k}$ sends its local model parameters $w_{t}^{k}$ to other agents, and receives weights $w_{t}^{i}$ from others. Based on the collected parameters, each agent aggregates global models distributedly. 
Once receiving the model parameters of other participants, it immediately updates the global model without waiting for other agents to complete the update.
To further improve the communication efficiency, we set a termination threshold $L$ for asynchronous updates. 
$L$ is a positive integer between 1 and $C$, and $C$ is the total number of agents.
By properly modifying the value of the termination threshold $L$ based on specific problems, the ideas in this paper can be applied to the collaborative training of various numbers of agents.
$\mathcal{L}$ is a set containing the agents whose model parameters have been received. After an agent receives model weights from other $L-1$ participants, it starts to aggregate global model $G_{t}$. 
Thus we have:
\begin{equation}
G_{t}=\sum_{i \in \mathcal{L}} \alpha_{i} w_{t}^{i}+\sum_{j \notin \mathcal{L}} \beta_{j} w_{t-1}^{j},
\label{sum}
\end{equation}
where $\alpha_{i}$ and $\beta_{j}$ are the mixing weights for the models. The global model of consists of two parts. For the first $L$ participants (including itself), the agent obtains their latest local model weights $w_{t}^{i}$, and for the remaining participants, the expired local weights $w_{t-1}^{j}$ are used instead. 

The selection of mixing weights is significant to the effectiveness of model aggregation. It is worth noting that the training data of each participating agent is changing in each round. Therefore, data size is significant to model aggregation. Inspired by \cite{25LiuYi}, the following mixing weights method is chosen:
\begin{equation}
\alpha_{i} = \frac{d_{i}}{D} , \beta_{j} = \frac{d_{j}}{D},
\label{pra}
\end{equation}
where $d_{i}=\left|D_{i}\right|$ and $d_{j}=\left|D_{j}\right|$ denote the local data size of agent $C_{i}$ and agent $C_{j}$, respectively. The data size of all agents is represented as $D=\sum\left|D_{i}\right|$. 

To speed up the convergence, the optimal model selection mechanism is also designed. When each agent participates in global model aggregation for the first time, the aggregated model $G_{t}$ is directly regarded as the new local model $w_{t+1}^{k}$. Otherwise, agent $C_{k}$ compares the accuracy of the latest global model $G_{t}$ and the current local model $w_{t}^{k}$. The model with higher accuracy is selected as the local fault diagnosis model $w_{t+1}^{k}$ for subsequent rounds of global model aggregation. The detailed steps are shown in Algorithm 1.

\begin{algorithm}[htbp] 
	\caption{Asynchronous Decentralized Model Aggregation} 
	\label{alg:Framwork} 
	\begin{algorithmic}[1] 
		\REQUIRE ~~\\ 
		Local datesets $D_{k}$;\\
		Local model $w_{t}^{k}$;\\
		Model aggregation round $t$;\\
		Mini-batch size $B$\\
		Latest local weights from $L-1$ agents $w_{t+1}^{i}$;\\
		Expired local weights from other agents $w_{t}^{j}$;\\
		Initialization CNN parameters $m_{0}$, $v_{0}$, $l_{r}$, $\epsilon$, $\beta_{1}$ and $\beta_{2}$\\
		\ENSURE ~~\\ 
		Best local fault diagnosis model $w_{t+1}^{k}$  
		\STATE $\textbf{Local Model Learning:}$ For agent $C_{k}$, parameters update is performed to train the local fault diagnosis model as $w_{t+1}^{k}=w_{t}^{k}- {l_{r}} \frac{\hat{m}_{t}}{\sqrt{v_{t}}+\epsilon}$ by running Adam method over a number of mini-batches.
		\label{ code:fram:extract }
		\STATE  $\textbf{Model Broadcasting:}$ Agent $C_{k}$ broadcasts its local model $w_{t}^{k}$ to other participants; 
		\label{code:fram:trainbase}
		\STATE $\textbf{Asynchronous Decentralized Model Aggregation:}$
		Once agent $C_{k}$ receives the model parameters of other participant $C_{i}$, it immediately calculates the global model components $\alpha_{i} w_{t}^{i}$. After receiving the model parameters of $L-1$ other agents, the global model is aggregated as $G_{t}=\sum_{i \in \mathcal{L}} \alpha_{i} w_{t}^{i}+\sum_{j \notin \mathcal{L}} \beta_{j} w_{t-1}^{j}$; 
		\label{code:fram:add}
		\STATE $\textbf{Local Model Optimization:}$
		For the first time the agent $C_{k}$ participates in the global model aggregation, $w_{t+1}^{k} = G_{t}$.
		Otherwise, agent $C_{k}$ compares the fault diagnosis accuracy of the aggregated global model $G_{t}$ and the current local model $w_{t}^{k}$. Then, the model with higher accuracy is selected as new local model $w_{t+1}^{k}$; 
		\label{code:fram:classify}
		\RETURN $w_{t+1}^{k}$; 
	\end{algorithmic}
\end{algorithm}

\subsection{Convergence Analysis}
The convergence of the global model of FL has been proved by some studies. However, since we design a new decentralized FL framework and adopt an asynchronous model aggregation mechanism, it is necessary to analyze the convergence of Algorithm 1.

In the proposed ADFL method, the following optimization problems need to be solved in each round of global model update:
\begin{equation}
\arg \min _{G} F(G, t)= \sum_{i \in  \mathcal{L}} \frac{d_{i}}{D} F_{i}(G)+\sum_{j \notin \mathcal{L}} \frac{d_{j}}{D} F_{j}(G), 
\label{opt}
\end{equation}
where $G$ represents the global model parameters, and $F_{k}(G)$ is the average loss of the local data on agent $C_{k}$. $F_{k}(G)$ is calculated by (\ref{li}):
\begin{equation}
F_{k}(G)=\frac{1}{d_{k}} \sum_{\left(x_{k}, y_{k}\right) \in D_{k}} f\left(G ; x_{k}, y_{k}\right),
\label{li}
\end{equation}
where $f(\cdot)$ denotes the loss function, and $d_{k}=\left|D_{k}\right|$ represents the local data size of agent $C_{k}$.

We prove the convergence of the proposed ADFL method by analyzing the upper bound of $F(G(t), t) - F (G^{*}; t)$, where $G^{*}$ denotes the optimal parameters of the global model. 
We mainly refer to the idea of convergence analysis in \cite{36Wang} and \cite{37Wu}, and make appropriate changes based on our situation. 
First, we also make the following assumption:

\noindent\textbf{\emph{Assumption 1.}} For each agent $C_{k}$, the loss function $F_{k}(w)$ is convex, $\rho$-Lipschitz and $H$-smooth. For any ${w}$, ${w^{\prime}}$, we have:\\
1)1)$F\left(w\right)-F\left(w^{\prime}\right) \leq \nabla F\left(w\right)^{\mathrm{T}}\left(w-w^{\prime}\right)$, \\
2)$\left\|F_{k}({w})-F_{k}\left({w}^{\prime}\right)\right\| \leq \rho \| {w}-{w}^{\prime} \mid$, \\
3)$\left\|\nabla F_{k}({w})-\nabla F_{k}\left({w}^{\prime}\right)\right\| \leq H\left\|{w}-{w}^{\prime}\right\|$.

Based on Assumption 1 and triangle inequality, we further introduce the definition of gradient divergence in \cite{36Wang}, to quantify the gradient difference between the local and global loss function.
As defined in \cite{36Wang}, the $\delta_{k}$ is the upper bound of the gradient divergence of agent $C_{k}$, and $\bar{\delta}$ denotes the upper limit of $\delta_{k}$:
\begin{equation}
\left\|\nabla F_{k}(w)-\nabla F(w)\right\| \leq \delta_{k} \leq \bar{\delta}.
\label{com}
\end{equation}

Considering that each agent trains the local model $\tau$ epochs before sharing parameters, we use $e$ to represent the index of the epoch, $e = 1, 2, \ldots, \tau$. Formally, we define a virtual global model $G_{t}([e])$, which is obtained by the aggregation of all $w_{t}^{k}([e])$ at epoch $e$.
Inspired by \cite{37Wu}, we introduce an auxiliary model $z_{t}([e])$, which is initialized as $G_{t - 1}$ in round $t$ and updated by centralized gradient descent for optimizing the same target $F(G, t)$. Based on the definition, we have: 
\begin{equation}
G_{t}([e])=\sum_{i \in  \mathcal{L}} \frac{d_{i}}{D} w_{t}^{i}([e]) + \sum_{j \notin \mathcal{L}} \frac{d_{j}}{D} w_{t}^{j}([e]),
\label{vg}
\end{equation}
\begin{equation}
w_{t}^{i}([e])=w_{t}^{i}([e-1])-l_{r} \nabla F_{i}\left(w_{t}^{i}([e-1])\right),
\label{vg1}
\end{equation}
\begin{equation}
w_{t}^{j}([e])= w_{t}^{j}([e-1]),
\label{vg2}
\end{equation}
\begin{equation}
z_{t}([e])=z_{t}([e-1])-l_{r} \nabla F\left(z_{t}([e-1])\right),
\label{vg3}
\end{equation}
where $w_{t}^{i}([e])$ is updated from $w_{t}^{i}([e-1])$ and $w_{t}^{j}([e])$ does not change with epoch $e$ because agent $j$ does not participate in the $t$-th round of model aggregation.

\noindent\textbf{\emph{Theorem 1:}} For any epoch e in round t, we have the loss divergence bound:
\begin{equation}
F(G_{t}([e]))-F\left(z_{t}([e])\right) \leq \rho \bar{u}(e),
\end{equation}
where 
\begin{equation}
\bar{u}(x) \triangleq \frac{\bar{\delta}}{H}\left((l_{r} H+1)^{x}-1\right)-l_{r} \bar{\delta} x.
\end{equation}

\noindent\textbf{\emph{Proof.}} The proof of Theorem 1 is mainly based on Lemma 2 in \cite{36Wang} and the details are in Appendix A.

Theorem 1 quantifies the difference in loss between the $G_{t}([e])$ aggregated by the local models and the $z_{t}([e]$ learned from centralized training in each epoch $[e]$.
It is worth noting that $G_{t}([e])=z_{t}([e]) \leq \rho \bar{u}(\tau)$ at $e = \tau$ and $\tau = 1$ since $\bar{u}(1) = 0$. Considering that $G_{t}=G_{t}([t \cdot\tau])$, the convergence upper bound of $G_{t}$ is further analyzed in Theorem 2.

\noindent\textbf{\emph{Theorem 2:}} The convergence upper bound of the global model in round $t$ with $\tau$ epochs is given by:
\begin{equation}
F(G_{t})-F\left(G^{*}\right) \leq \frac{1}{T\left(\chi \varphi-\zeta\right)},
\end{equation}
when it satisfies the following conditions: \\
1) $l_{r} \leq \frac{1}{H}$, \\
2) $\chi \varphi-\zeta>0$, \\
3) $F\left(z_{t}([e])\right)-F\left(G^{*}\right) \geq \varepsilon, \forall e$, \\
4) $F(G_{t})-F\left(G^{*}\right) \geq \varepsilon$, \\
where $\varepsilon > 0$, $T = t\tau$, $\chi \triangleq \min _{t} \frac{1}{\left\|z_{t}([0])-G^{*}\right\|^{2}}$, $\varphi \triangleq l_{r}(1-\frac{H l_{r}}{2})$, and $\zeta\triangleq \frac{\rho \bar{u}(\tau)}{\tau \varepsilon^{2}}$.

\noindent\textbf{\emph{Proof.}} 
Condition 1) indicates that the learning rate $l_{r}$ cannot be too large. Condition 2) restricts the auxiliary model $z_{t}([e])$ from deviating too much from the virtual global model $G_{t}([e])$. Conditions 3) and 4) ensure that the loss of the approximate model is not less than the theoretical optimal model $G^{*}$ \cite{37Wu}.
By combining the result of Theorem 1 and the Lemmas in \cite{36Wang} and \cite{38Bubeck}, Theorem 2 is proved and see Appendix B for details.

\section{Experiments and Simulation Studies}
In this section, experiments and numerical simulations are carried out to verify the effectiveness of the proposed ADFL method. First, experiments based on the temperature and irradiance data in a small range are carried out to illustrate the basic properties of the method. Then, numerical simulation is investigated to verify that the method is suitable for various meteorological conditions.

\subsection{Experiments Configuration}
In the experiments, we consider the collaborative fault diagnosis of three agents. Each agent has a certain number of local fault samples, which are divided into the training set and test set at a ratio of 7 : 3. It is worth noting that each fault type is divided according to this ratio instead of randomly shuffled samples. This ensures that all the local fault types are covered in the training set, which can maximize the accuracy and generalization performance of the local model.
The model receiving termination threshold $L$ is set to two. Some key hyperparameter settings are shown in Table \ref{table_hyper}.
The PV module installed in the actual PV array is HT60-156M-C-330, and the detailed parameters are shown in Table \ref{module_parameter}.
\begin{table}[!h]
	\caption{Settings of the hyper-parameters}
	\label{table_hyper}
	\centering
	\begin{tabular}{c c c c}
		\hline
		\textbf{Hyper-parameters} & \textbf{Value} & \textbf{Hyper-parameters} & \textbf{Value}\\
		\hline
		$B$ & 128 & $l_{r}$ & 1e-3\\
		$\beta_{1}$ & 0.995 & $\beta_{2}$ & 0.999\\
		Epoch & 50 & $\epsilon$ & 1e-8\\
		\hline
	\end{tabular}
\end{table}
\begin{table}[!htbp]
	\caption{Technial data at standard test condition}
	\label{module_parameter}
	\centering
	\begin{tabular}{c c}
		\hline
		\textbf{Module parameters} & \textbf{Value} \\
		\hline
		Maximum power point $M_{\rm pp}$ (W) & 99.925\\
		Open circuit voltage $V_{\rm oc}$ (V) & 21.5\\
		Short-circuit current $I_{\rm sc}$ (A) & 6.03\\
		Maximum power voltage $V_{\rm mp}$ (V) & 17.5\\
		Maximum power curren $I_{\rm mp}$ (A) & 5.71\\
		Fuse current (A) & 15\\
		Maximum system voltage (V) & 1000\\
		\hline
	\end{tabular}
\end{table}

\begin{figure} [!t]
	\centering 
	\subfigure[I-V tester PROVA1011]{ 
		\includegraphics[width=0.15\textwidth]{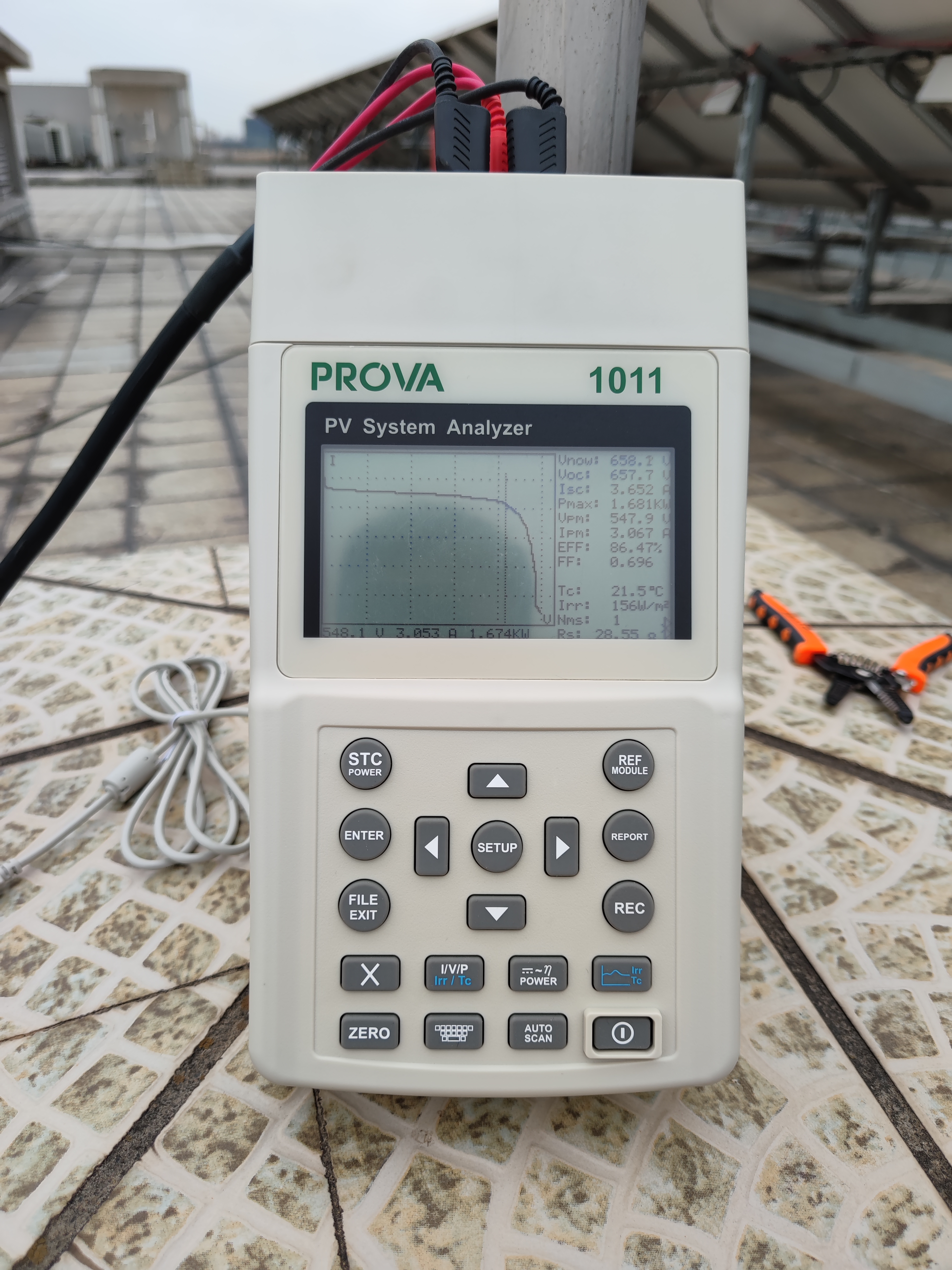} 
	} 
	\subfigure[Environmental tester]{ 
		\includegraphics[width=0.15\textwidth]{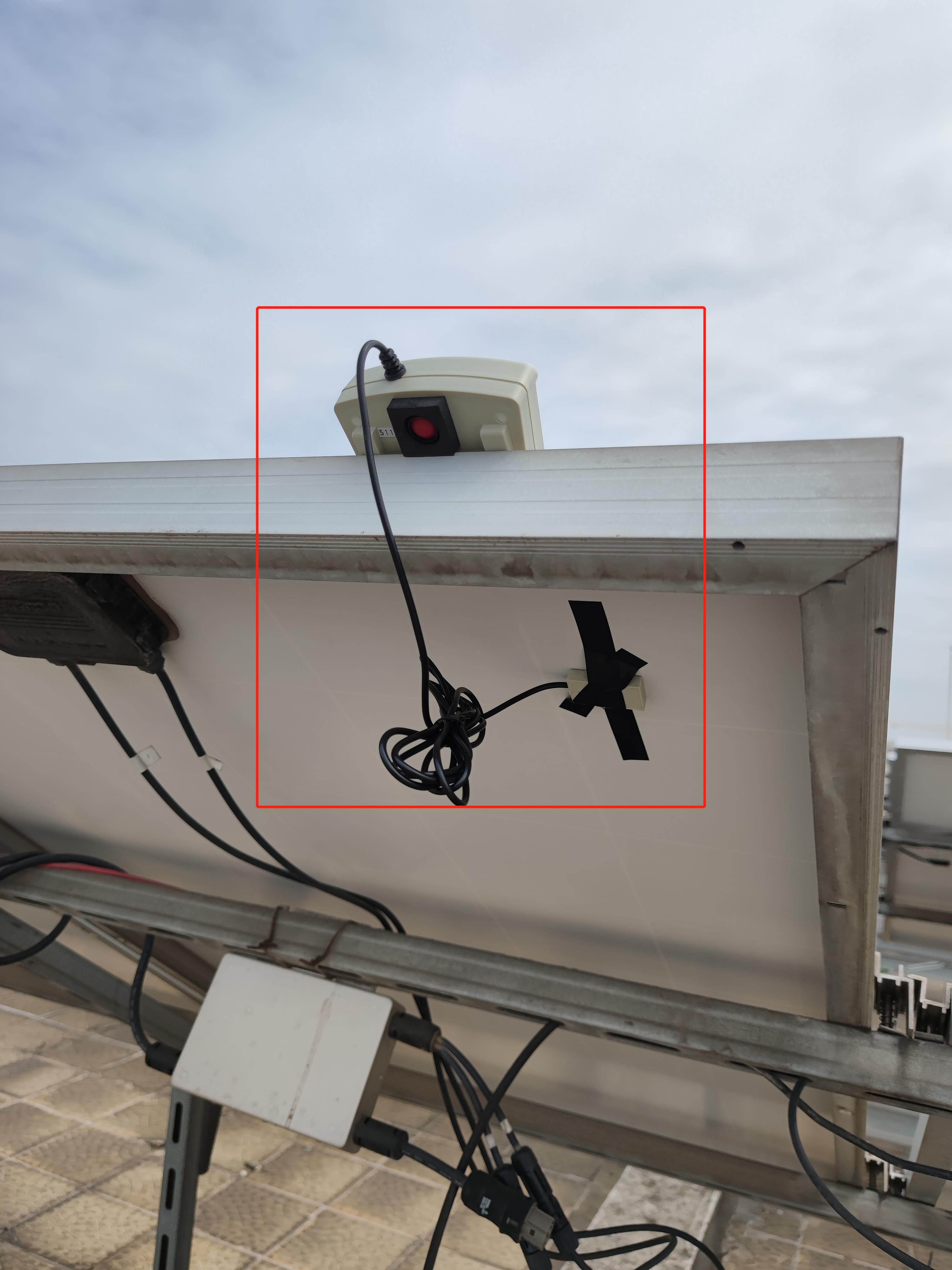} 
	} 	
	\caption{The I-V tester PROVA1011 and environmental tester} 
	\label{fig_IV tester}
\end{figure}

\begin{figure} [!t]
	\centering 
	\subfigure[Short-circuit]{ 
		\includegraphics[width=0.14\textwidth]{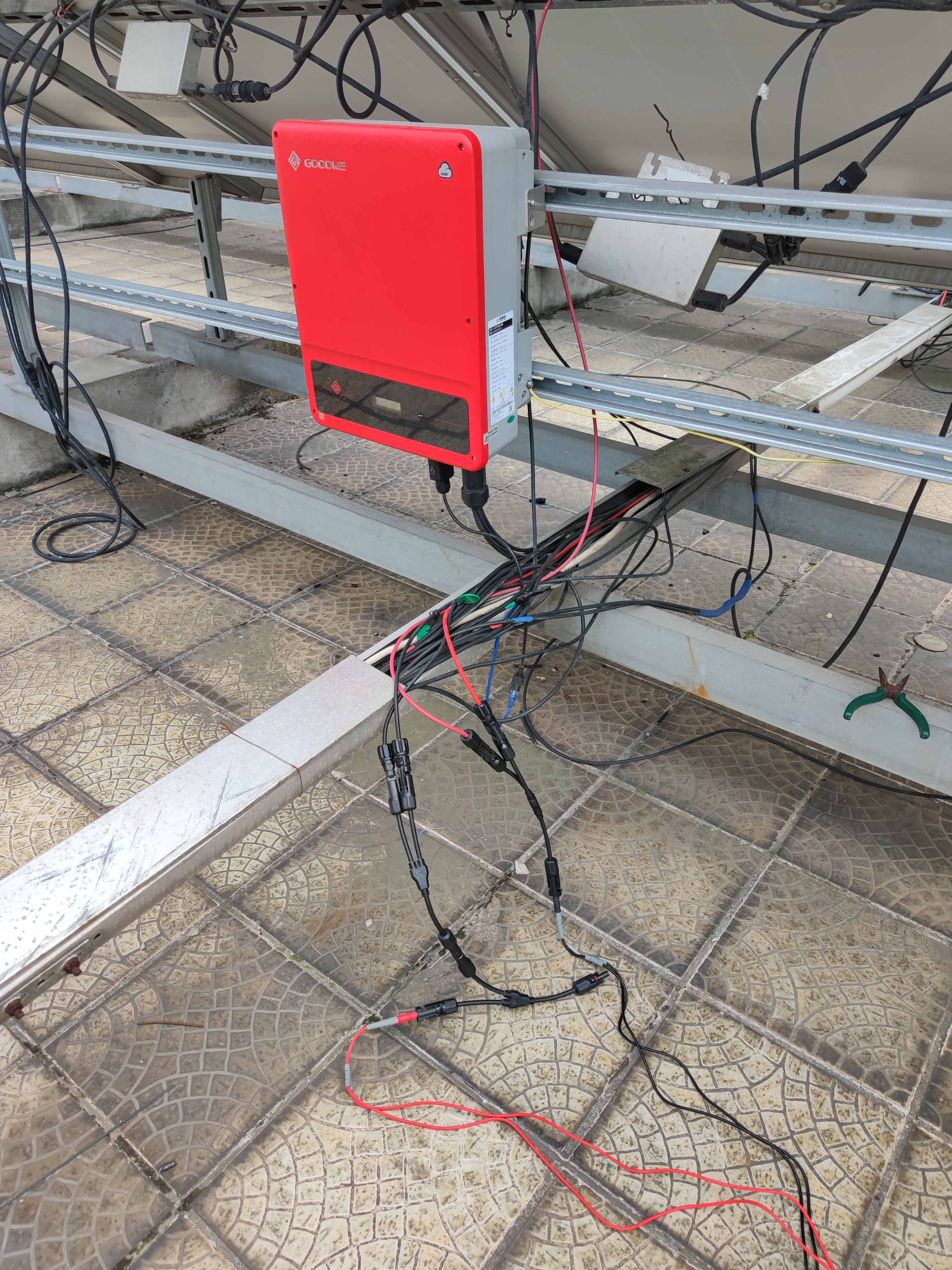} 
	} 
	\subfigure[Partial shading]{ 
		\includegraphics[width=0.14\textwidth]{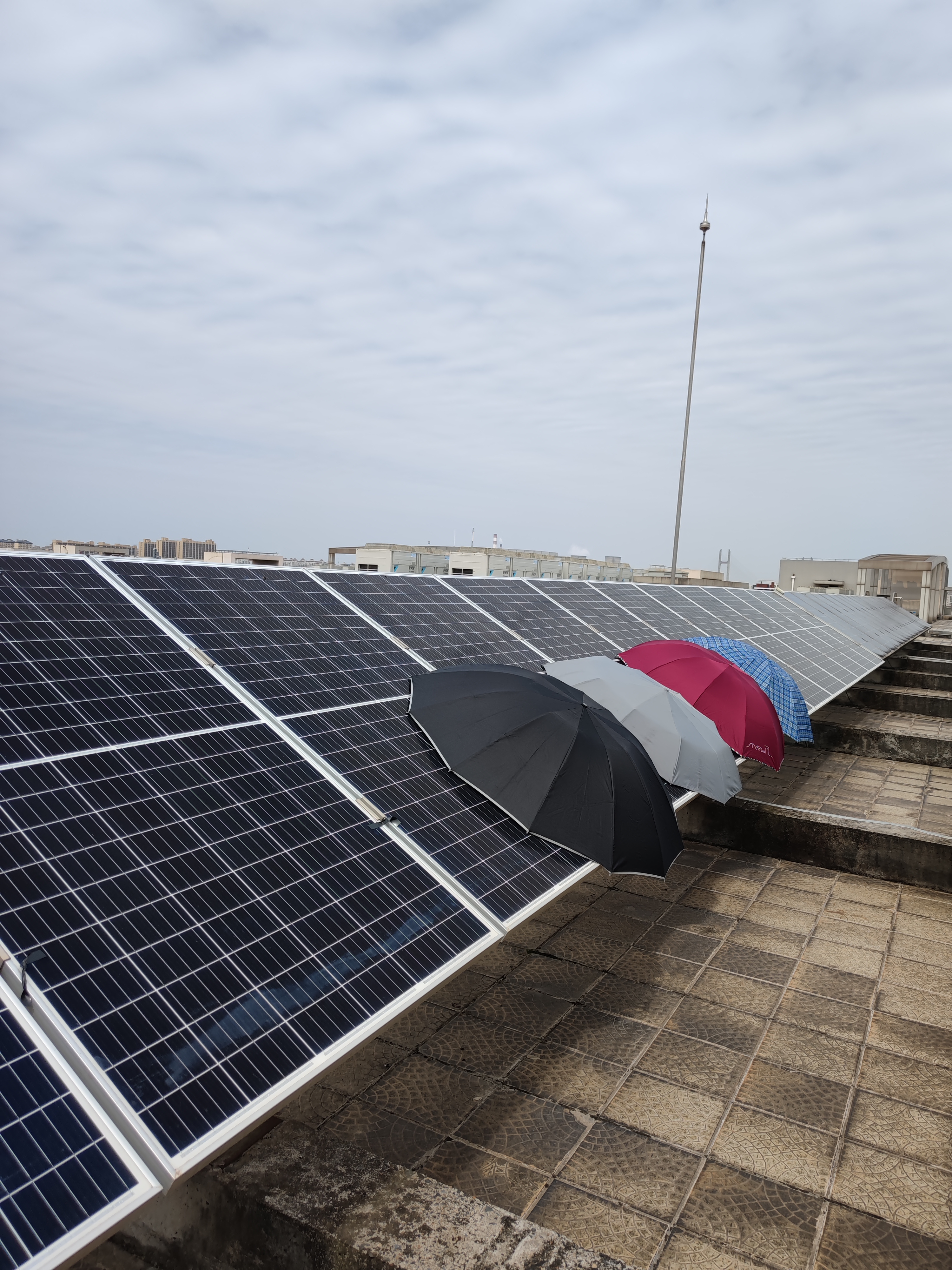} 
	} 
	\subfigure[Degradation]{ 
		\includegraphics[width=0.14\textwidth]{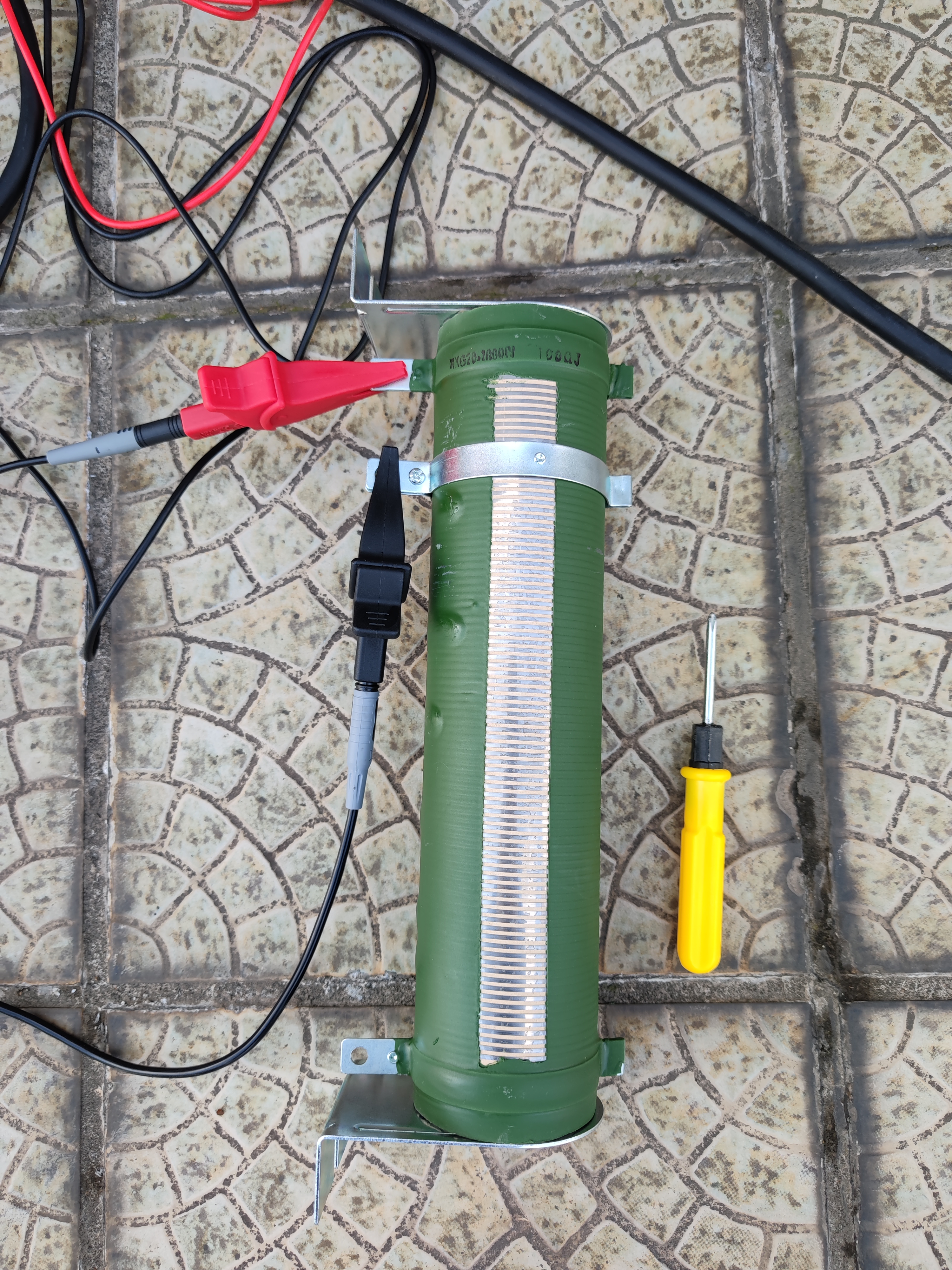} 
	} 	
	\caption{Fault simulation in actual PV array} 
	\label{fig_real experiments}
\end{figure}

\begin{table*}[!t]
	\renewcommand{\arraystretch}{1.3}
	\caption{Dataset configuration for each agent in six real experiments}
	\label{table_real experiment}
	\centering
	\resizebox{\textwidth}{!}{
		\begin{tabular}{c c c c c c}
			\hline
			\diagbox{\textbf{Participants}}{\textbf{Sample numbers of each experiment }}{\textbf{Types}} & \textbf{Normal} & \textbf{Short-current} & \textbf{Degradation} & \textbf{Partial shading}\\
			\hline
			Agent 1 & 100/ 100/ 100/ 100/ 100/ 100 & 100/ 100/ 100/ 100/ 100/ 100 & 0/ 100/ 0/ 100/ 100/ 100 & 0/ 0/ 100/ 100/ 0/ 100\\
			Agent 2 & 100/ 100/ 100/ 100/ 100/ 100 &  0/ 0/ 0/ 0/ 0/ 100 & 100/ 100/ 100/ 100/ 100/ 100 & 0/ 0/ 0/ 0/ 100/ 100\\ 
			Agent 3 & 100/ 100/ 100/ 100/ 100/ 100 &  0/ 0/ 0/ 0/ 100/ 100 & 0/ 0/ 0/ 0/ 0/ 100 & 100/ 100/ 100/ 100/ 100/ 100\\	
			\hline
	\end{tabular}}
\end{table*}
The actual PV array consists of two parallel-connected PV strings with twenty-two PV modules in series. 
The I-V tester (Model: PROVA1011) equipped with an environmental tester with Bluetooth communication function is selected to collect data in various situations, as shown in Fig. \ref{fig_IV tester}.
Based on the actual PV array, short-circuit faults, partial shading faults, and degradation faults are simulated respectively. Specifically, the short-circuit faults are realized by connecting the two panels directly with wires, the partial shading faults are implemented by covering part of the PV panels with umbrellas, and the degradation faults are simulated by connecting a high-power resistor of 8 ohms in series to the PV array, as shown in Fig. \ref{fig_real experiments}.
The I-V characteristic curves and corresponding temperature and irradiance of normal and three fault states are collected under different weather conditions. 
For each state, 100 pieces of data are collected and reconstructed into 40*4 two-dimensional matrices as diagnostic samples through the data preprocessing method described above.

The experiments are implemented by using Pytorch and are conducted on HP ZBook Create G7 with Intel (R) Core (TM) i9-10885H CPU, 32GB RAM, NVIDIA GeForce RTX 2070 Max-Q.

\subsection{Experimental Results}
To verify the effectiveness of the proposed method in practical applications, six groups of experiments are carried out based on the actual PV array in terms of accuracy, communication overhead, and training time, as shown in Table \ref{table_real experiment}.
In each experiment, we set the local data of different agents to contain different types of fault samples. Specifically, if the local data of an agent contains a certain type, the number of samples is 100. Otherwise, it is 0. Taking agent 1 as an example, the first row below the normal column, i.e. \textbf{100 / 100 / 100 / 100 / 100}, indicates that the sample numbers in the normal state of agent 1 are 100 in all six experiments. These samples are further divided into the training set and test set by 7:3 as described in Section 5.1.
To ensure the validity of the accuracy comparison, centralized FL and centralized training use the same dataset of the above six experiments as ADFL. It is worth noting that only centralized training can share data. At this time, the training sets of the three agents are combined to train the model in each experiment.

\subsubsection{Accuracy of Local Model without ADFL}
\begin{figure*} [htbp]
	\centering 
	\subfigure[Accuracy of agent 1]{ 
		\includegraphics[width=0.25\textwidth]{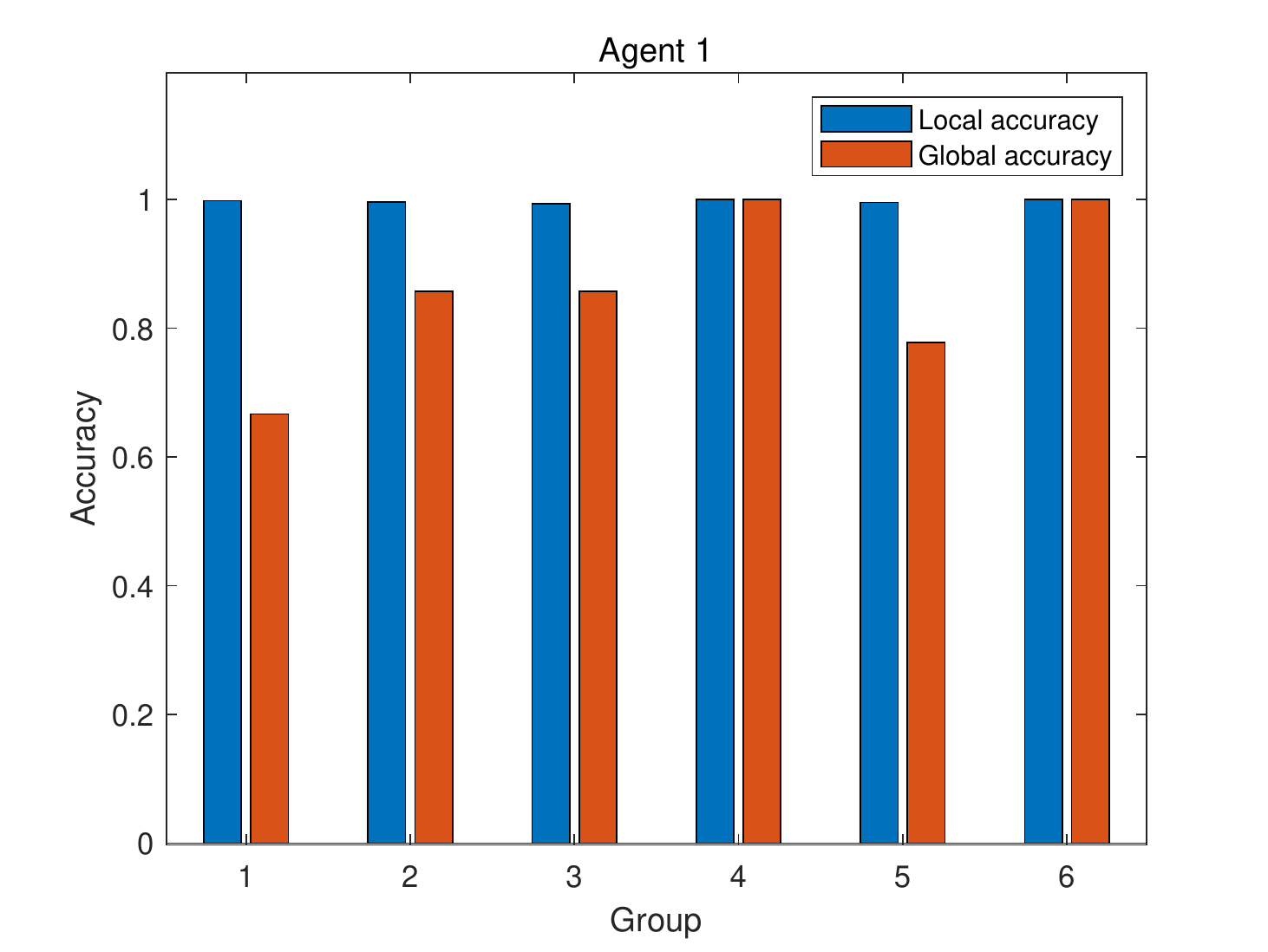} 
	} 
	\subfigure[Accuracy of agent 2]{ 
		\includegraphics[width=0.25\textwidth]{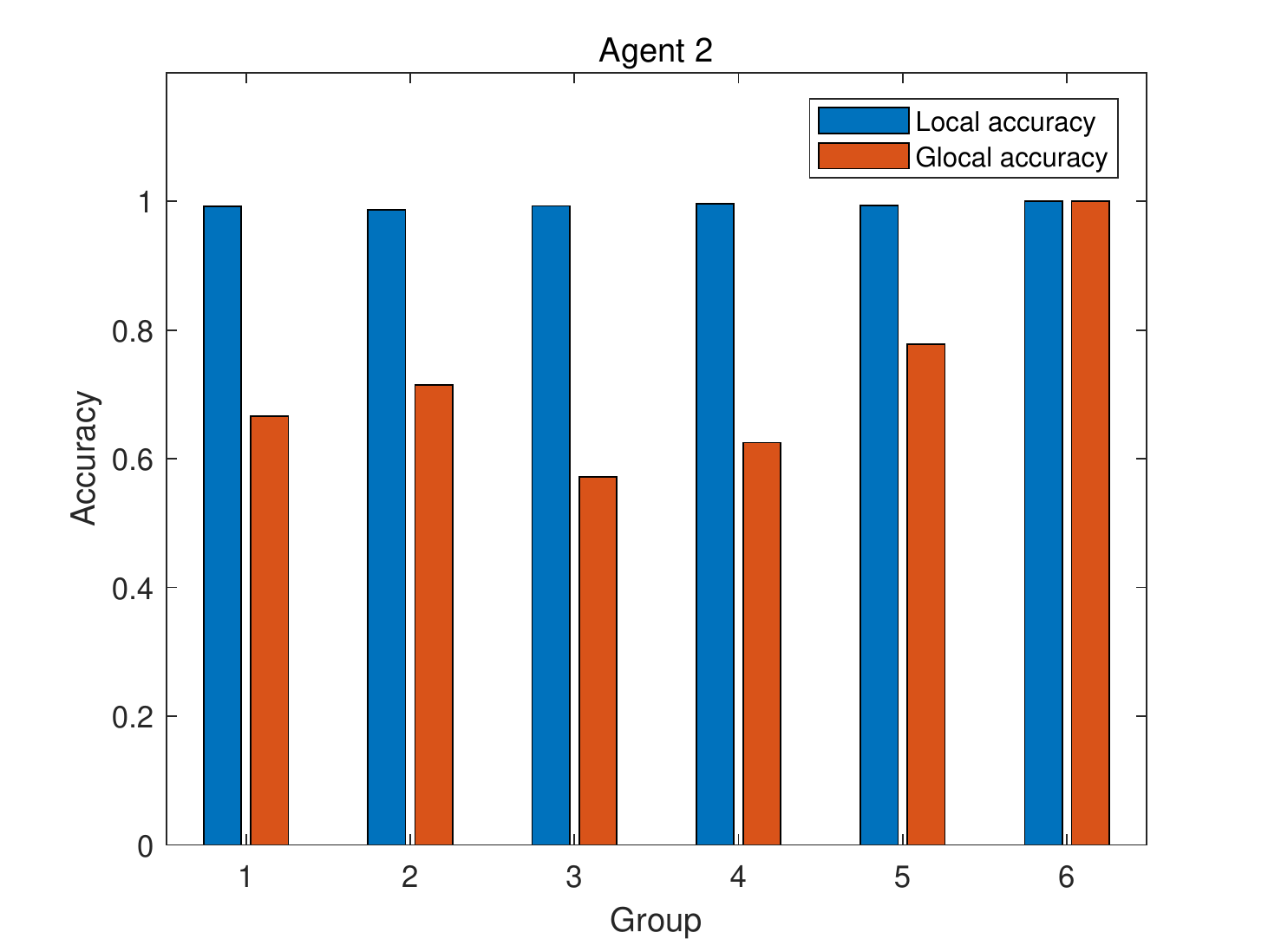} 
	} 
	\subfigure[Accuracy of agent 3]{ 
		\includegraphics[width=0.25\textwidth]{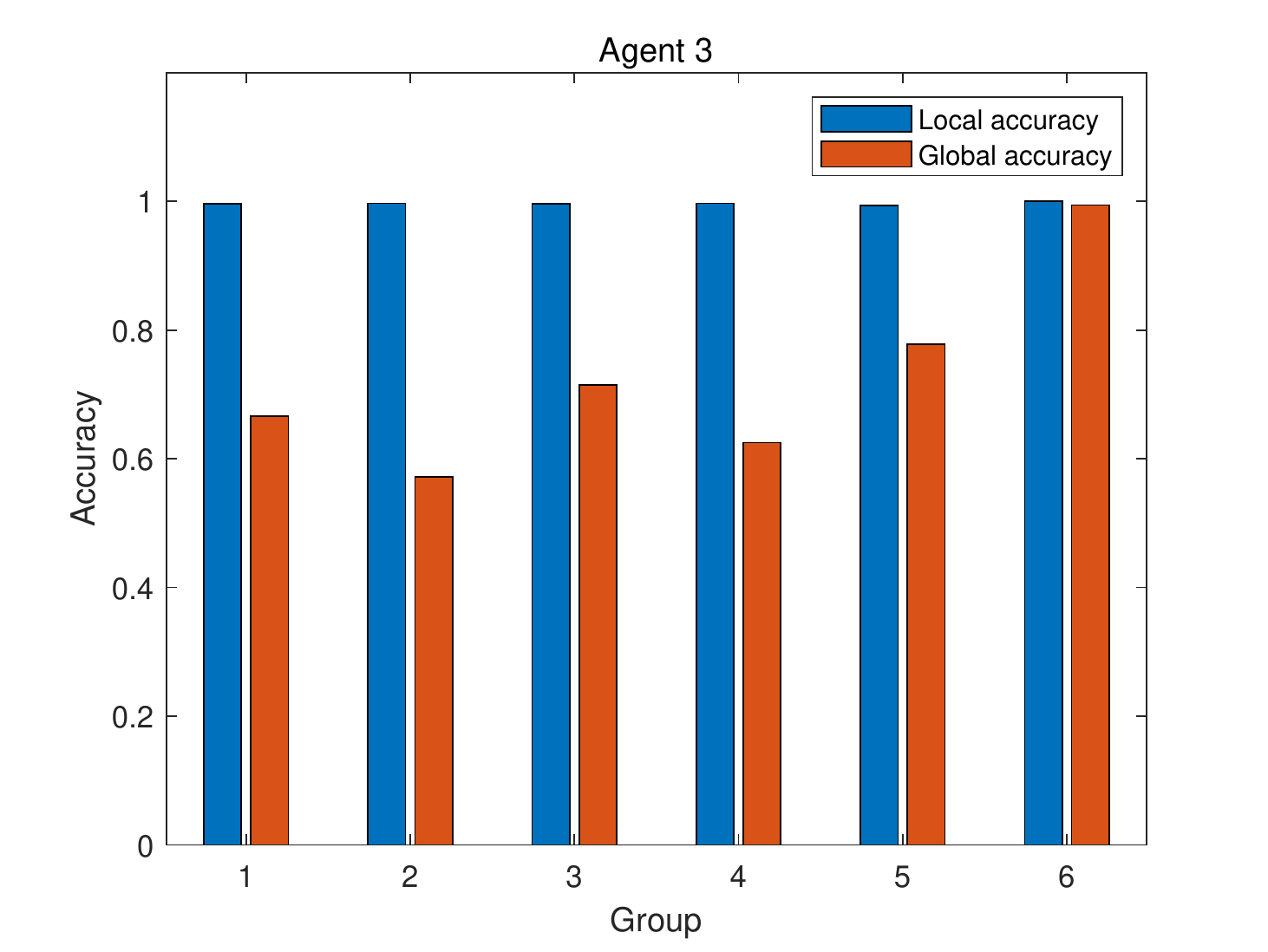} 
	} 
	\caption{The fault diagnosis accuracy of each agent without the proposed method} 
	\label{fig_client1}
\end{figure*}

We first evaluate the local and global fault diagnosis accuracy of the local models of the three agents in the six groups of experiments without introducing the proposed ADFL method, as shown in Fig. \ref{fig_client1}.
It is worth noting that the local here refers to the samples owned by the agent itself, and the global refers to all the samples of the three agents.

\begin{figure*} [htbp]
	\centering 
	\subfigure[Accuracy of agent 1]{ 
		\includegraphics[width=0.25\textwidth]{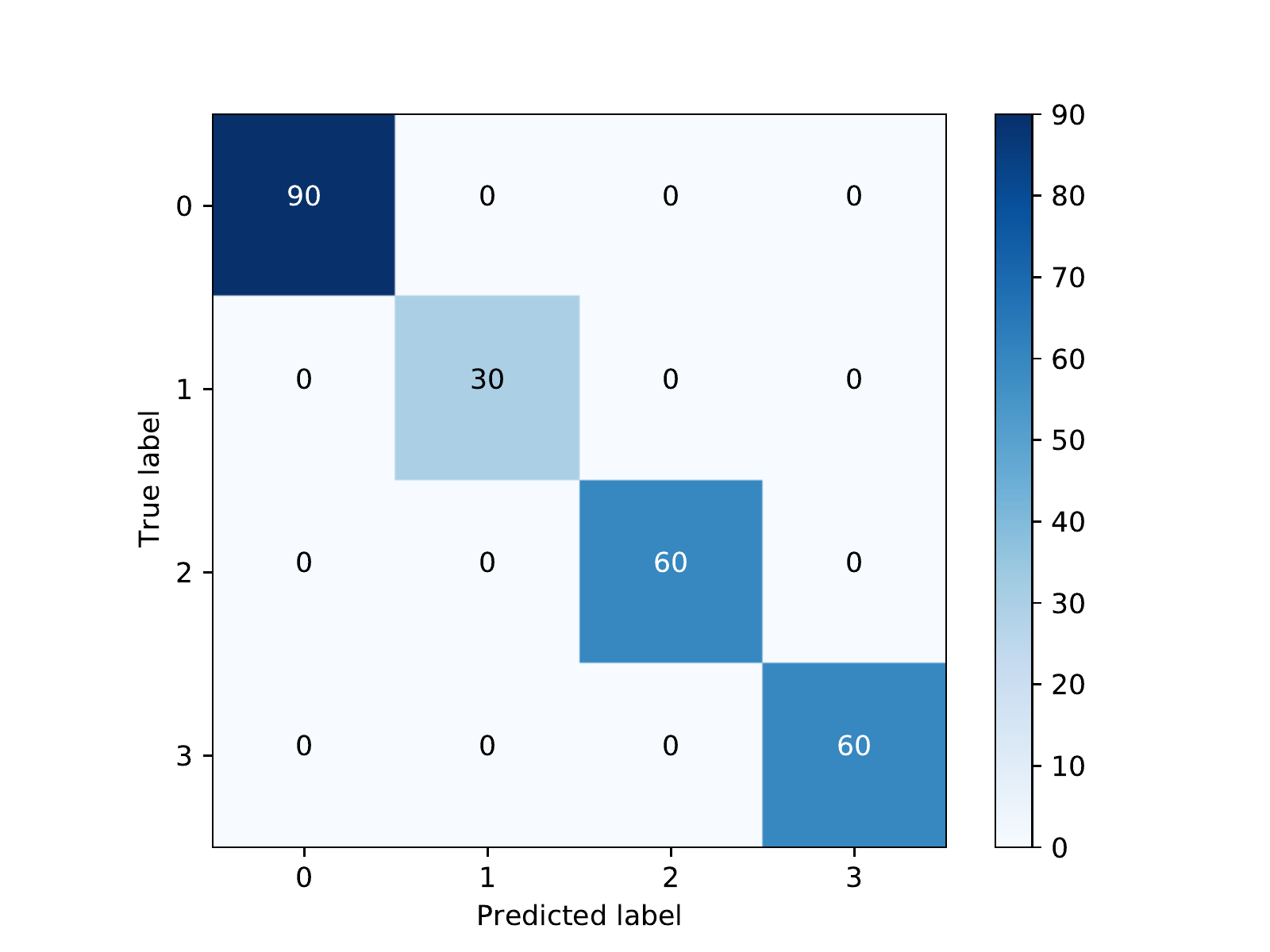} 
	} 
	\subfigure[Accuracy of agent 2]{ 
		\includegraphics[width=0.25\textwidth]{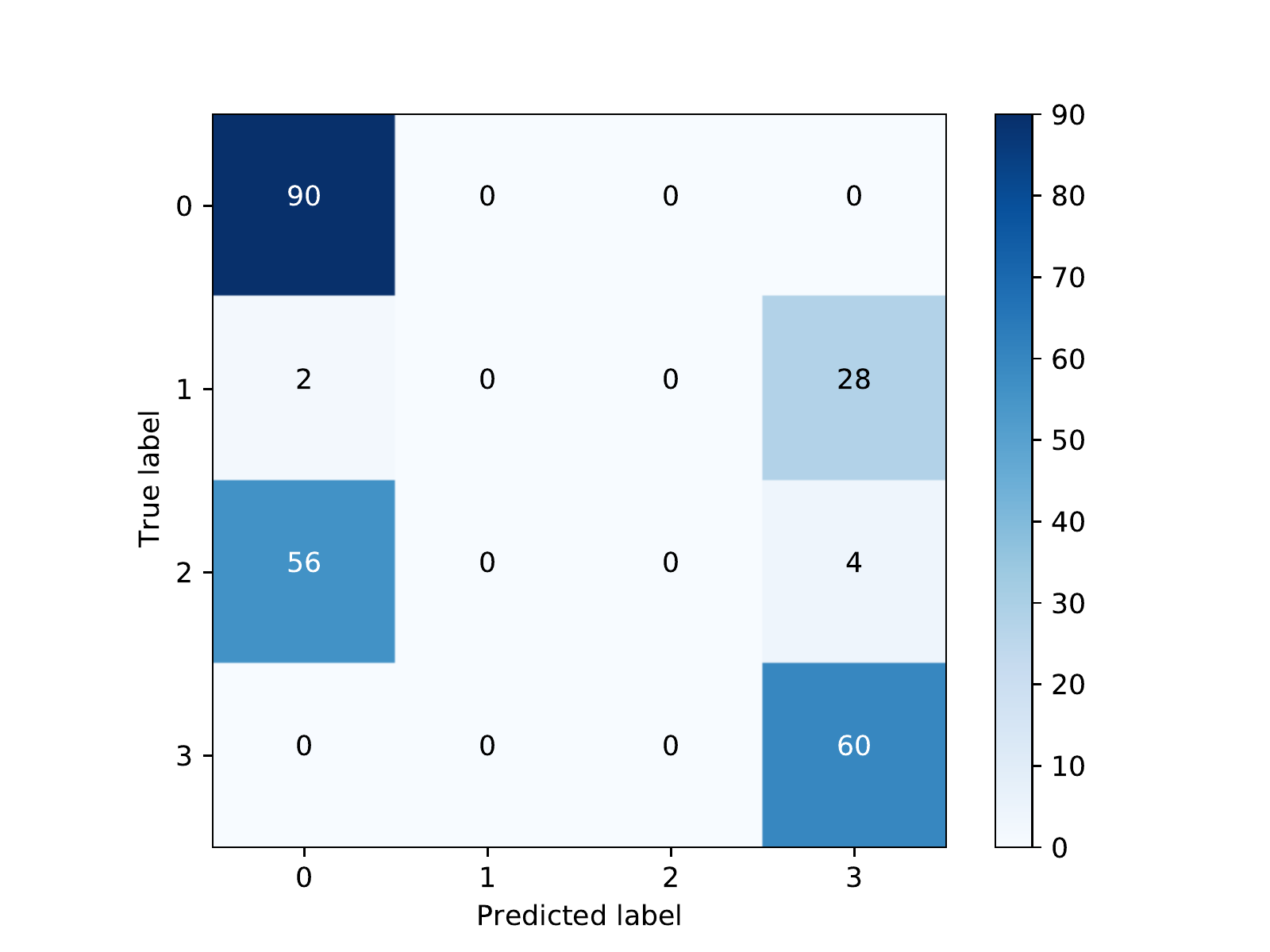} 
	} 
	\subfigure[Accuracy of agent 3]{ 
		\includegraphics[width=0.25\textwidth]{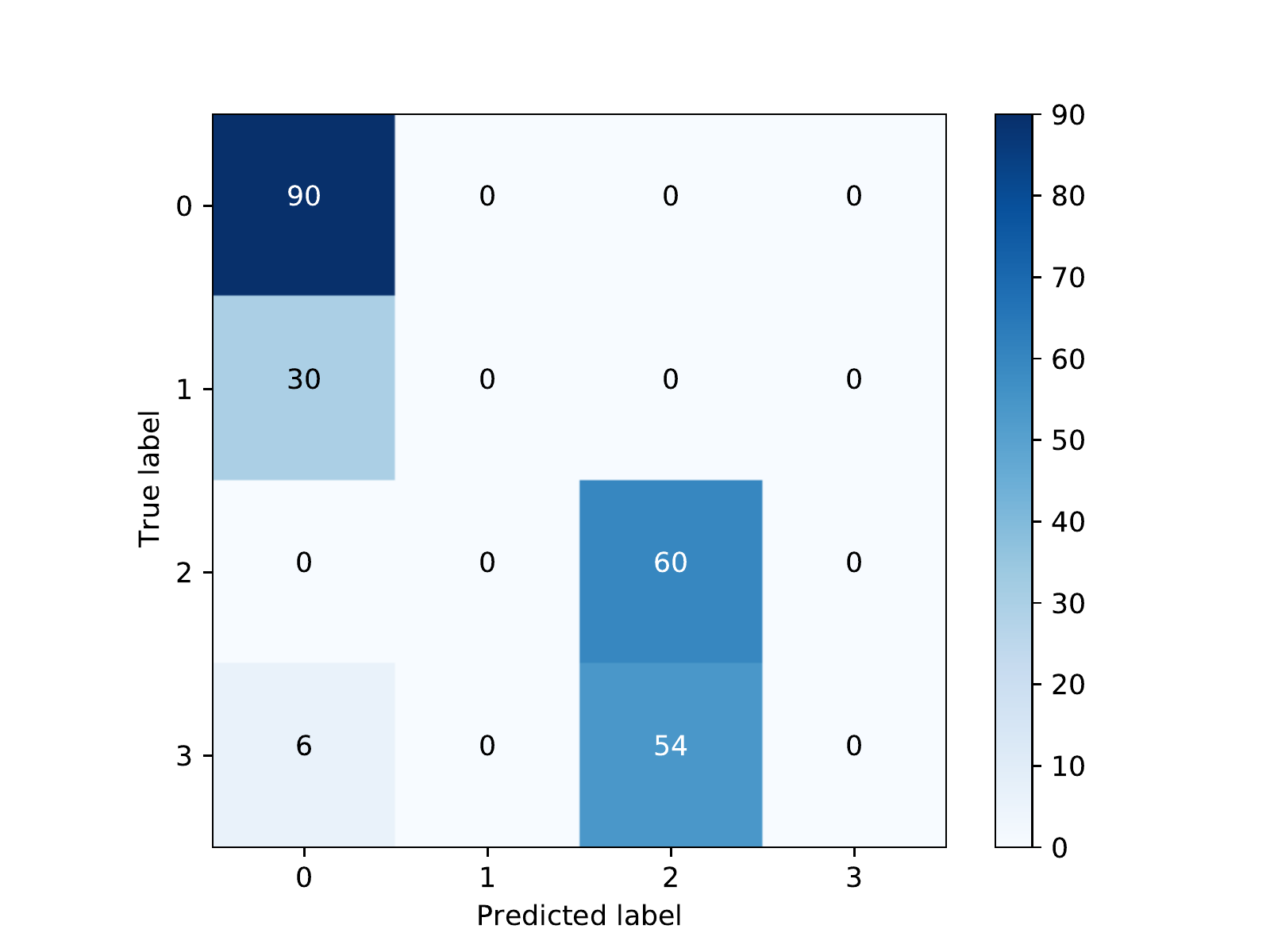} 
	} 
	\caption{The global fault classification confusion matrix of the three agents} 
	\label{fig_e4_global}
\end{figure*}

Obviously, in all experiments, each agent reaches 99\% accuracy on its local dataset, which shows that the CNN used for training accurately extracts fault features. However, it is difficult for a single PV station to effectively diagnose all the fault types based on its local model in the case that both the amount and types of fault samples are insufficient. 
Specifically, taking experiment 4 as an example, the global fault classification confusion matrix of the three agents is plotted in Fig. \ref{fig_e4_global}. 
It can visually reveal the accuracy of the classifier for each type of fault sample.
It can be seen that the global fault diagnosis accuracy of agent 2 and agent 3 is only 62.5\% because the CNN cannot learn the fault characteristic of the missing types.
Only when the agent has samples of all fault types, the local model can achieve high global accuracy, such as agent 1.
Therefore, it is significant to achieve collaborative fault diagnosis of multiple PV stations through the proposed ADFL method, which improves the generalization ability of the model.

\subsubsection{Accuracy of The Global Model with ADFL}
\begin{figure*} [!tbp]
	\centering 
	\subfigure[Accuracy of experiment 1]{ 
		\includegraphics[width=0.25\textwidth]{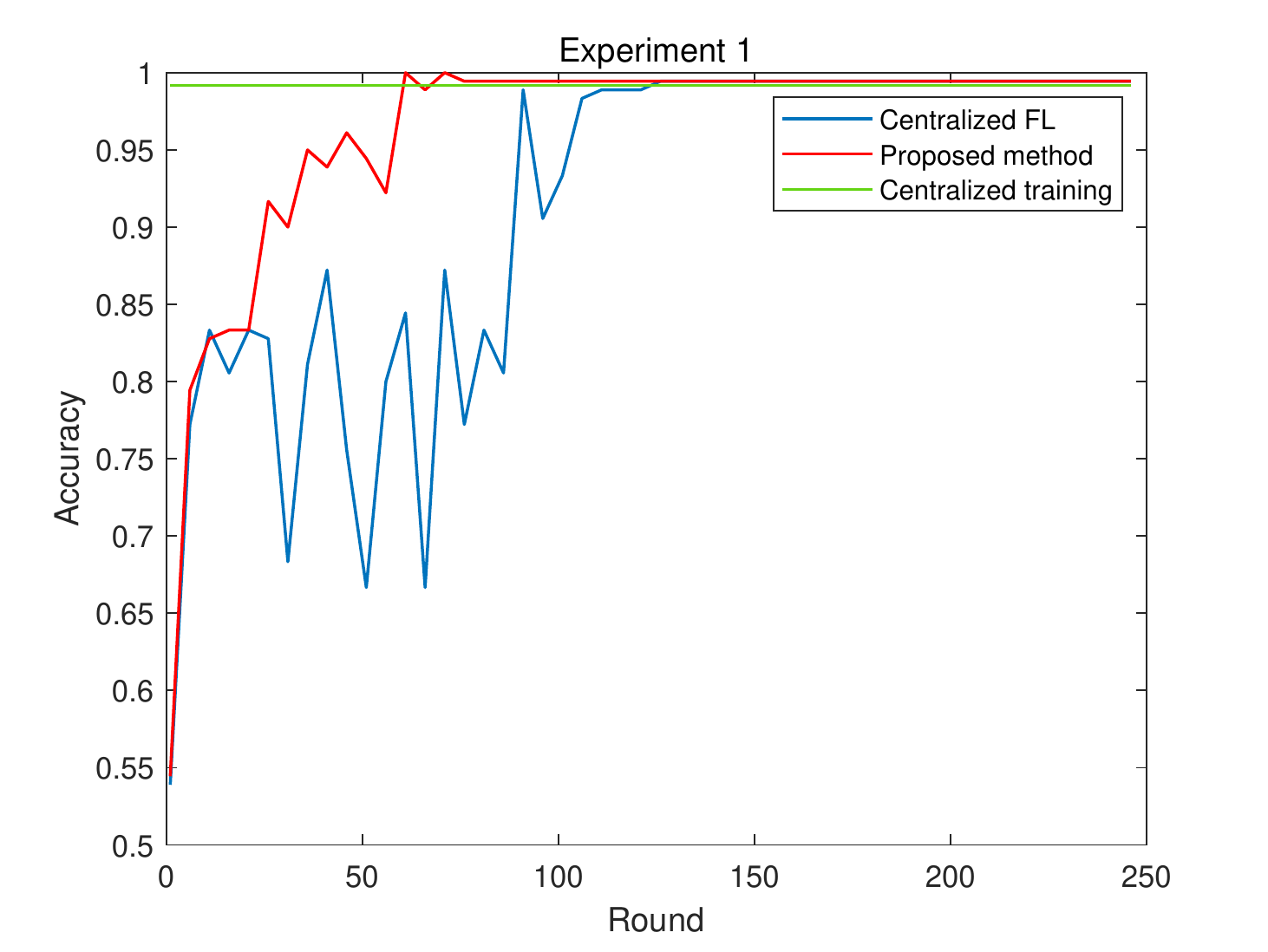} 
	} 
	\subfigure[Accuracy of experiment 2]{ 
		\includegraphics[width=0.25\textwidth]{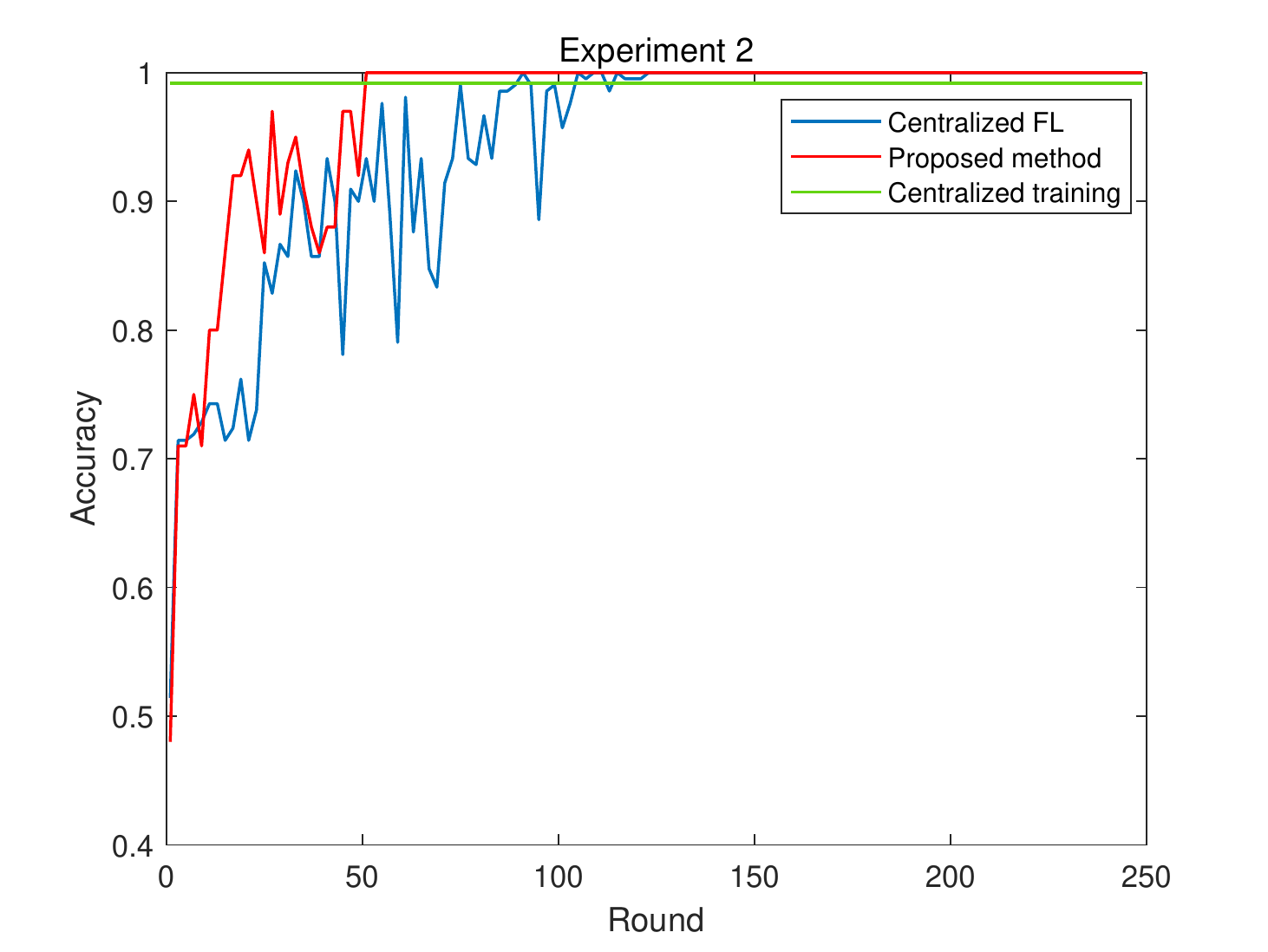} 
	} 
	\subfigure[Accuracy of experiment 3]{ 
		\includegraphics[width=0.25\textwidth]{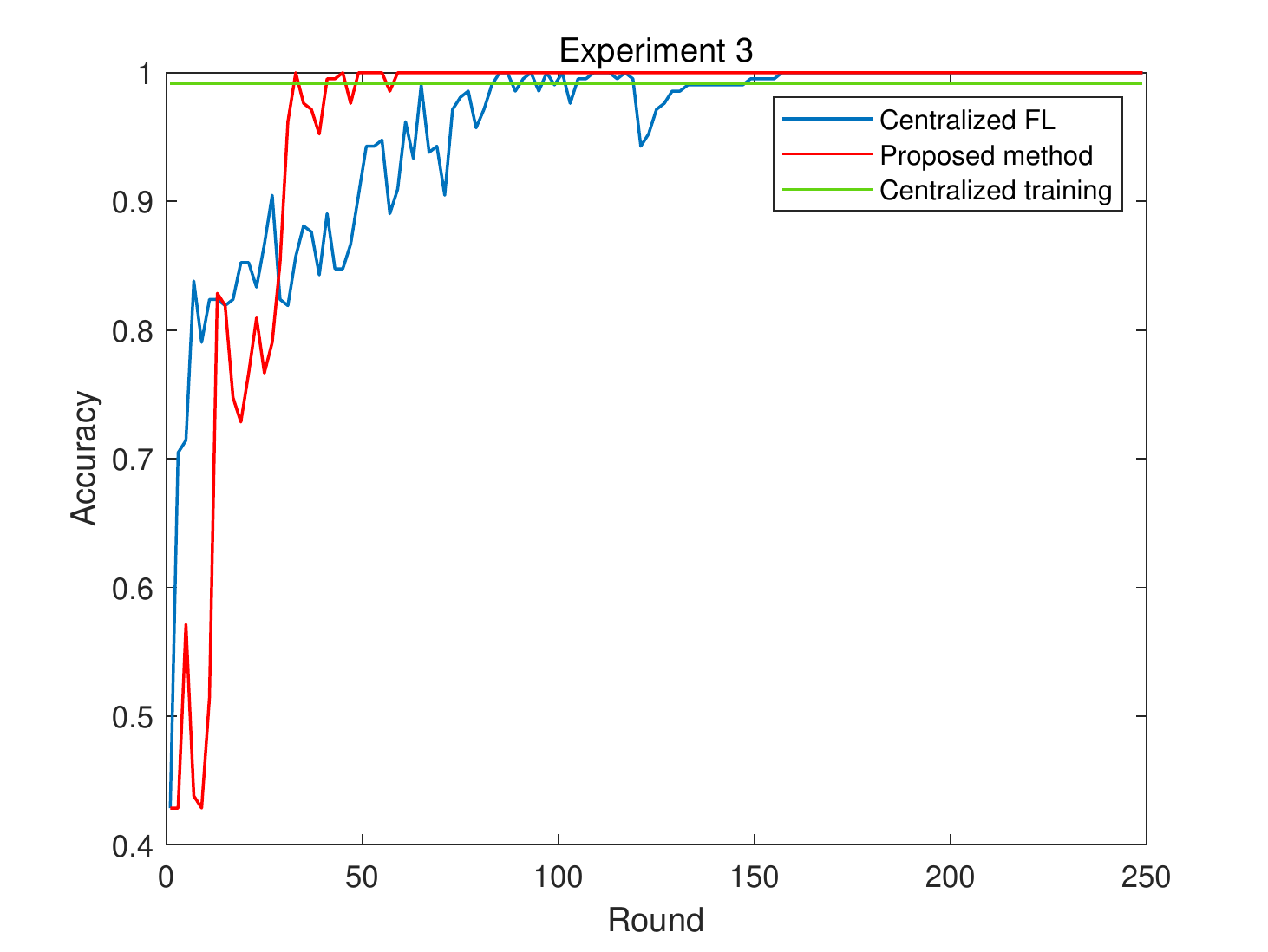} 
	} 
	\subfigure[Accuracy of experiment 4]{ 
		\includegraphics[width=0.25\textwidth]{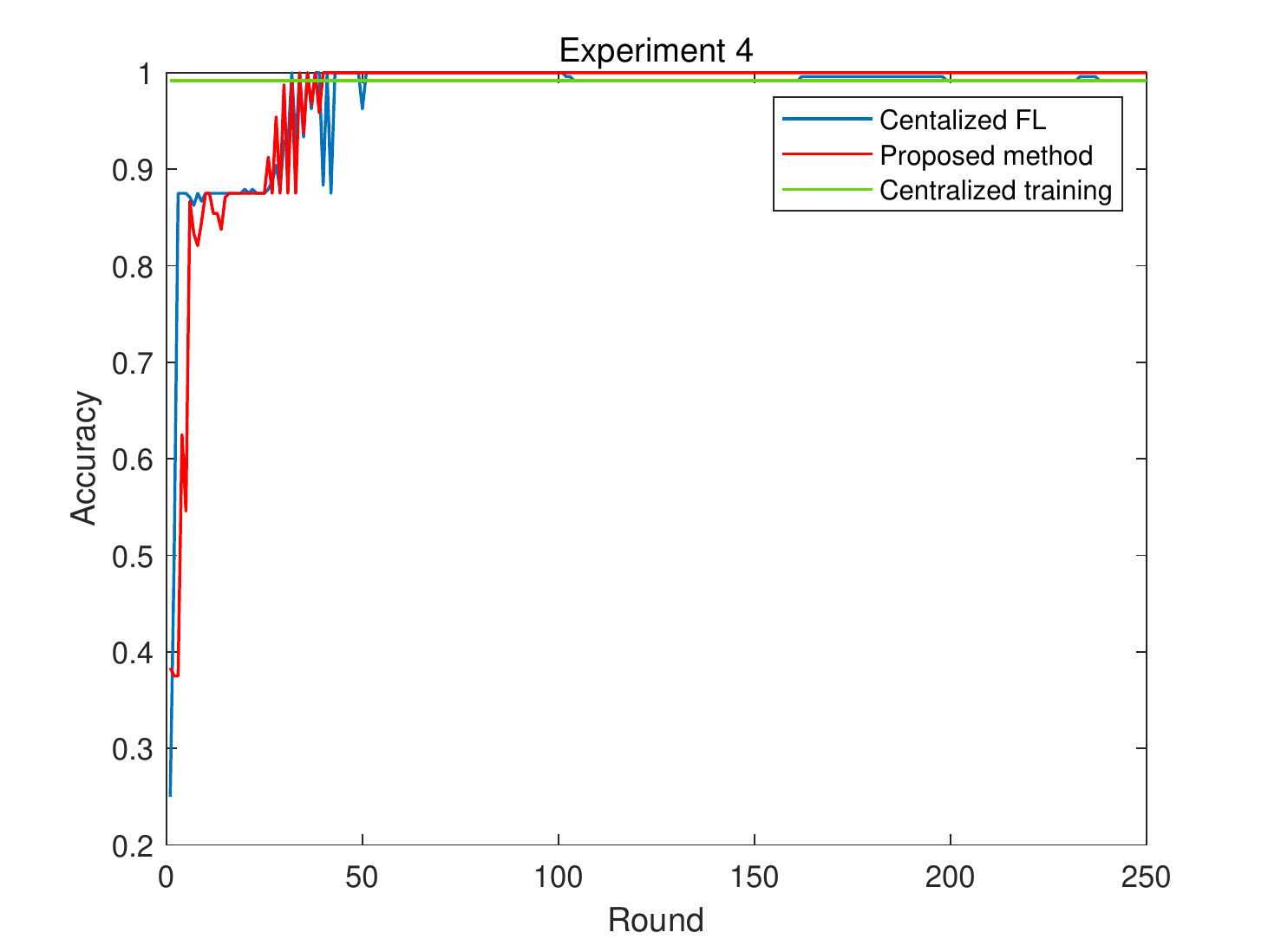} 
	} 
	\subfigure[Accuracy of experiment 5]{ 
		\includegraphics[width=0.25\textwidth]{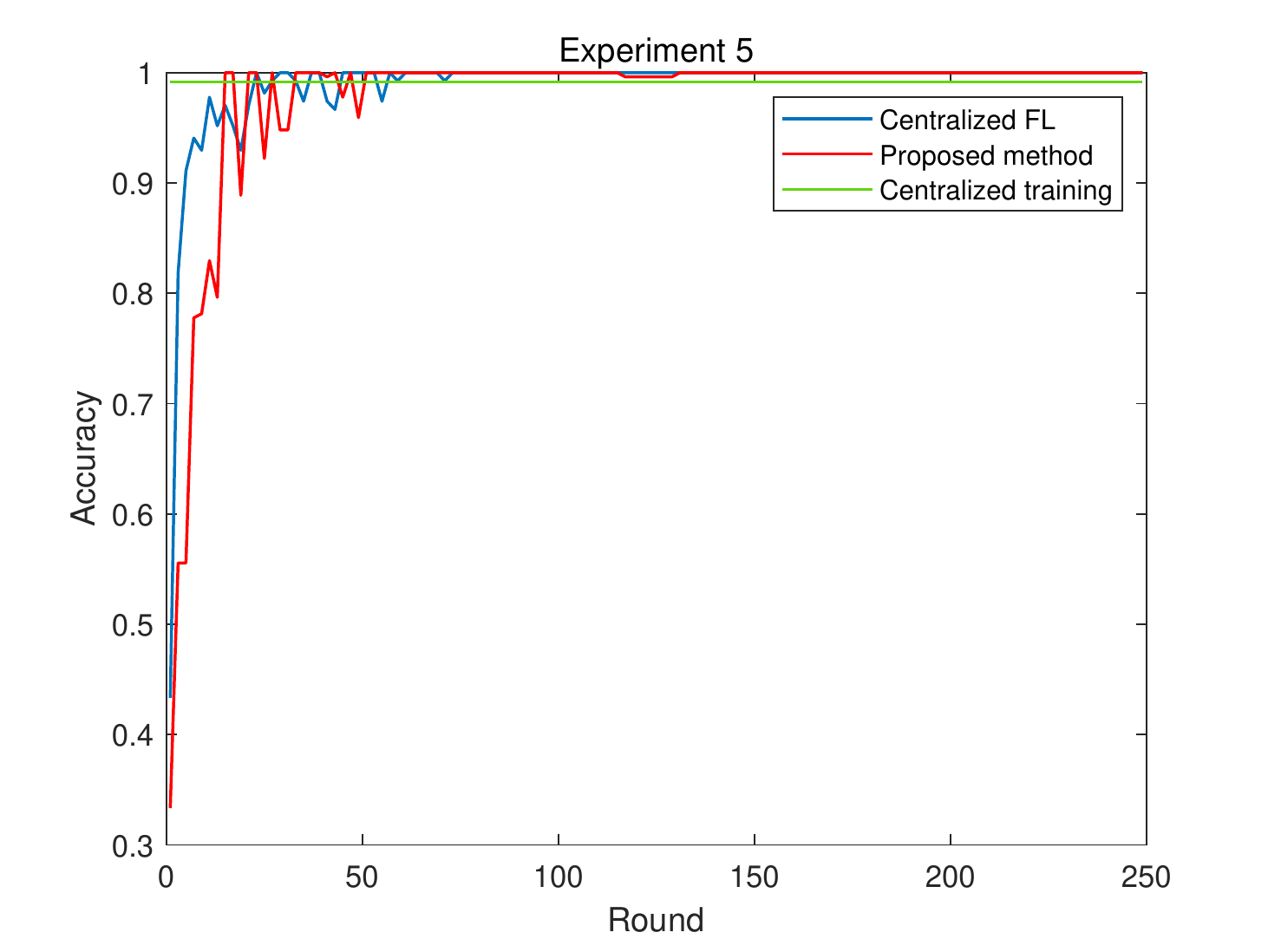} 
	} 
	\subfigure[Accuracy of experiment 6]{ 
		\includegraphics[width=0.25\textwidth]{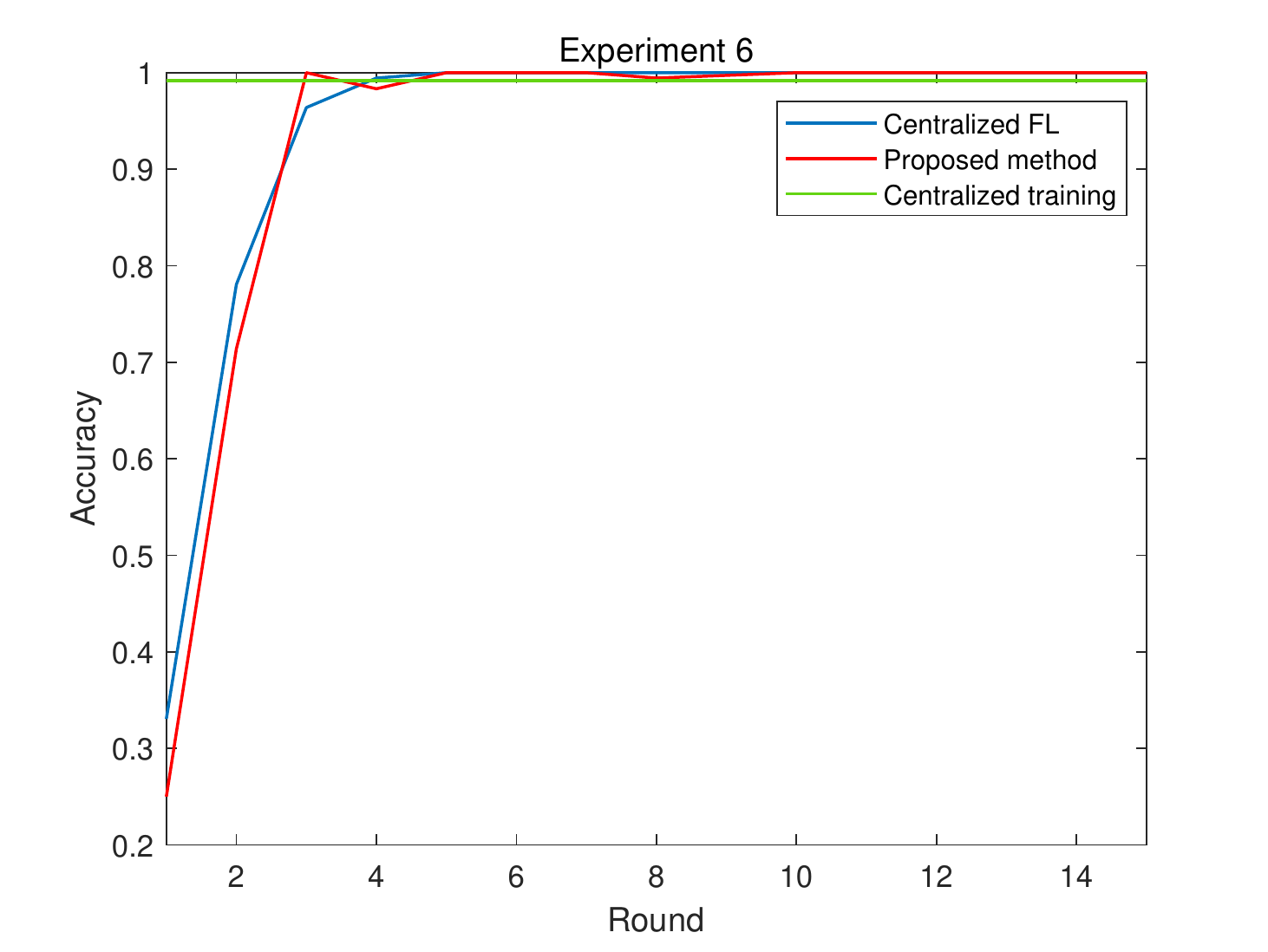} 
	} 
	\caption{Global fault diagnosis accuracy of six real experiments} 
	\label{fig_real global}
\end{figure*}

Collaborative fault diagnosis modeling is carried out under six groups of experiments to verify the effectiveness of the ADFL method. The global accuracy of the collaborative models varies with aggregation rounds and is further compared with the centralized FL and the centralized training, as shown in Fig. \ref{fig_real global}.
Since the centralized training does not involve the model aggregation and has nothing to do with the rounds, its accuracy is represented by a straight line with a constant value.

It can be observed that the proposed method converges to 99\% accuracy after a certain round of aggregation, which is almost the same as the centralized training method.
Taking agent 2 as an example, the global accuracy in the first five groups of experiments is improved by about 33\%, 28\%, 43\%, 38\%, and 23\% respectively. 
This is mainly because the proposed ADFL method enables the collaborative model to learn the fault features in the local data of all agents while ensuring data privacy.
In addition, compared with the traditional centralized FL, the proposed method achieves convergence with fewer aggregation rounds. 

\subsubsection{Communication and Training Efficiency}
\begin{figure} [!tbp]
	\centering 
	\subfigure[Communication efficiency]{ 
		\includegraphics[width=0.23\textwidth]{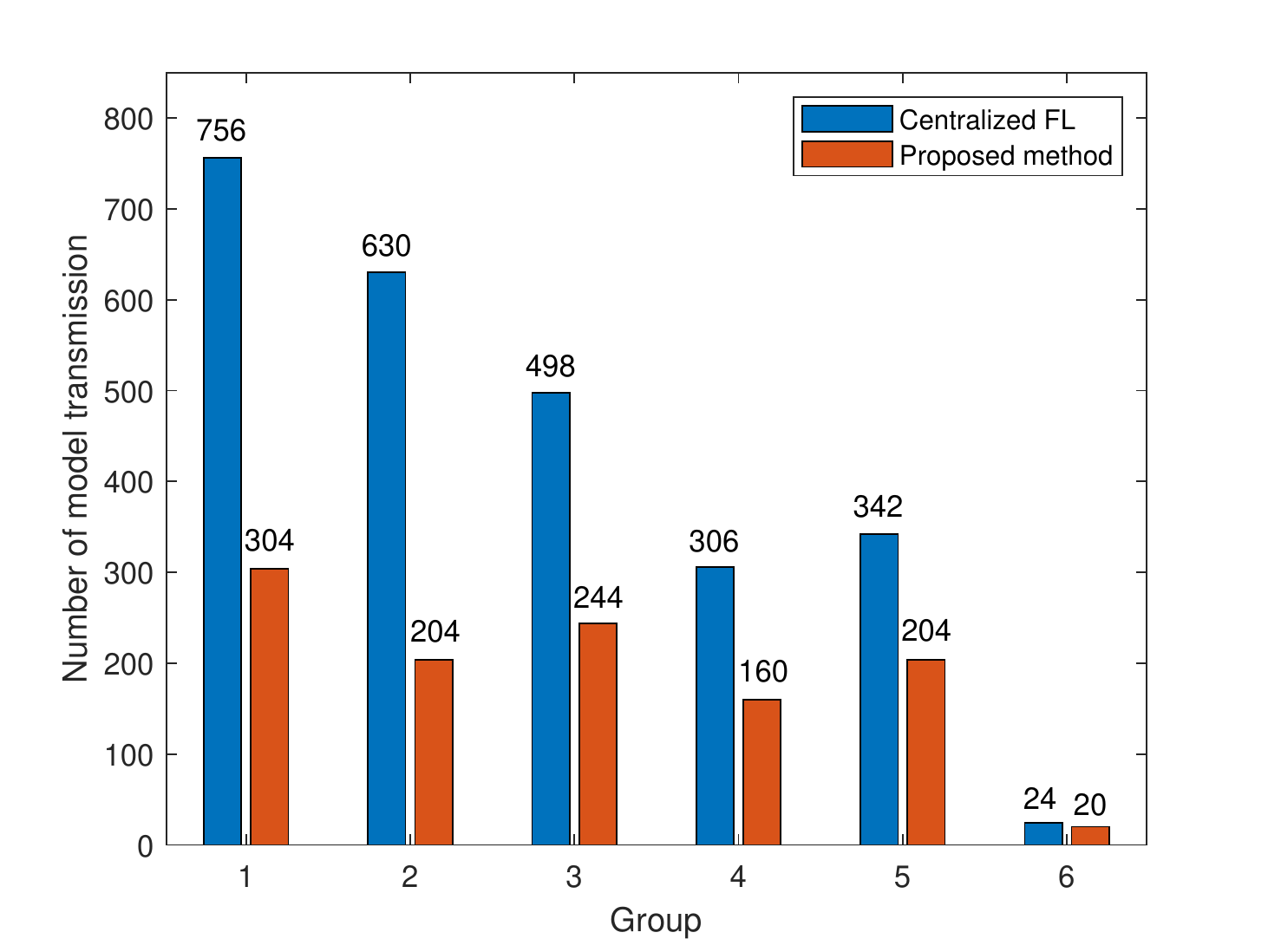} 
	} 
	\subfigure[Training time]{ 
		\includegraphics[width=0.23\textwidth]{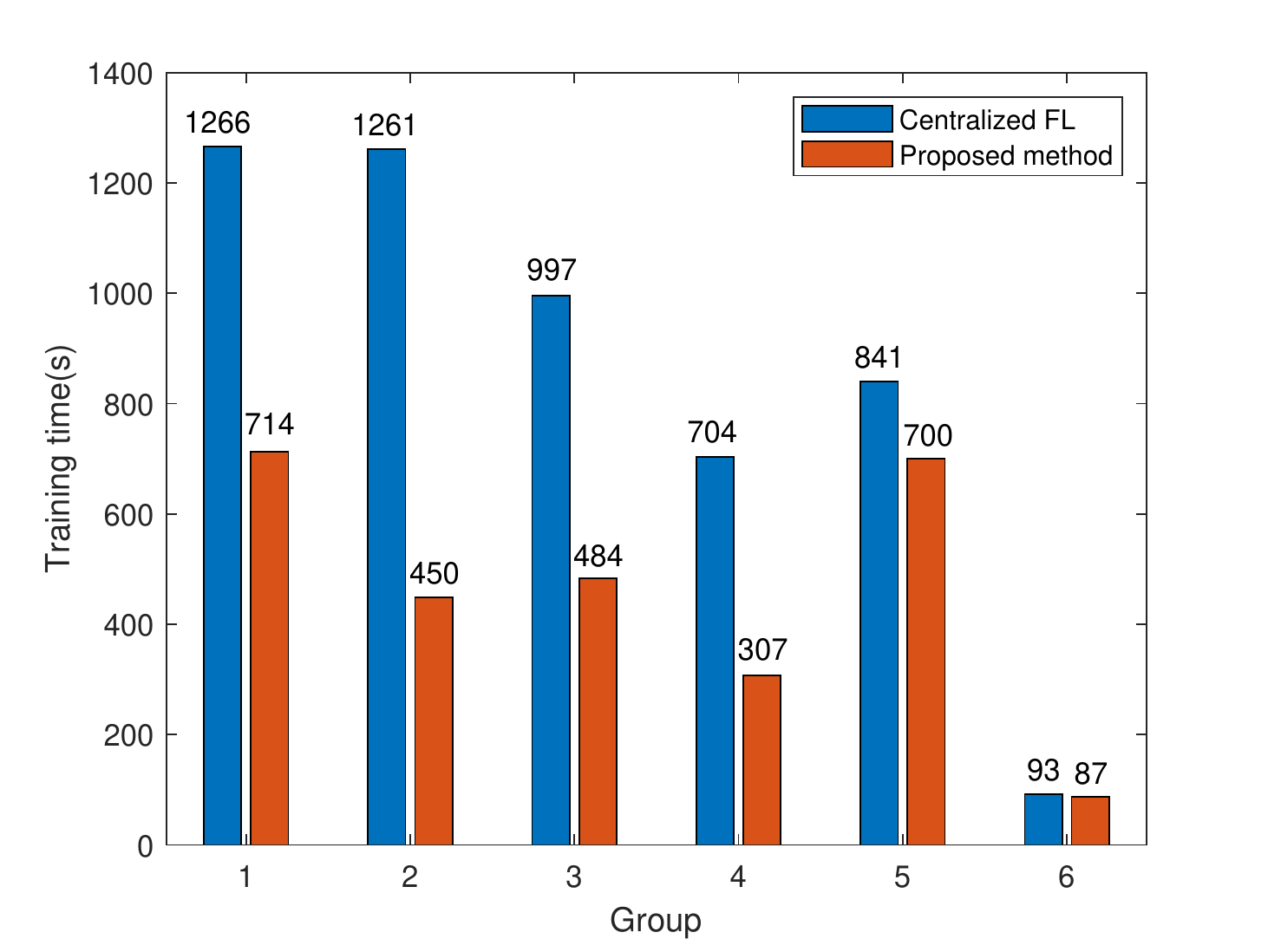} 
	} 	
	\caption{Communication overhead and training time in experiments} 
	\label{real_efficiency}
\end{figure}

The main factors affecting communication overhead include the number of parameters transmission in each round of global model aggregation and the number of rounds required for model convergence.
In general, the greater the local sample difference between agents, the more aggregation rounds required for the global model to converge. 
In this paper, the communication overhead is set as the total number of parameters transmission when the global model converges. 
The communication overhead is compared with that of centralized FL under six groups of experiments, as shown in Fig. \ref{real_efficiency} (a).
Obviously, the proposed method significantly reduces the communication overhead in the process of model aggregation, especially when the local data of agents is greatly heterogeneous. 
For example, the communication overhead is reduced by approximately 60\%, 67\%, and 51\% in experiment 1, experiment 2, and experiment 3. 
Even if the fault samples of agents are similar, this method still saves more than 16\% of the transmission costs, such as experiment 6.
\begin{table*}[!t]
	\renewcommand{\arraystretch}{1.3}
	\caption{Dataset configuration for each agent in six numerical simulation experiments}
	\label{table_experiment}
	\centering
	\resizebox{\textwidth}{!}{
		\begin{tabular}{c c c c c c}
			\hline
			\diagbox{\textbf{Participants}}{\textbf{Sample numbers of each simulation}}{\textbf{Types}} & \textbf{Normal} & \textbf{Short-current} & \textbf{Degradation} & \textbf{Partial shading}\\
			\hline
			Agent 1 & 2976/ 2976/ 2976/ 2976/ 2976/ 2976 & 2976/ 2976/ 2976/ 2976/ 2976/ 2976 & 0/ 2976/ 0/ 2976/ 2976/ 2976 & 0/ 0/ 2976/ 2976/ 0/ 2976\\
			Agent 2 & 2976/ 2976/ 2976/ 2976/ 2976/ 2976 &  0/ 0/ 0/ 0/ 0/ 2976 & 2976/ 2976/ 2976/ 2976/ 2976/ 2976 & 0/ 0/ 0/ 0/ 2976/ 2976\\ 
			Agent 3 & 2976/ 2976/ 2976/ 2976/ 2976/ 2976 &  0/ 0/ 0/ 0/ 2976/ 2976 & 0/ 0/ 0/ 0/ 0/ 2976 & 2976/ 2976/ 2976/ 2976/ 2976/ 2976\\
			\hline
	\end{tabular}}
\end{table*}

\begin{table}[htbp]
	\caption{Setting of the simulation module parameters}
	\label{table_parameter}
	\centering
	\begin{tabular}{c c}
		\hline
		\textbf{Module parameters} & \textbf{Value} \\
		\hline
		Short-circuit resistance  $R_{\rm short}$ (ohms) & 0.001\\
		Degradation resistance $R_{\rm degradation}$ (ohms) & 3\\
		Partial shading gain $Gain_{\rm PS}$ & 0.5\\ 
		Maximum power point $M_{\rm pp}$ (W) & 99.925\\ 
		Open circuit voltage $V_{\rm oc}$ (V) & 21.5\\
		Voltage at maximum power point $V_{\rm mp}$ (V) & 17.5\\
		Short-circuit current $I_{\rm sc}$ (A) & 6.03\\
		Current at maximum power point $I_{\rm mp}$ (A) & 5.71\\
		Light-generated current $I_{\rm L}$ (A) & 6.0576\\
		Diode saturation current $I_{\rm 0}$ (A) & 2.0517e-10\\
		Diode ideality factor & 0.96445\\
		Shunt resistance $R_{\rm sh}$ (ohms) & 551.8793\\
		Series resistance $R_{\rm s}$ (ohms) & 0.2392\\
		Cells per module (Ncell) & 36\\
		Temperature coefficient of $V_{\rm oc}$ (\%/deg.C) & -0.36\\
		Temperature coefficient of $I_{\rm sc}$ (\%/deg.C) & 0.06\\
		\hline
	\end{tabular}
\end{table}

\begin{table}[!ht]
	\renewcommand{\arraystretch}{1.3}
	\caption{The longest continuous upload times of the same model for agents}
	\label{table_longest}
	\centering
	\begin{tabular}{c c c c c c c c}
		\hline
		\diagbox{\textbf{Participants}}{\textbf{Experiments}} & \textbf{1l} & \textbf{2} & \textbf{3} & \textbf{4} & \textbf{5} & \textbf{6}\\
		\hline
		Agent 1 & 5 & 6 & 6 & 3 & 2 & 0\\
		Agent 2 & 6 & 9 & 6 & 6 & 4 & 1\\ 
		Agent 3 & 26 & 15 & 13 & 12 & 17 & 2\\
		\hline
	\end{tabular}
\end{table}

Although the above results have shown that the proposed method converges faster to high accuracy and reduces the communication overhead, this may not necessarily reduce the time of collaborative fault diagnosis modeling, because the time consumed in each round is also a key factor \cite{26XiaWenchao}.
Especially for power systems, the speed of data processing is particularly important \cite{8727940,app11073101}.
To further evaluate the training efficiency of the proposed framework, the training time required for global model convergence in six groups of experiments is recorded separately and compared with that of centralized FL, as shown in Fig. \ref{real_efficiency} (b).
In each round of local updates, the epochs of local model training are set to 50 for agents. It can be seen that the proposed method greatly reduces the model training time, especially when the samples vary greatly between agents. 
For example, the training time is reduced by approximately 65\% in experiment 2. 


\subsection{Numerical Simulation Results}
\begin{figure*}[htbp]
	\centering
	\includegraphics[width=0.55\textwidth]{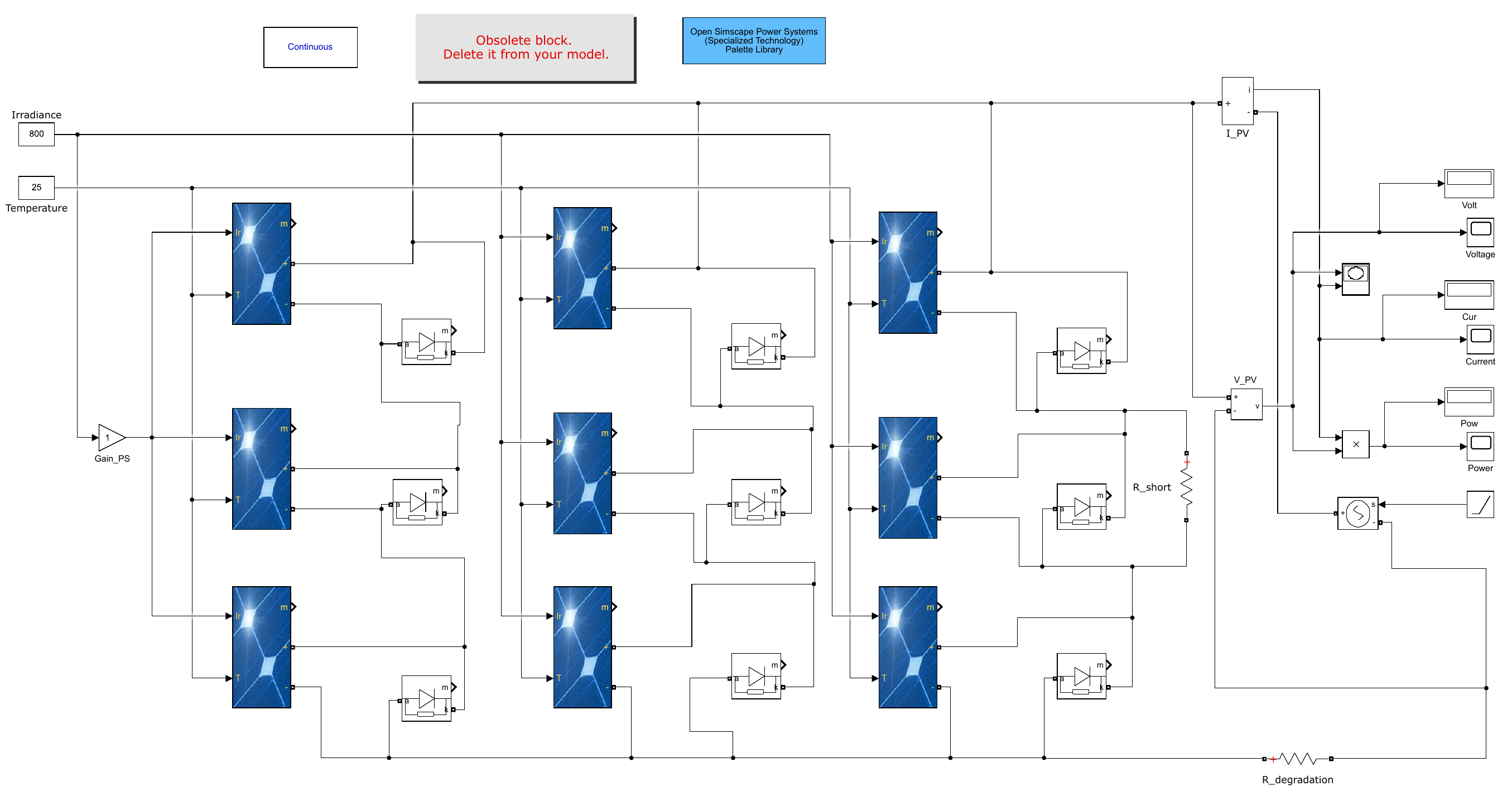}
	\caption{PV module fault simulation model}
	\label{fig_ZL}
\end{figure*}

Due to the limitation of weather and environmental conditions, the actual experimental samples are collected in situations where the temperature and irradiance change within a small range. 
To further verify that the proposed method is suitable for various meteorological conditions, a wider range of samples need to be collected. Therefore, a PV module fault simulation model based on Simulink is built as shown in Fig. \ref{fig_ZL}.
The simulation model consists of three parallel-connected PV strings with six PV modules in series. 
To save space, only three PV panels are shown in each column in the figure, where the second and third panels actually represent two and three PV modules respectively.

Specifically, short-circuit faults are simulated by connecting a resistor $R_{\rm short}$ with a small resistance in parallel to the PV module. 
A resistor $R_{\rm degradation}$ is connected in series on the trunk road of the PV array to increase the equivalent series resistance to simulate the degradation faults. 
Partial shading faults are simulated by connecting a gain module $Gain_{\rm PS}$ with a value between 0 and 1 in series between the PV panels and the irradiance module.
Some parameter settings of PV modules in the simulation model are listed in Table \ref{table_parameter}.

By setting the temperature module and irradiation module of the simulation model, the temperature change range is set from 10$^{\circ}$C to 70$^{\circ}$C, and the step length is 2$^{\circ}$C. At the same time, the irradiance change range is set from 50$W/m^{2}$ to 1000$W/m^{2}$, and the step length is 10$W/m^{2}$ \cite{23ZhicongChen3}. The I-V characteristic curves and the corresponding temperature and irradiance under all of these situations are collected by the I-V tester and temperature irradiance recording module.
Based on the simulation model, we select samples in the status of normal, short-circuit, degradation, and partial shading. There are 2976 samples of each state and the total is 11904. 
We also design six groups of numerical simulation experiments to evaluate the effectiveness of the proposed ADFL method in terms of accuracy, communication overhead, and training time, as shown in Table \ref{table_experiment}. 
The structure and hyperparameter settings of CNN are the same as the previous experiments.



\subsubsection{Accuracy of The Global Model with ADFL}
\begin{figure*} 
	\centering 
	\subfigure[Accuracy of experiment 1]{ 
		\includegraphics[width=0.25\textwidth]{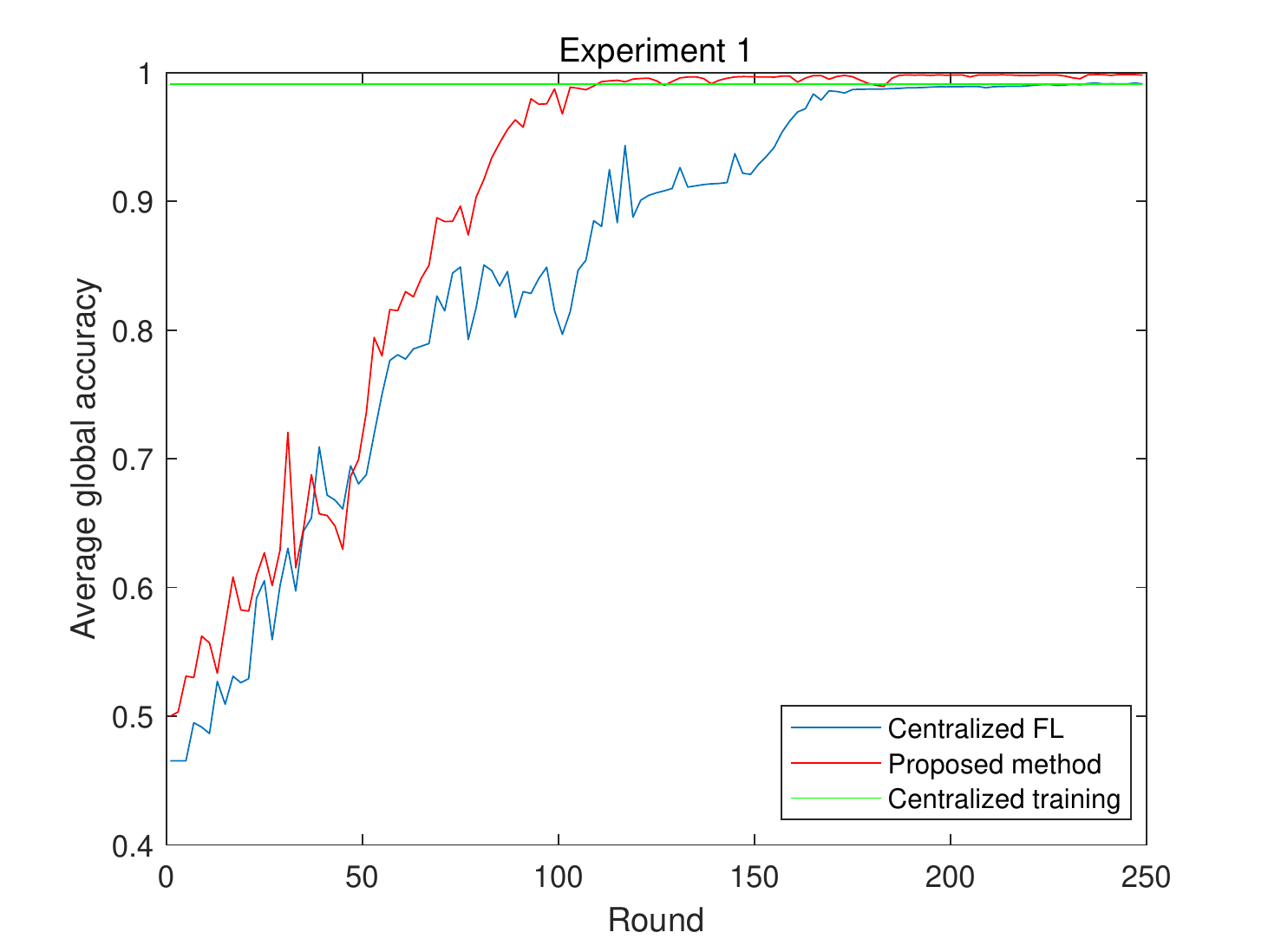} 
	} 
	\subfigure[Accuracy of experiment 2]{ 
		\includegraphics[width=0.25\textwidth]{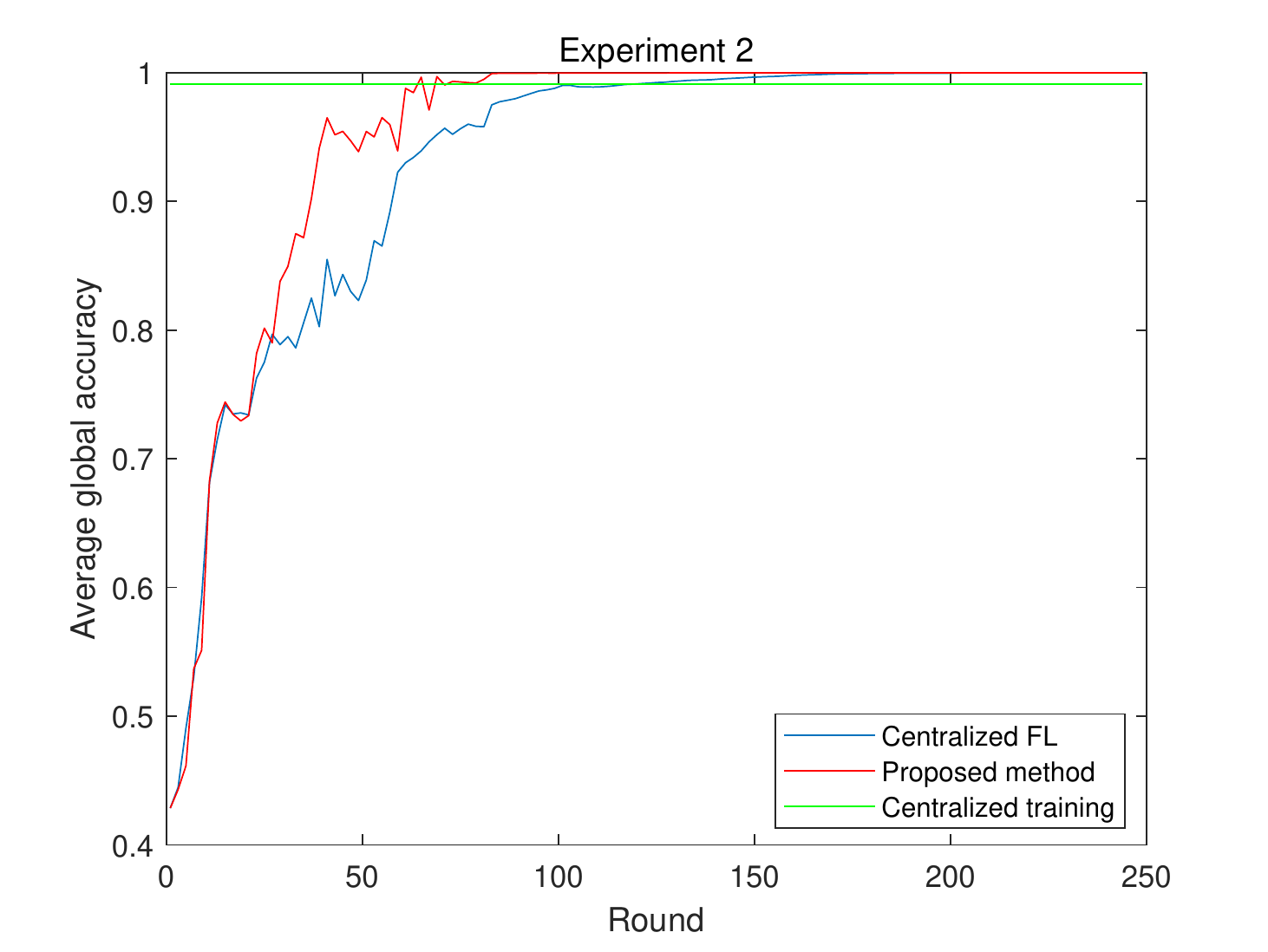} 
	} 
	\subfigure[Accuracy of experiment 3]{ 
		\includegraphics[width=0.25\textwidth]{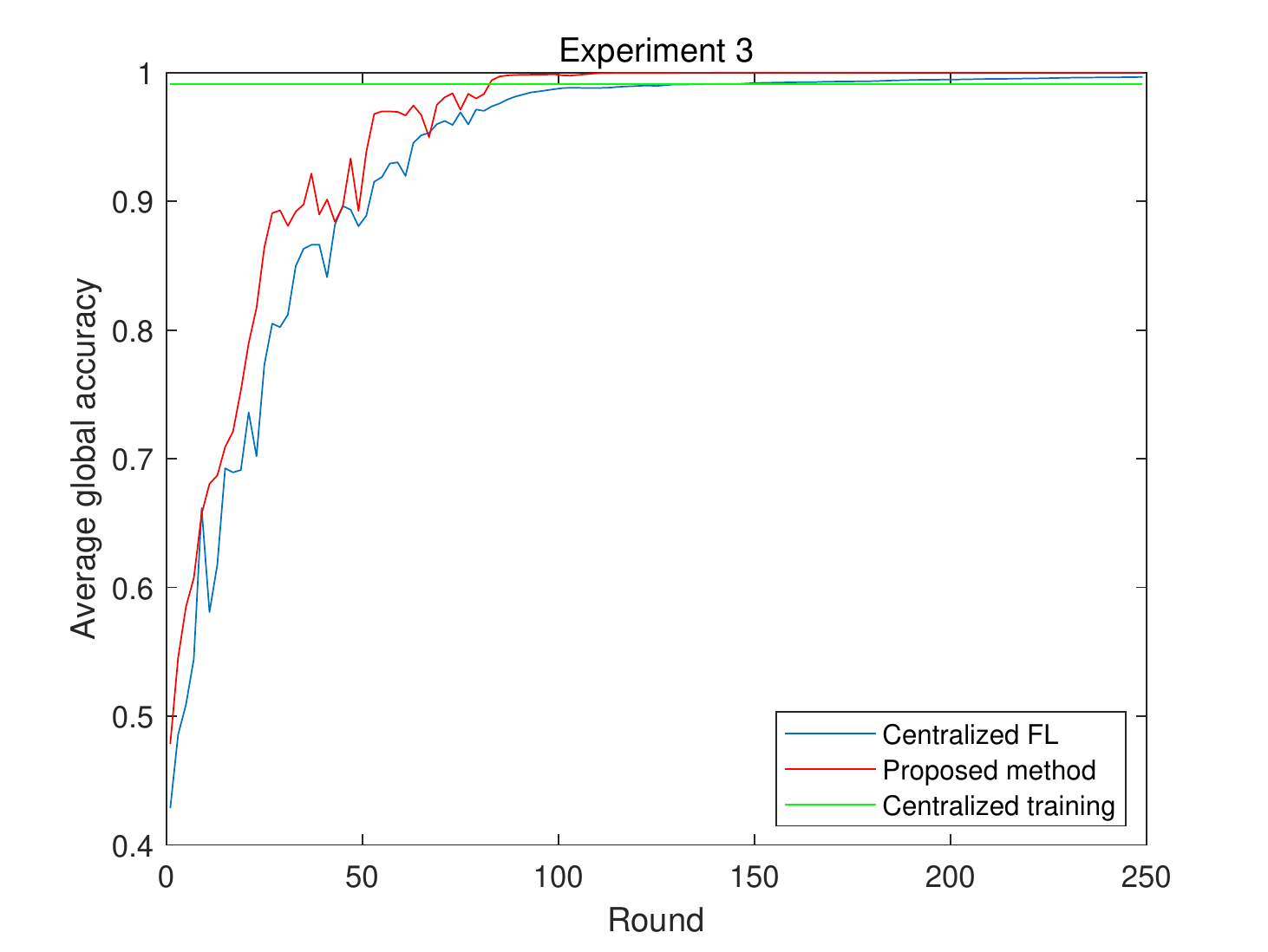} 
	} 
	\subfigure[Accuracy of experiment 4]{ 
		\includegraphics[width=0.25\textwidth]{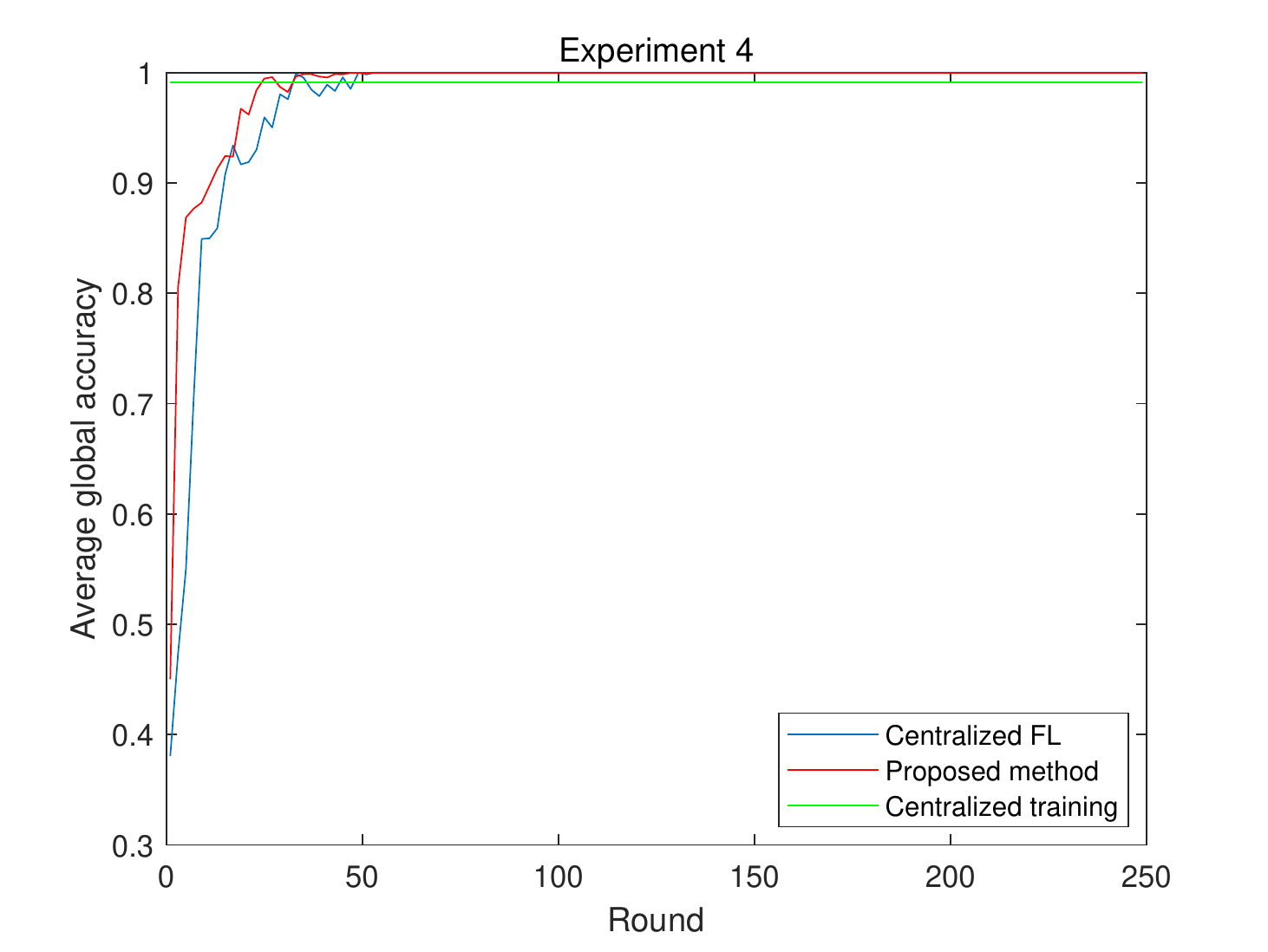} 
	} 
	\subfigure[Accuracy of experiment 5]{ 
		\includegraphics[width=0.25\textwidth]{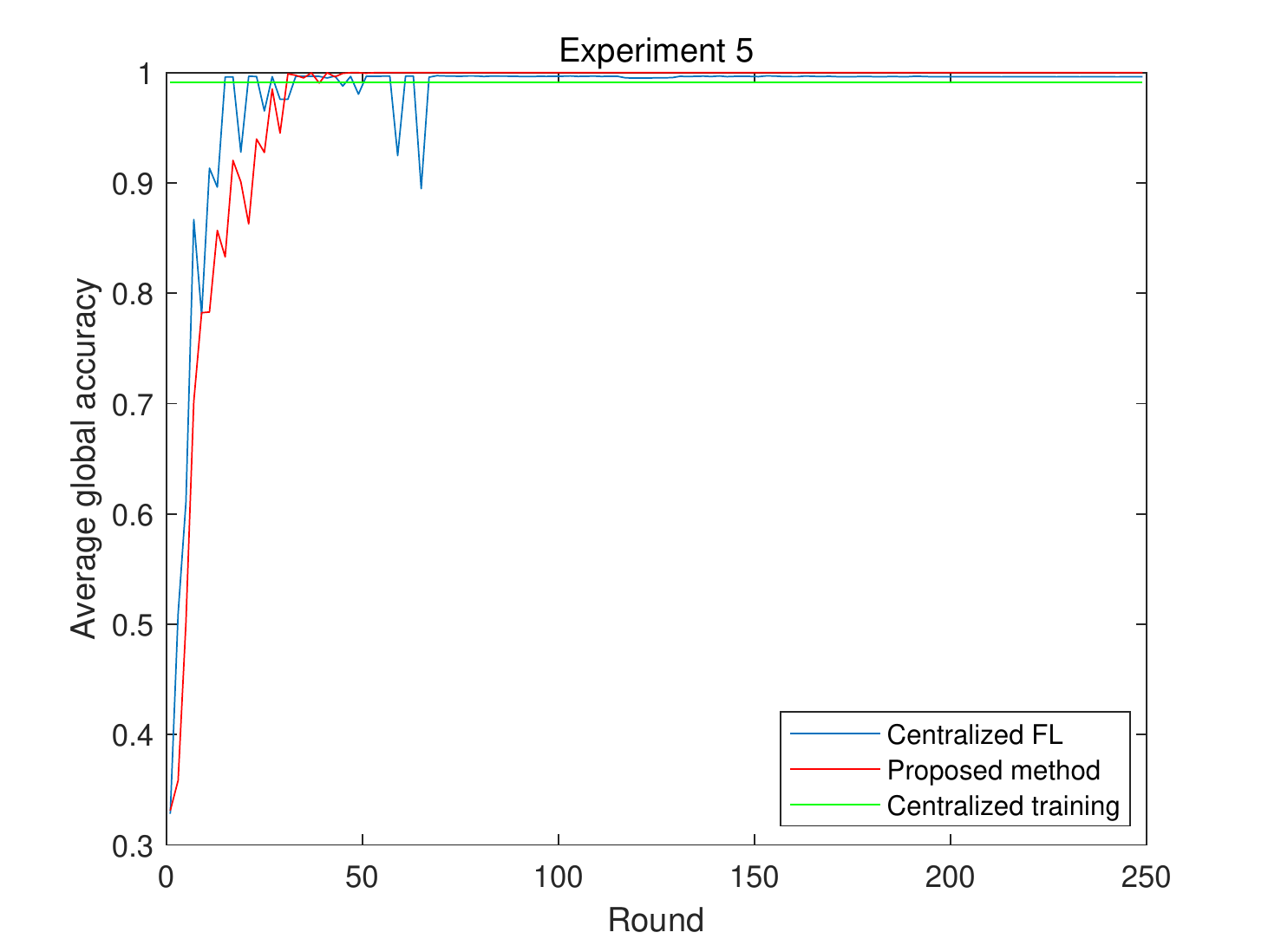} 
	} 
	\subfigure[Accuracy of experiment 6]{ 
		\includegraphics[width=0.25\textwidth]{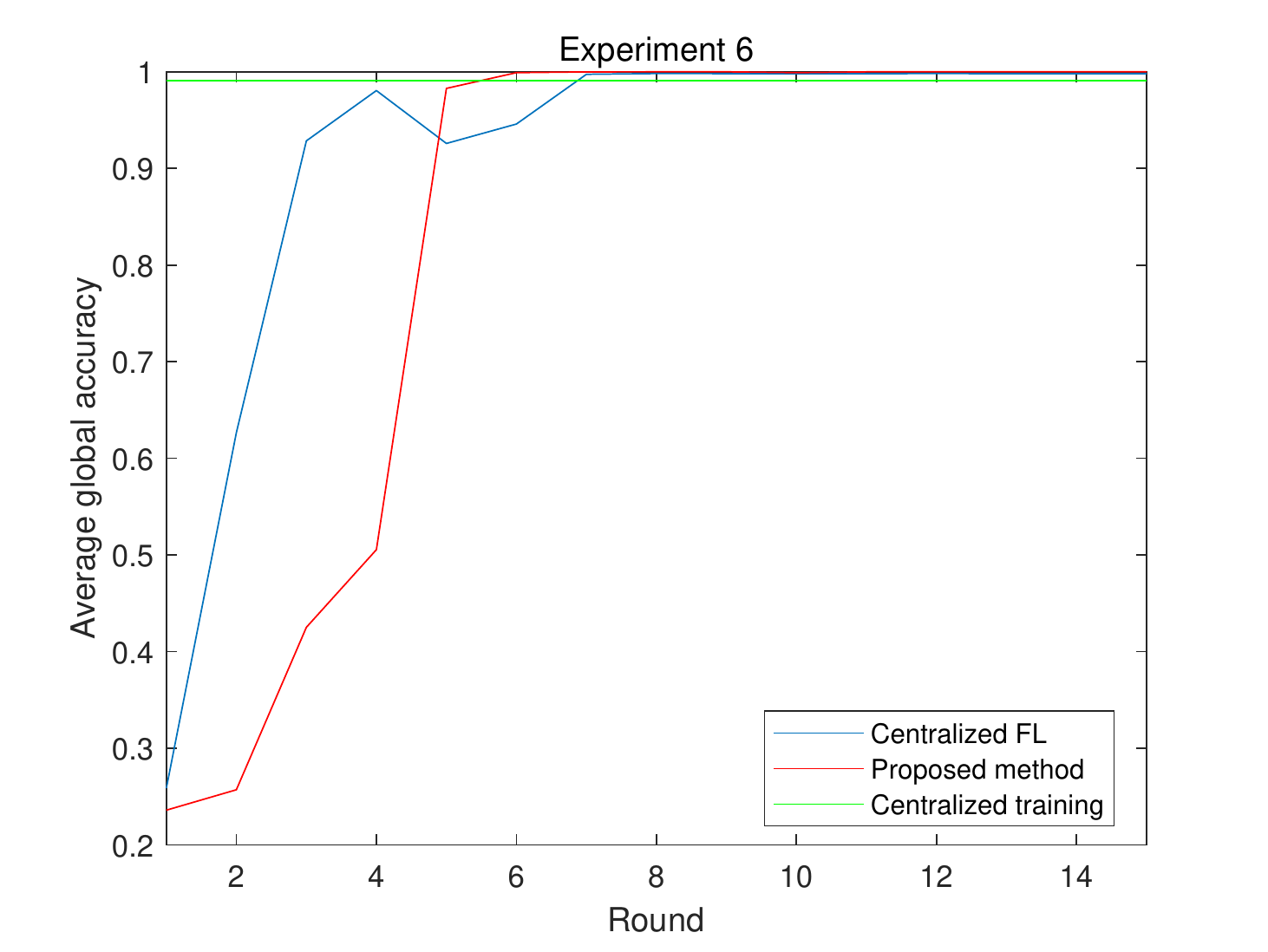} 
	} 
	\caption{Average accuracy of six simulation experiments} 
	\label{fig_global}
\end{figure*}

The six experiments designed in Table \ref{table_experiment} can be divided into three cases. Experiment 1 represents the first case, where the local data of each agent contains normal data and a kind of fault data, as shown in Fig. 15(a). Moreover, the fault type of each agent is different, which has the strongest data heterogeneity.
Experiments 2, 3, 4, and 5 represent the second case, where the local dataset of each agent contains normal data and a part of fault data, as shown in Fig. 15(b)(c)(d)(e). The fault types of various agents overlap to a certain extent, which reduces the heterogeneity of data. Experiment 6 represents the third case, where the local data of each agent contains normal data and all three types of fault data, as shown in Fig. 15(f).
At this time, the fault types of each agent are the same, which conforms to independent and identical distribution (IID), and the data heterogeneity is the lowest.
To avoid accidental errors, the result of each experiment is the average accuracy of 20 times, as shown in Fig. \ref{fig_global}.

It can be observed that the proposed method is significantly better than the centralized FL in convergence speed for the first case. For the second case, the proposed method still converges slightly faster than centralized FL under various data heterogeneity situations. For the third case, the convergence speed of the proposed method is almost the same as that of the centralized FL due to low data heterogeneity and few rounds required for convergence. 
In all cases, the proposed method can achieve 99\% accuracy after certain aggregation rounds.
In fact, there is significant data heterogeneity among actual PV stations due to differences in geographic environment and operating conditions, and the proposed method is more suitable to be adopted.

\subsubsection{Communication and Training Efficiency}
\begin{figure} [!htbp]
	\centering 
	\subfigure[Communication overhead]{ 
		\includegraphics[width=0.23\textwidth]{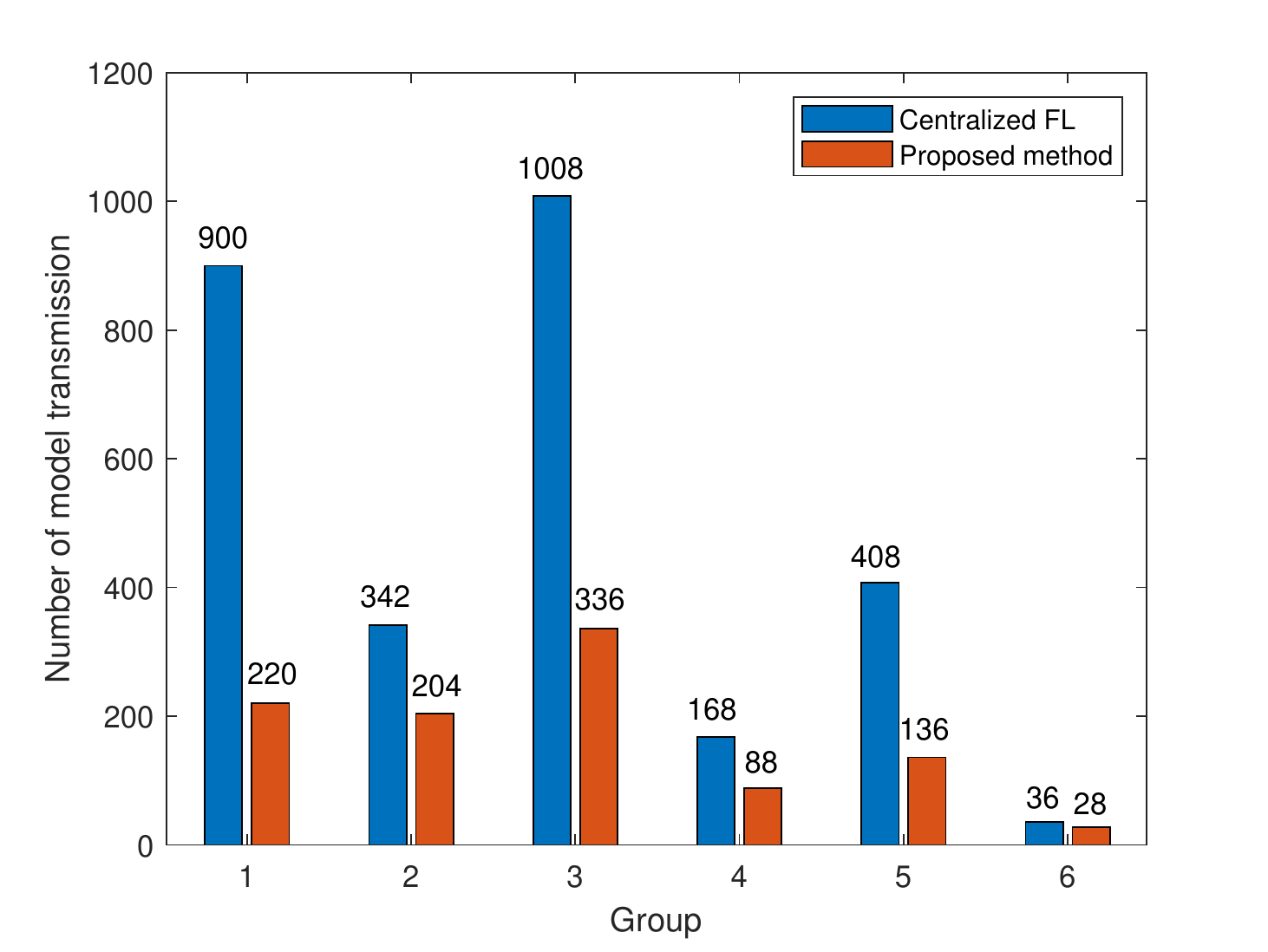} 
	} 
	\subfigure[Training time]{ 
		\includegraphics[width=0.23\textwidth]{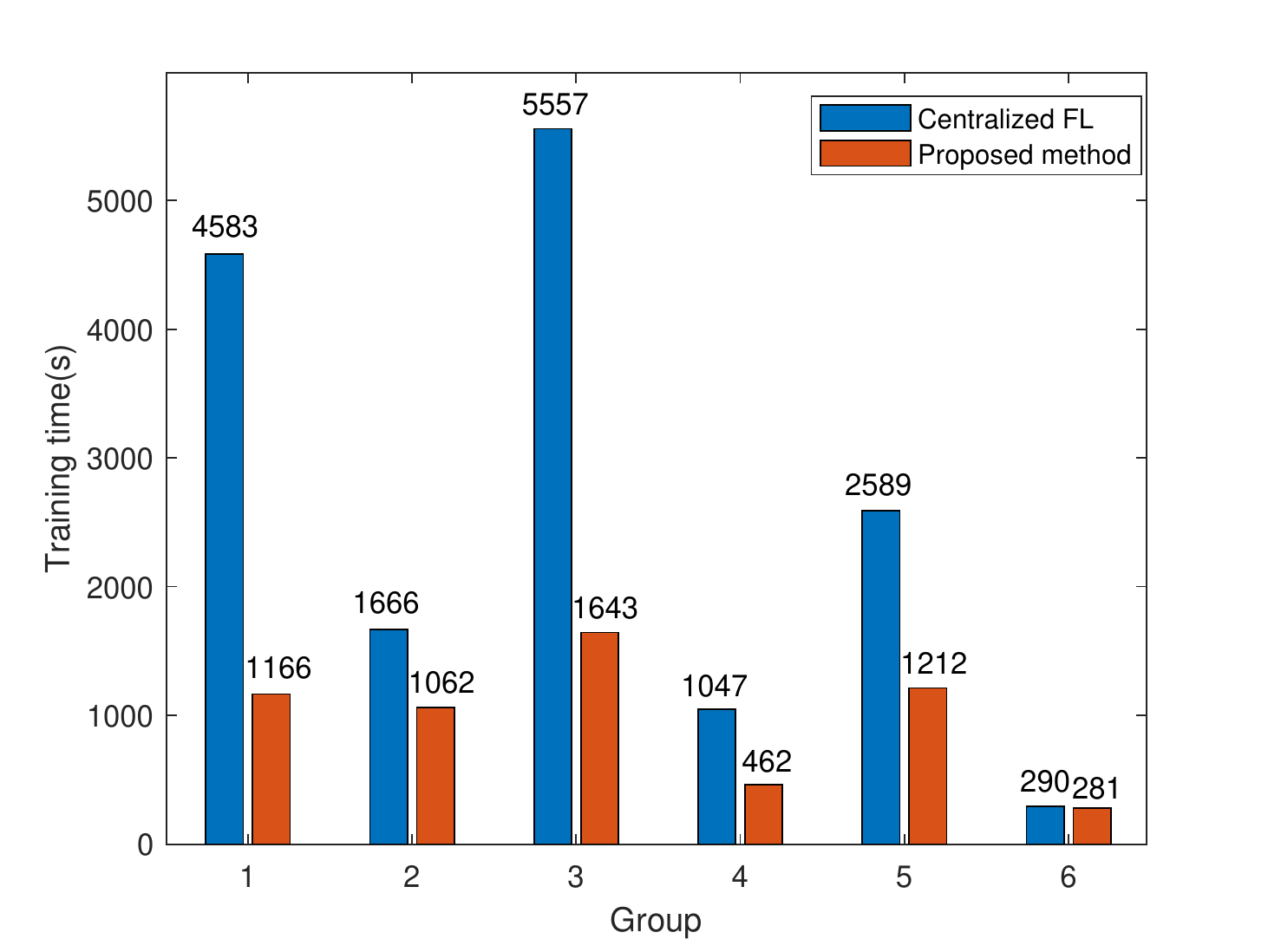} 
	} 
	\caption{Communication overhead and training time in simulations} 
	\label{fig_last}
\end{figure}

Under six groups of simulation experiments, the communication overhead and training time of the proposed method are compared with the centralized FL respectively, as shown in Fig. \ref{fig_last}.
The proposed method converges with lower communication overhead and less training time, which is consistent with previous experimental conclusions.

The above experimental and numerical simulation results show that the proposed ADFL method is applicable to various meteorological conditions.
Through the collaborative fault diagnosis of multiple PV stations, the problem of uneven distribution of fault samples is well solved. 

\begin{figure*} [!t]
	\centering 
	\subfigure[Accuracy of experiment 1]{ 
		\includegraphics[width=0.25\textwidth]{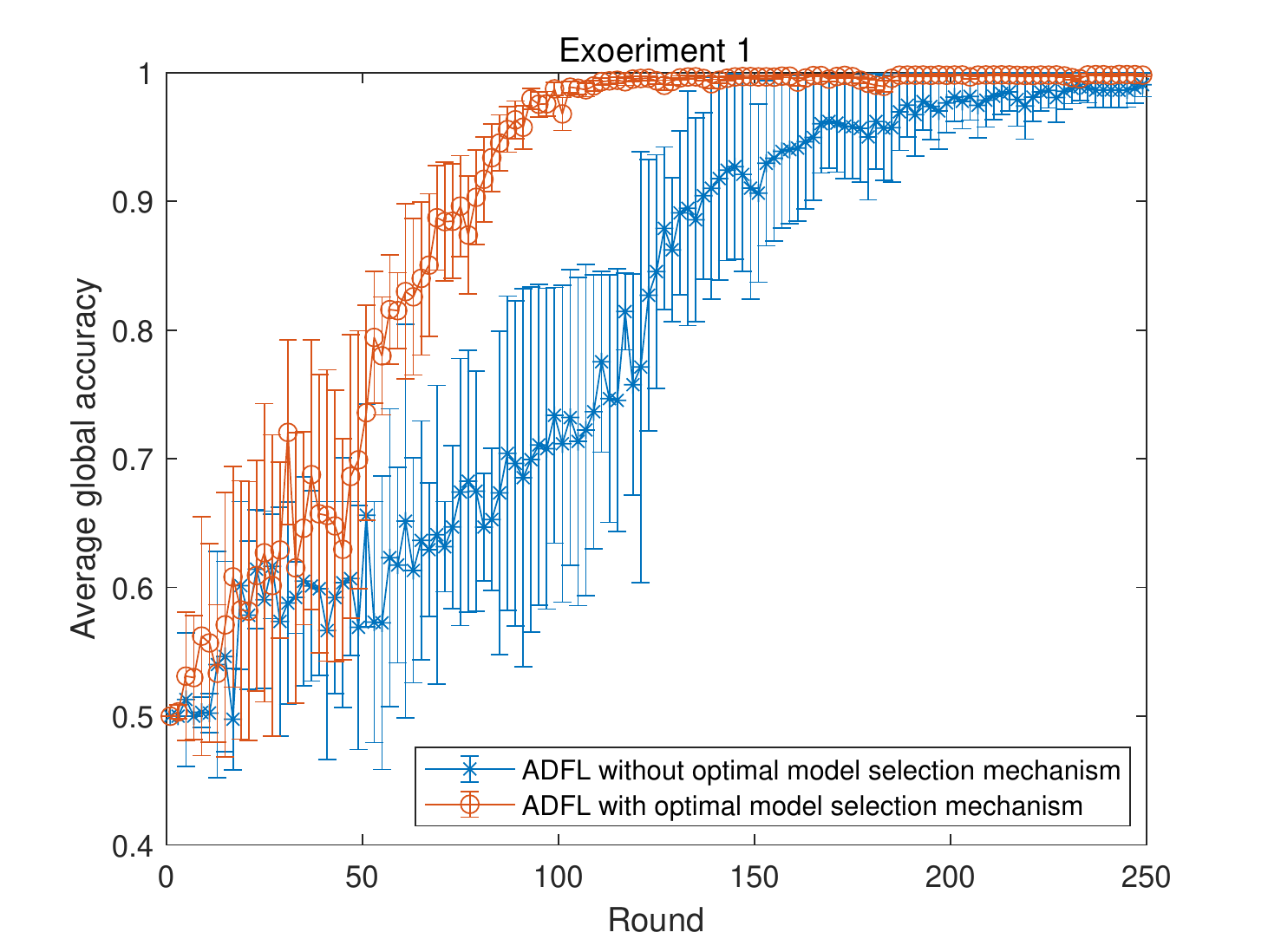} 
	} 
	\subfigure[Accuracy of experiment 2]{ 
		\includegraphics[width=0.25\textwidth]{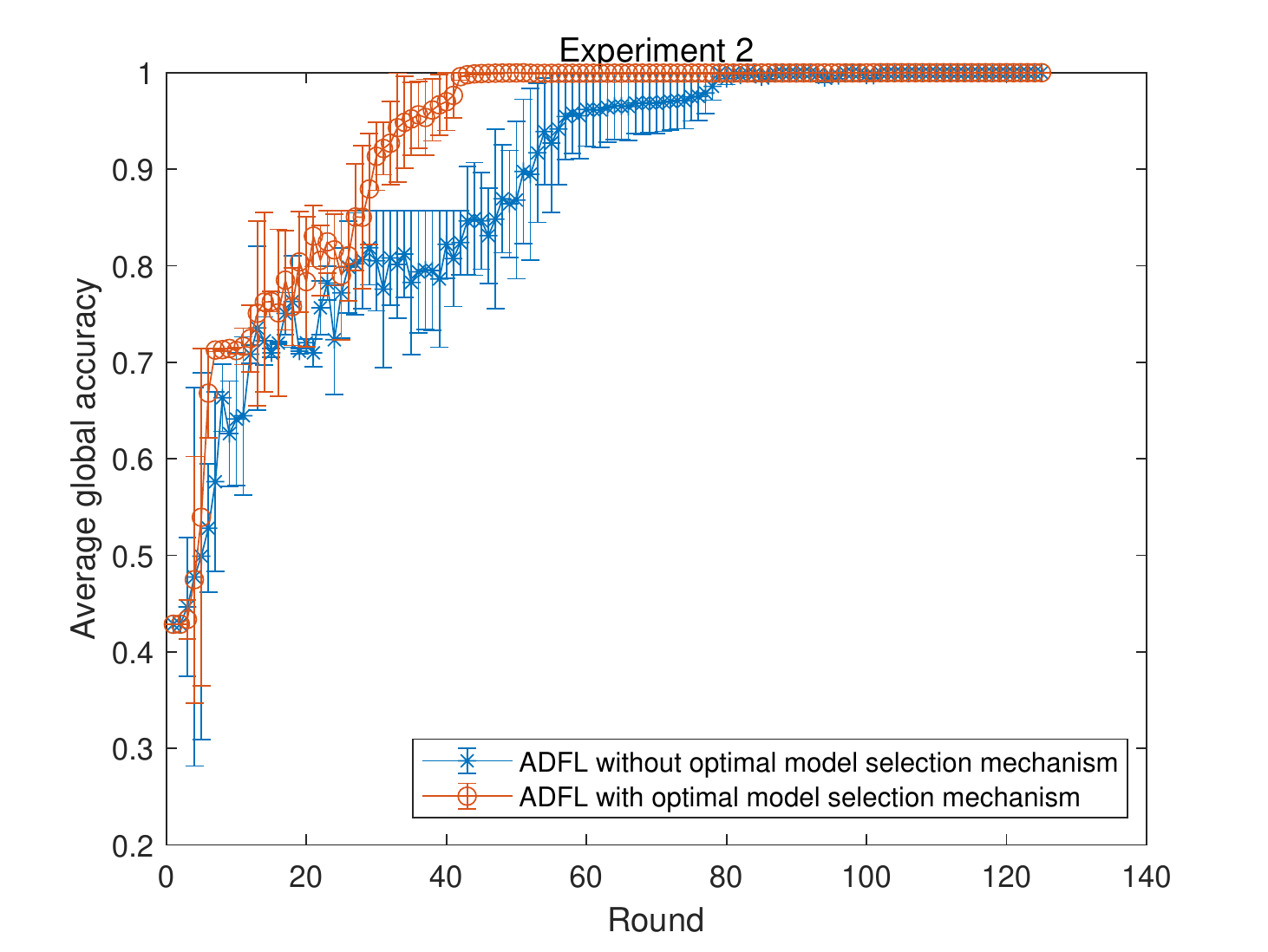} 
	} 
	\subfigure[Accuracy of experiment 3]{ 
		\includegraphics[width=0.25\textwidth]{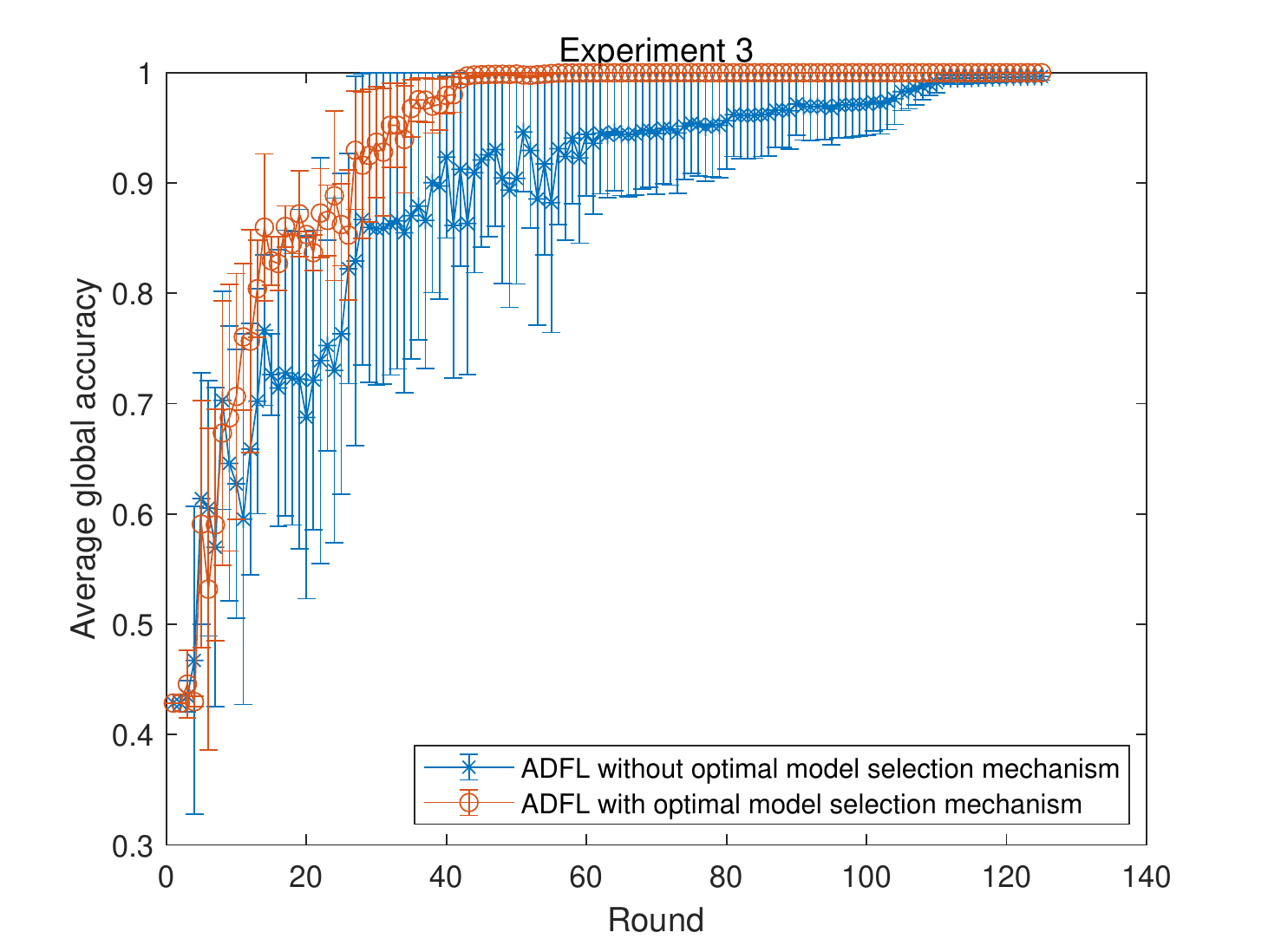} 
	} 
	\subfigure[Accuracy of experiment 4]{ 
		\includegraphics[width=0.25\textwidth]{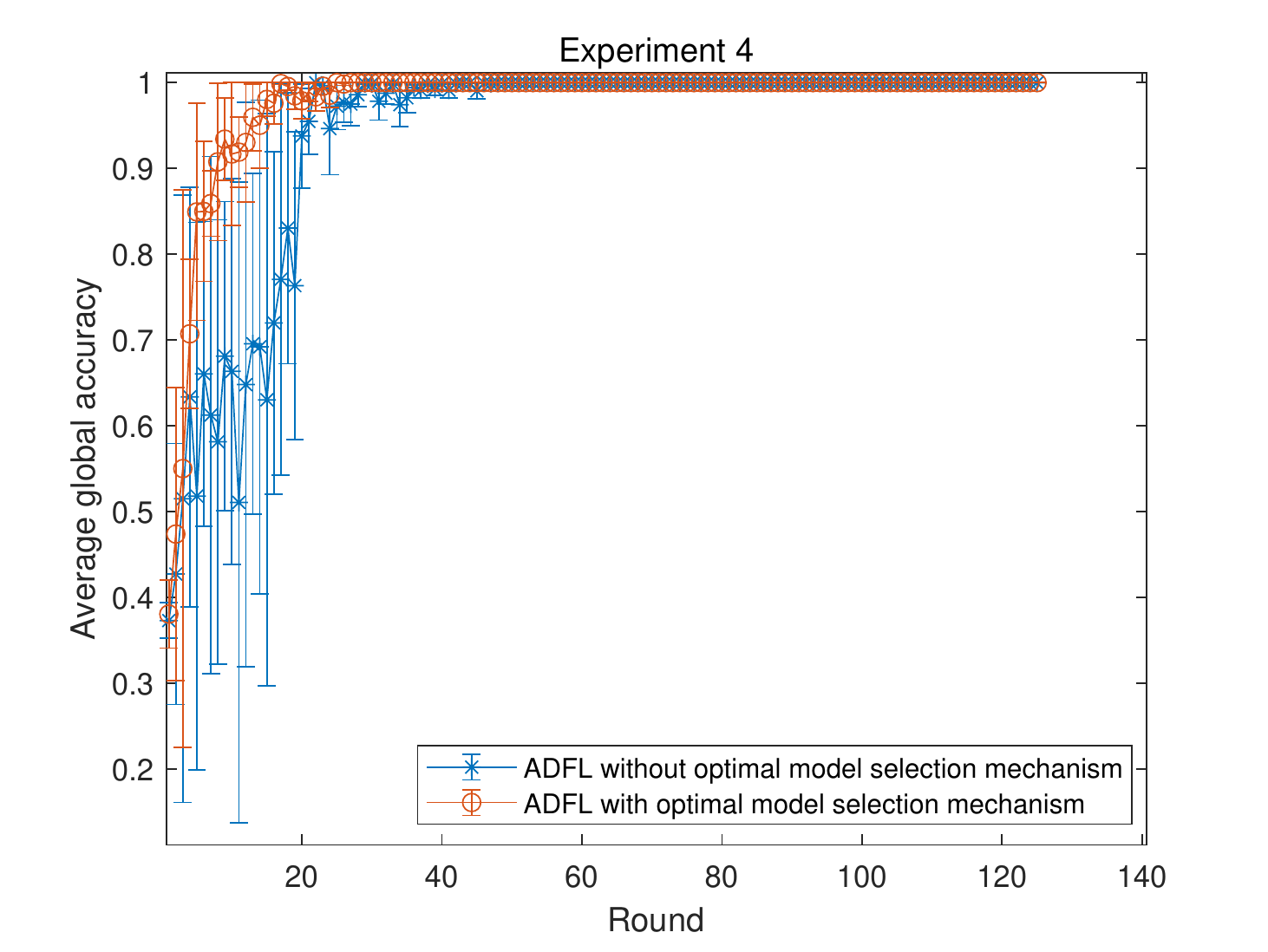} 
	} 
	\subfigure[Accuracy of experiment 5]{ 
		\includegraphics[width=0.25\textwidth]{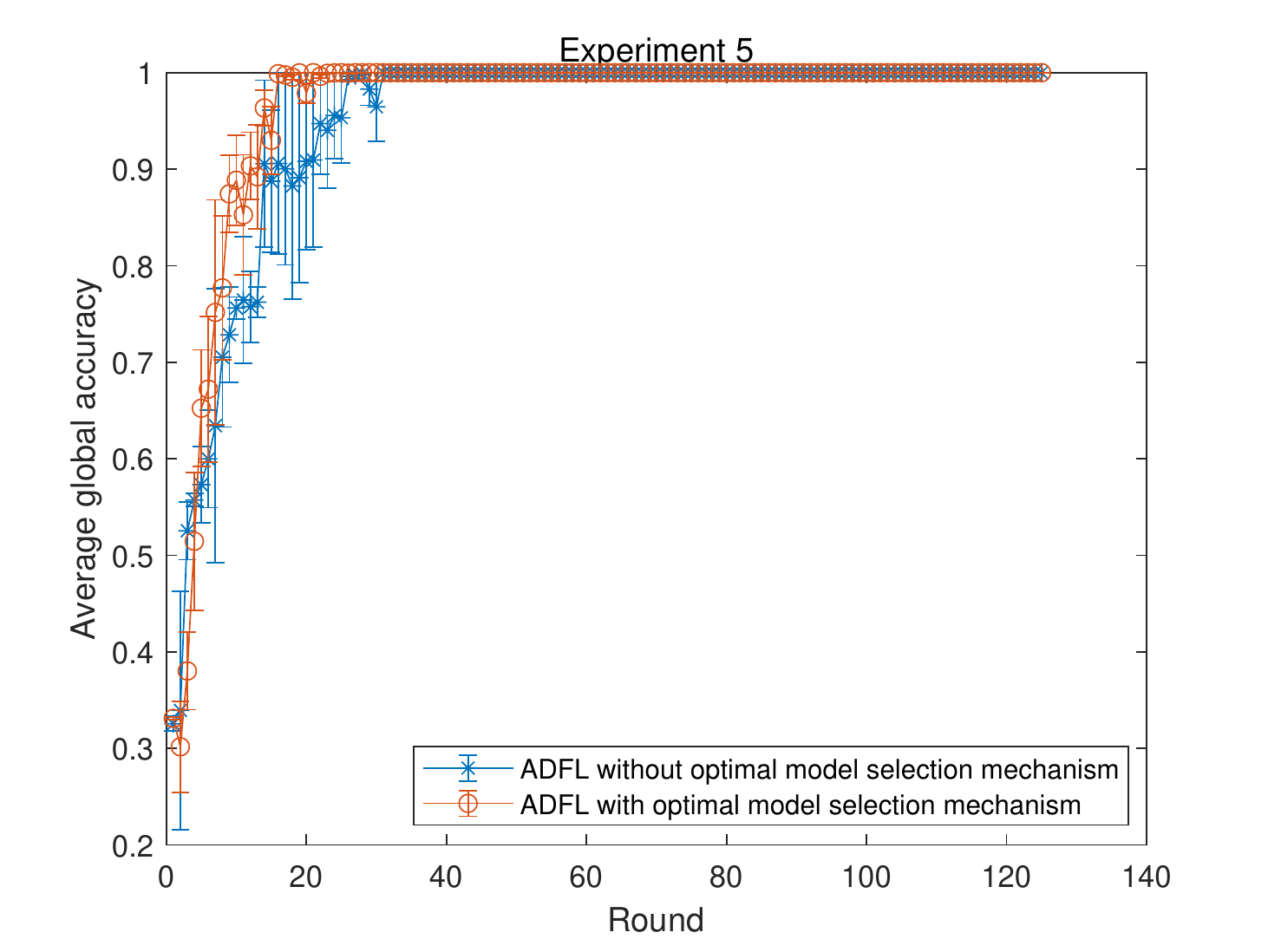} 
	} 
	\subfigure[Accuracy of experiment 6]{ 
		\includegraphics[width=0.25\textwidth]{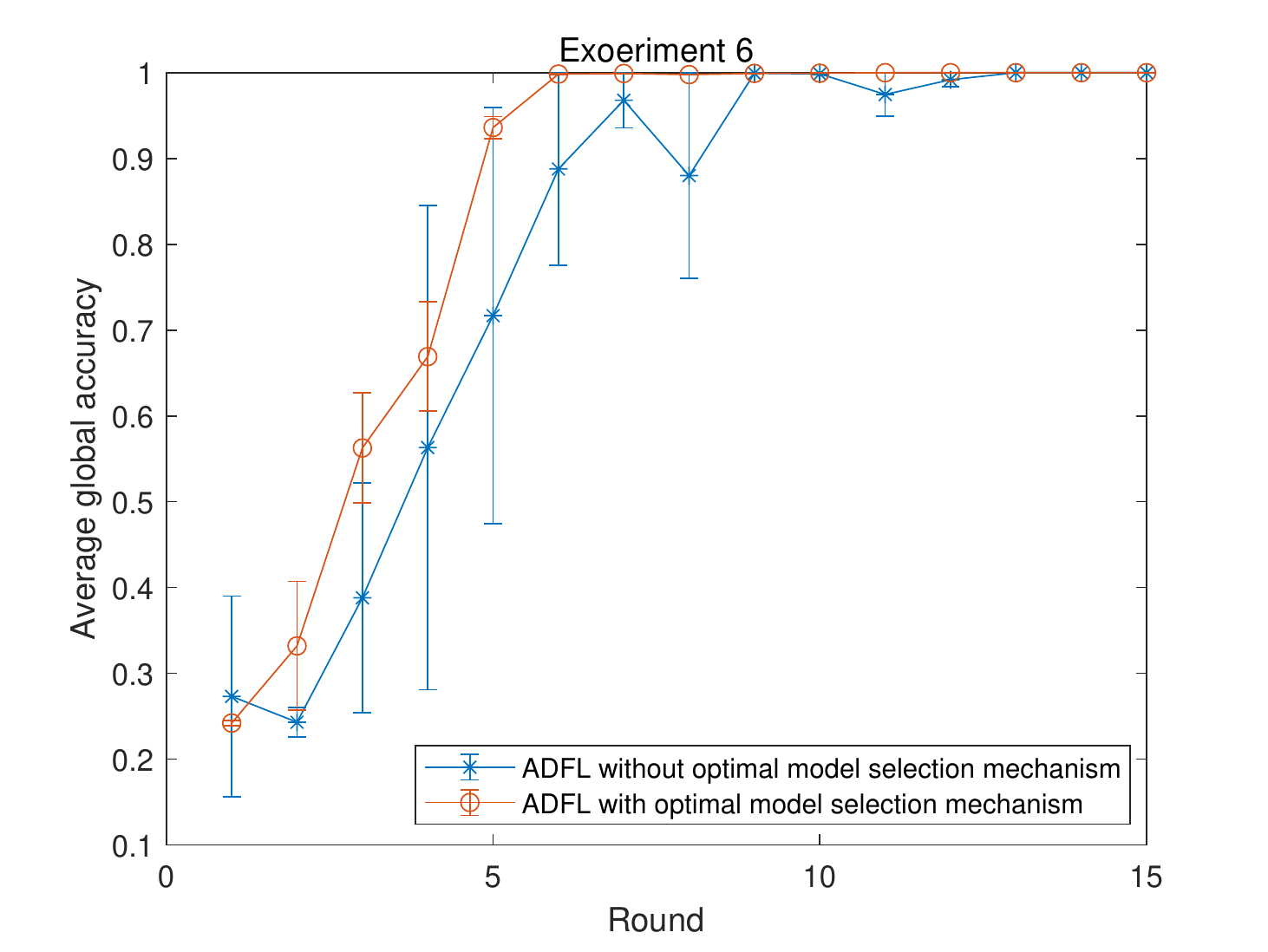} 
	} 
	\caption{Average accuracy of six simulation experiments} 
	\label{fig_Average}
\end{figure*}

\subsubsection{Optimality analysis}
To further prove that the proposed method will not converge to the local optimum in a single dataset, for the six simulation experiments in Section 5.3.1, we test the average accuracy of the global model 20 times, as shown in Fig. \ref{fig_Average}.
The error bar clearly shows the average accuracy and standard deviation of each round.
Obviously, the optimal model selection mechanism performs better in terms of average accuracy, convergence speed and fluctuation, which indicates that it is sufficient to select the model by comparing the accuracy.

In the meanwhile, the average longest continuous upload times of the same model for each agent with the optimal model selection mechanism are recorded, as shown in Table \ref{table_longest}. 
It is worth noting that the values in Table \ref{table_longest} are rounded up.
It can be seen that the number of consecutive uploads of the same model for each agent is not high, which also indicates that the proposed method will not converge to the local optimum in a single dataset.

\section{Conclusion}
In this paper, a novel asynchronous decentralized federated learning framework is proposed for PV fault diagnosis. 
The data island problem among multiple PV stations is addressed through collaborative fault diagnosis, which enhances the generalization of the model without losing accuracy. 
Different from centralized learning, each PV station trains its local model and only shares model parameters rather than original data to ensure privacy.
Considering the framework robustness and training efficiency, the global model is aggregated distributedly to avoid central node failure. 
By designing the asynchronous update strategy, the communication overhead and training time are greatly reduced.
Both experimental and numerical simulation results show that the global model obtained by the proposed method successfully diagnoses the types of faults contained in the data of all participants, and reaches the accuracy of centralized training. 
The performance of the proposed method in convergence speed, communication overhead, and training time is better than centralized federated learning.


%

\clearpage

\appendices
\section{Proof of Theorem 1}
According to (\ref{vg})-(\ref{vg2}), we have:
\begin{equation}
G_{t}([e])-G_{t}([e-1])=-l_{r} \sum_{i \in L} \frac{d_{i}}{D} \nabla F_{i}\left(w_{t}^{i}([e-1])\right)
\label{a1}
\end{equation}

Considering the (\ref{vg3}) and (\ref{a1}), we have:

\begin{subequations}
	\begin{align}
	&\left\|G_{t}([e])-z_{t}([e])\right\| \nonumber\\ 
	&=\| G_{t}([e-1])-l_{r} \sum_{i \in L} \frac{d_{i}}{D} \nabla F_{i}\left(w_{t}^{i}([e-1])\right) \nonumber\\
	&\left.\left.\quad-z_{t}([e-1\right])+l_{r} \nabla F\left(z_{t}(e-1]\right), t\right) \| \nonumber\\
	&=\| G_{t}([e-1])-l_{r} \sum_{i \in L} \frac{d_{i}}{D} \nabla F_{i}\left(w_{t}^{i}([e-1])\right) \nonumber\\
	&\left.\quad-z_{t}([e-1\right])+l_{r} \sum_{i \in L} \frac{d_{i}}{D} \nabla F_{i}\left(z_{t}([e-1]), t\right) \| \nonumber\\
	&=\| G_{t}([e-1])-z_{t}([e-1]) \nonumber\\
	&\quad-l_{r} \sum_{i \in L} \frac{d_{i}}{D}\left(\nabla F_{k}\left(w_{t}^{i}([e-1])\right)-\nabla F_{k}\left(z_{t}([e-1])\right)\right) \| \nonumber\\
	&\leq \| G_{t}([e-1])-z_{t}([e-1]) \| \nonumber\\
	&\quad+l_{r} \sum_{i \in L} \frac{d_{i}}{D}\|\left(\nabla F_{k}\left(w_{t}^{i}([e-1])\right)-\nabla F_{k}\left(z_{t}([e-1])\right)\right) \| \\
	&\leq \| G_{t}([e-1])-z_{t}([e-1]) \| \nonumber\\
	&\quad+l_{r} H \sum_{i \in L} \frac{d_{i}}{D}\| w_{t}^{i}([e-1])- z_{t}([e-1]\| \\\
	&\leq \| G_{t}([e-1])-z_{t}([e-1]) \| \nonumber\\
	&+l_{r} H \sum_{i \in L} \frac{d_{i}}{D} \frac{\delta_{k}}{H}\left((l_{r} H+1)^{e-1}-1\right) \\	
	&=\| G_{t}([e-1])-z_{t}([e-1]) \| \nonumber\\
	&\quad +l_{r} \sum_{i \in L} \frac{d_{i}}{D} \delta_{k}\left((l_{r} H+1)^{e-1}-1\right) \nonumber\
	\end{align}	
	\label{t1}
\end{subequations}

\noindent where (\ref{t1}a) is from triangle inequality, (\ref{t1}b) is obtained beacause $F_{k}(\cdot)$ is $H$-smooth from Assumption 1, and (\ref{t1}c) is based on Lemma 2 in \cite{36Wang}.

Since $\sum_{i \in L} \frac{d_{i}}{D} \leq \left(\sum_{i \in L} \frac{d_{i}}{D}+\sum_{j \notin L} \frac{d_{j}}{D}\right)=1\, and\, \delta_{k} \leq \bar{\delta}$, we have: \\				
\begin{equation}
\begin{aligned}	
&\left\|G_{t}([e])-z_{t}([e])\right\| \\
& \leq\| G_{t}([e-1])-z_{t}([e-1]) \| \\
&\quad +l_{r} \sum_{i \in L} \frac{d_{i}}{D} \delta_{k}\left((l_{r} H+1)^{e-1}-1\right) \\		
&\leq\| G_{t}([e-1])-z_{t}([e-1]) \| +l_{r} \bar{\delta} \left((l_{r} H+1)^{e-1}-1\right) \\	
\end{aligned}
\end{equation}

Thus, we have:
\begin{equation}
\begin{aligned}	
&\| G_{t}([e])-z_{t}([e]) \| - \| G_{t}([e-1])-z_{t}([e-1]) \| \\
&\leq l_{r} \bar{\delta} \left((l_{r} H+1)^{e-1}-1\right)
\end{aligned}
\label{vsum}
\end{equation}

According to the definition of the auxiliary model $z_{t}([e])$, we have $\|z_{t}([0]) - G_{t}([0])\|=0$. By summing up (\ref{vsum}) over $e \in (0,\tau]$, we have:
\begin{equation}
\begin{aligned}
&\left\|G_{t}([e])-z_{t}([e])\right\|\\
&=\sum_{i = 1}^{e}\left(\left\|G_{t}([i])-z_{t}([i])\right\|-\| G_{t}([i-1])-z_{t}([i-1]) \|\right)\\
&\leq l_{r} \bar{\delta} \sum_{i=1}^{e}\left((l_{r} H+1)^{i-1}-1\right)\\
&=l_{r} \bar{\delta} \frac{\left(1-(l_{r} H+1)^{z}\right)}{-l_{r} H}-l_{r} \bar{\delta}(z)\\
&=\frac{\bar{\delta}}{H}\left((l_{r} H+1)^{e}-1\right)-l_{r} \bar{\delta}(z)\\
&=\bar{u}(e)
\end{aligned}
\end{equation}

Considering that $F(w,t)$ in Assumption 1 is $\rho$-Lipschitz, we have:
\begin{equation}
\begin{aligned}
F(G_{t}([e]))-F(z_{t}([e])) & \leq \|F(G_{t}([e]))-F(z_{t}([e]))\| \\
&\leq \rho\|G_{t}([e])-z_{t}([e])\|\\
&\leq \rho\bar{u}(e)
\end{aligned}
\end{equation}
Thus, Theorem 1 is proved.

\section{Proof of Theorem 2}
%
First, we analyze the convergence of $\left\|{z}_{t}(e+1)-{G}^{*}\right\|^{2} $ when $l_{r} \leq \frac{1}{H}$:
\begin{subequations}
	\begin{align}
	&\left\|{z}_{t}(e+1)-{G}^{*}\right\|^{2} \nonumber\\
	&=\left\|{z}_{t}([e])-l_{r} \nabla F\left({z}_{t}([e])\right)-{G}^{*}\right\|^{2}  \nonumber\\
	&=\left\|{z}_{t}([e])-{G}^{*}\right\|^{2}-2 l_{r} \nabla F\left({z}_{t}([e])\right)^{{T}}\left({z}_{t}([e])-{G}^{*}\right) \nonumber\\
	&\quad+l_{r}^{2}\left\|\nabla F\left({z}_{t}([e])\right)\right\|^{2} \\
	&<\left\|{z}_{t}([e])-{G}^{*}\right\|^{2}-l_{r} \frac{\left\|\nabla F\left({z}_{t}([e])\right)\right\|^{2}}{H}+l_{r}^{2}\left\|\nabla F\left({z}_{t}([e])\right)\right\|^{2} \\
	&=\left\|{z}_{t}([e])-{G}^{*}\right\|^{2}-l_{r}\left(\frac{1}{H}-l_{r}\right)\left\|\nabla F\left({z}_{t}([e])\right)\right\|^{2} \nonumber\\
	&\leq \left\|{z}_{t}([e])-{G}^{*}\right\|^{2} 
	\end{align}
	\label{t2}
\end{subequations}

\noindent where (\ref{t2}a) is from expanding the squared normis, (\ref{t2}b) is based on lemma 3.5 in \cite{38Bubeck}.

Then, considering that $F(\cdot)$ is $H$-smooh, we derive the upper bound of $F\left({z}_{t}([e+1])\right)-F\left({z}_{t}([e])\right)$:

\begin{subequations}
	\begin{align}
	&F\left({z}_{t}([e+1])\right)-F\left({z}_{t}([e])\right) \nonumber\\
	&\leq \nabla F\left({z}_{t}([e])\right)^{\mathrm{T}}\left({z}_{t}([e+1])-{z}_{t}([e])\right) \nonumber\\
	&\quad+\frac{H}{2}\left\|{z}_{t}(e+1)-{z}_{t}([e])\right\|^{2} \\
	&\leq-l_{r} \nabla F\left({z}_{t}([e])\right)^{\mathrm{T}} \nabla F\left({z}_{t}([e])\right)+\frac{H l_{r}^{2}}{2}\left\|\nabla F\left({z}_{t}([e])\right)\right\|^{2} \\
	&\leq-l_{r}\left(1-\frac{H l_{r}}{2}\right)\left\|\nabla F\left({z}_{t}([e])\right)\right\|^{2} 
	\end{align}
	\label{t3}
\end{subequations}
\noindent where, (\ref{t3}a) is based on lemma 3.4 in \cite{38Bubeck}.

We define $\theta_{t}([e]) = F\left({z}_{t}([e])\right)-F\left({G}^{*}\right)$. Obviously, we have $\theta_{t}([e]) > 0$ for any round  $t$ and epoch $e$. From (\ref{t3}) , we obtain:
\begin{equation}
\theta_{t}([e+1]) \leq \theta_{t}([e])  -l_{r}\left(1-\frac{H l_{r}}{2}\right)\left\|\nabla F\left({z}_{t}([e])\right)\right\|^{2}
\label{lee1}
\end{equation}

Based on the convexity condition and Cauchy-Schwarz inequality, we have:
\begin{equation}
\begin{aligned}
\theta_{t}([e]) &= F\left({z}_{t}([e])\right)-F\left(G^{*}\right) \leq \nabla F\left({z}_{t}([e])\right)^{\mathrm{T}}\left({z}_{t}([e])-G^{*}\right)\\ 
&\leq \left\|\nabla F\left({z}_{t}([e])\right)\right\|\left\|{z}_{t}([e])-G^{*}\right\|
\end{aligned}
\label{lee2}
\end{equation}

Recalling (\ref{t1}) and $0<\theta_{t}([e+1])<\theta_{t}([e])$, we define $\chi \triangleq \min _{t} \frac{1}{\left\|z_{t}([0])-G^{*}\right\|^{2}}$. Combining (\ref{lee1}) and (\ref{lee2}), and dividing both sides by $\theta_{t}([e+1])\theta_{t}([e])$, we have:
\begin{equation}
\frac{1}{\theta_{t}([e+1])}-\frac{1}{\theta_{t}([e])} \geq \frac{\chi l_{r}\left(1-\frac{H l_{r}}{2}\right) \theta_{t}([e])}{\theta_{t}([e+1])} \geq \chi l_{r}\left(1-\frac{H l_{r}}{2}\right)
\end{equation}

\begin{equation}
\begin{aligned}
\frac{1}{\theta_{t}([\tau])}-\frac{1}{\theta_{t}([0])} &=\sum_{e = 0}^{\tau-1}\left(\frac{1}{\theta_{t}([e+1])}-\frac{1}{\theta_{t}([e])}\right) \\
&\geq \tau \chi l_{r}\left(1-\frac{H l_{r}}{2}\right)
\end{aligned}
\end{equation}

By summing up all the rounds, for $i=1, 2, \ldots, t$ and defining $T=t\tau$, we have:
\begin{subequations}
	\begin{align}
	&\sum_{i=1}^{t}\left(\frac{1}{\theta_{i}( [\tau])}-\frac{1}{\theta_{i}([0])}\right) \nonumber\\ &=\frac{1}{\theta_{t}([\tau])}-\frac{1}{\theta_{1}([0])}-\sum_{i=1}^{t-1}\left(\frac{1}{\theta_{i+1}([0])}-\frac{1}{\theta_{i}([\tau])}\right) \\
	&\geq T \chi l_{r}\left(1-\frac{H l_{r}}{2}\right) 
	\end{align}
	\label{t4}
\end{subequations}

For the summation term in (\ref{t4}a), we have:
\begin{subequations}
	\begin{align}
	&\sum_{i=1}^{t-1}\left(\frac{1}{\theta_{i+1}([0])}-\frac{1}{\theta_{i}([\tau])}\right) \nonumber\\
	&=\sum_{i=1}^{t-1}\frac{\theta_{i}( [\tau])-\theta_{i+1}([0])}{\theta_{i}([\tau]) \theta_{i+1}([0])} \nonumber\\
	&=\sum_{i=1}^{t-1}\frac{F\left({z}_{i}([\tau])\right)-F\left({z}_{i+1}([0])\right)}{\theta_{i}([\tau]) \theta_{i+1}([0])} \nonumber\\
	&=\sum_{i=1}^{t-1}\frac{F\left({z}_{i}([\tau])\right)-F\left({G}_{i}([\tau])\right)}{\theta_{i}([\tau]) \theta_{i+1}([0])} \\		
	& \geq \sum_{i=1}^{t-1}\frac{-\rho \bar{u}(\tau)}{\theta_{i}([\tau]) \theta_{i+1}([0])} \\
	& \geq -(t-1)\frac{-\rho \bar{u}(\tau)}{\varepsilon^{2}}  
	\end{align}
	\label{t5}
\end{subequations}

\noindent where (\ref{t5}a) is based on the definition of ${z}_{t}([e])$, and (\ref{t5}b) is derived by Theorem 1.

Combining (\ref{t4}b), (\ref{t4}c) and (\ref{t5}c), we have:\\
\begin{equation}
\frac{1}{\theta_{t}([\tau])}-\frac{1}{\theta_{1}([0])} =T \chi l_{r}\left(1-\frac{H l_{r}}{2}\right) -(t-1)\frac{-\rho \bar{u}(\tau)}{\varepsilon^{2}}
\label{fin}
\end{equation}

Based on Condition 4 in Theorem 2, we have $F(G_{t})-F\left(G^{*}\right) \geq \varepsilon$. By combining Theorem 1 and $\theta_{t}(\tau) \geq \varepsilon$, we can derive:
\begin{equation}
\begin{aligned}
&\frac{1}{F(G_{t})-F\left({G}^{*}\right)}-\frac{1}{\theta_{t}([\tau])} \\
&=\frac{F\left({z}_{t}([\tau])\right)-F(G_{t})}{\left(F(G_{t})-F\left({G}^{*}\right)\right) \theta_{t}([\tau])} \\
&\geq \frac{-\rho \bar{u}(\tau)}{\left(F(G_{t})-F\left({G}^{*}\right)\right) \theta_{t}([\tau])} \\
&\geq-\frac{\rho \bar{u}(\tau)}{\varepsilon^{2}} \\
\end{aligned}
\label{fin1}
\end{equation}   

Summing up (\ref{fin}) and (\ref{fin1}), we have:
\begin{equation}
\begin{aligned}
\frac{1}{F(G_{t})-F\left({G}^{*}\right)}-\frac{1}{\theta_{1}([0])} & \geq T \chi l_{r}\left(1-\frac{H l_{r}}{2}\right) -t\frac{\rho \bar{u}(\tau)}{\varepsilon^{2}}\\
&=T\left(\chi l_{r}\left(1-\frac{H l_{r}}{2}\right) -\frac{\rho \bar{u}(\tau)}{\tau\varepsilon^{2}}\right) \\
&=T\left(\chi \varphi -\zeta\right) \\
\end{aligned}
\label{fin2}
\end{equation} 

Based on Condition 2 in Theorem 2 and $\theta_{1}(0)>0$, we have:
\begin{equation}
\begin{aligned}
F(G_{t})-F\left({G}^{*}\right) & = \frac{1}{T\left(\chi \varphi -\zeta\right)+\frac{1}{\theta_{1}(0)}} \\
&\leq \frac{1}{T\left(\chi \varphi -\zeta\right)} \\
\end{aligned}
\end{equation}

Thus, Theorem 2 is proved.

\ifCLASSOPTIONcompsoc
  \section*{Acknowledgments}
\else
  \section*{Acknowledgment}
\fi

This work was supported by the National Key Research and Development Program of China (Grant No. 2018YFB1702300), and in part by the NSF of China (Grants No. 61731012, 62103265, 62122065 and 92167205).

\ifCLASSOPTIONcaptionsoff
  \newpage
\fi



%

\bibliographystyle{IEEEtran}
\bibliography{IEEEabrv,mybibfile}
\end{document}